\documentclass[pdflatex,sn-mathphys]{sn-jnl}
\pdfoutput=1

\jyear{2021}%

\theoremstyle{thmstyleone}%
\usepackage{subfigure}
\usepackage{multicol}
\usepackage{float}
\usepackage{natbib}
\usepackage{url}
\usepackage{hyperref} 
\usepackage[english]{babel}

\usepackage{amsmath}
\usepackage{makecell}
\usepackage{multirow}
\usepackage{hhline}
\usepackage{cellspace}
\usepackage[english]{babel}

\usepackage{booktabs}%
\newcommand{\tabitem}{~\llap{\textbullet}~~}

\theoremstyle{thmstyletwo}%

\theoremstyle{thmstylethree}%

\begin{document}

\title[Continual Learning for Real-World Autonomous Systems]{Continual Learning for Real-World Autonomous Systems: Algorithms, Challenges and Frameworks}


\author*[1]{\fnm{Khadija} \sur{Shaheen}}\email{kshaheen.msee18seecs@seecs.edu.pk}

\author[2]{\fnm{Muhammad} \sur{Abdullah Hanif}}\email{mh6117@nyu.edu}
\equalcont{These authors contributed equally to this work.}

\author[1]{\fnm{Osman} \sur{Hasan}}\email{osman.hasan@seecs.edu.pk}
\equalcont{These authors contributed equally to this work.}

\author[2]{\fnm{Muhammad} \sur{Shafique}}\email{muhammad.shafique@nyu.edu}
\equalcont{These authors contributed equally to this work.}

\affil*[1]{\orgdiv{School of Electrical Engineering and Computer Science}, \orgname{National University of Sciences and Technology (NUST)}, \orgaddress{\city{Islamabad}, \postcode{44000}, \country{Pakistan}}}

\affil[2]{\orgname{New York University Abu Dhabi (NYU AD)}, \orgaddress{\country{United Arab Emirates}}}


\abstract{



Continual learning is essential for all real-world applications, as frozen pre-trained models cannot effectively deal with non-stationary data distributions. The purpose of this study is to review the state-of-the-art methods that allow continuous learning of computational models over time. We primarily focus on the learning algorithms that perform continuous learning in an online fashion from considerably large (or infinite) sequential data and require substantially low computational and memory resources. We critically analyze the key challenges associated with continual learning for autonomous real-world systems and compare current methods in terms of computations, memory, and network/model complexity. We also briefly describe the implementations of continuous learning algorithms under three main autonomous systems, i.e., self-driving vehicles, unmanned aerial vehicles, and urban robots. The learning methods of these autonomous systems and their strengths and limitations are extensively explored in this article.

}

\keywords{Continual Learning, Online Learning, Real-World, Deep Neural Network, Memory, Computations, Real-time, Autonomous Systems, Image Recognition, Supervised Learning, Reinforcement Learning, Self-driving Cars, Unmanned Aerial Vehicles, Urban Robots. }



\maketitle

\section{Introduction}
\label{sec:introduction}

In recent years, Deep Learning (DL) has achieved remarkable success in many computer vision and audio processing applications~\cite{b1, lecun2015deep, dehghani2019universal}. However, the main focus of DL has been on developing high-accuracy Deep Neural Networks (DNNs) through offline training using a fixed and pre-defined/collected training dataset~\cite{b2}. These DNNs are designed to remain static after deployment and are incapable of adapting to changing environments. The real-world applications, specifically those related to autonomous agents, involve dealing with non-stationary data, i.e., where new data/tasks become available over time, and static models do not perform very well in such scenarios. A possible solution to this is to repeat the training process each time a distribution shift occurs. However, repeating the complete training process or even for a few epochs using an extended dataset is highly computationally intensive, which is not feasible in realistic resource-constrained scenarios. This leads to a need for an entirely new class of techniques/algorithms that can enable resource-efficient continuous learning in real-world systems. 

For real-world applications, continuous learning is characterized as a learning process over sequentially generated streaming data divided into multiple temporally bounded subsections known as tasks. While the state-of-the-art deep neural networks can be trained on a wide range of individual tasks to achieve impressive performance, learning multiple tasks sequentially remains a considerable challenge for deep learning. Standard neural networks, when trained on a new task, forget most of the information related to previous tasks; this phenomenon is referred to as catastrophic forgetting~\cite{b3, b4}.


A major challenge in realizing a real-world autonomous system capable of continuously learning and adapting over time is to prevent catastrophic forgetting. The learning model needs to maintain a balance between plasticity (the ability to adapt to new knowledge) and stability (the ability to retain prior knowledge)~\cite{b5, b6}. Excessive plasticity can cause forgetting the prior learned information while learning a new task. In addition, due to extreme stability, sequential task learning may become more difficult. This phenomenon is known as the stability-plasticity dilemma~\cite{b7, b8}. 

To address the above-mentioned challenge, several continuous learning (CL) algorithms have been proposed. These algorithms are based on three fundamental biologically inspired learning mechanism, i.e., synaptic regularization, structural plasticity, and memory replay~\cite{b9, b10, b11, b12}. 
Most CL works are focused on supervised task-based incremental learning in which one task is streamed at a time with independent and identically distributed (i.i.d) data, with the assumption that task identifier is available at all times. These assumptions are not directly applicable (in general) to real-world settings where streamed input samples/data are not i.i.d. and the task information is also not available. To realize more realistic scenarios, online continual learning (OCL) algorithms have been proposed that allow continuous learning from infinite data streams in an online fashion without the notion of task identity. 
In this survey, we highlighted some of the OCL~\cite{b23,b24,b25,b26,b27,b28,b29,b30,b31} methods that address the online data stream learning over gradually varying data distribution with sparse supervision. 

The applications of CL in real-world autonomous agents are practically infinite. These systems interact with the external environment through a wide range of sensors, so that they have a tendency to continuously adapt to the uncertainties and variations in the real-world environment. Most of the systems mainly focus only on visual perceptions to effectively understand their surroundings. Therefore, CL algorithms have been broadly proposed for image recognition applications. In this survey, we discuss CL techniques under three basic categories of real-world autonomous systems including self-driving cars~\cite{b39, b40, b41, b42, b43, b44, b45, b46, b47, b48, b49, b50, b51, b52, b53, b54, b55}, Unmanned Aerial Vehicles (UAVs)~\cite{b57, b58, b59, b60, b61} and urban robots~\cite{b62, b63, b65, b66, b67, b68, b69, b70, b64}. 

The primary objective of this study is to provide an overview of the learning strategies that establish an effective connection between CL and autonomous real-world systems. To the best of our knowledge, the previous CL reviews~\cite{b9}\cite{b13}, do not compare the state-of-the-art algorithms and experimental frameworks. In addition, they do not describe training methods, network complexity, and computing resources required to perform CL in a real environment. This survey also explains state-of-the-art learning algorithms to address all the long-standing challenges of real-world continual learning. 

\textbf{The major contributions of this  survey are as follows:}
\begin{itemize}
    \item We discuss the state-of-the-art OCL algorithms that address many significant challenges of real-world continuous learning. These algorithms make it possible to learn with high adaptability from continuous data streams, utilize less computational resources and require limited supervision. We also highlight the limitations and strengths of these algorithms.
    \item Most of the works on CL and OCL are proposed for static datasets and simulation environments that are far from realistic scenarios. In this regard, we analyze the scope and limitations of these methods for real-world autonomous systems.
   \item We present an overview of the applications of CL in real-world autonomous systems and briefly explain their experimental frameworks, network/algorithm complexity and required computing resources. We categorize these applications into three main types, i.e., self-driving cars, unmanned aerial vehicles and urban robots.  

\end{itemize}

The rest of the paper is organized as follows: Section~\ref{section2} gives an overview of the fundamental CL algorithms and their scenarios. Section~\ref{section3} explains the working process of some state-of-the-art OCL algorithms. Section~\ref{section4} presents the scope and challenges of CL under real-world scenarios. Section~\ref{section5} and~\ref{section6} summarizes applications of CL in real-world autonomous systems. Section~\ref{section7} discusses further look ahead in the field of CL. Finally, Section~\ref{section8} presents the conclusion.



\section{Continual Learning Strategies}
\label{section2}
In this section, we present an overview of three fundamental categories of CL algorithms that address the problem of catastrophic forgetting for task-based sequential learning~\cite{b38}. We also highlight the three main CL scenarios that are categorized based on differences in task structure and the information available at test time~\cite{b78}.

\subsection{Algorithms}
\subsubsection{Rehearsal}
The rehearsal method alleviates forgetting by replaying a few of the data samples of previously seen tasks. While learning new tasks, these samples enable to retain past knowledge of the model. Ideally, only those samples are stored that are representative of the overall data distribution of the previous tasks. The most famous replay technique is  Incremental Classifier and Representation Learning (iCARL)~\cite{b14}, which uses a prioritized exemplar selection methodology to choose few samples with a feature vector that best approximates the average feature vector from the current training set.
It provides an effective solution, but negatively affects resource utilization. It requires a large storage capacity to preserve the raw samples or representations learned from the previous task.
To overcome the storage challenge, pseudo-rehearsal strategies are proposed that generate past examples through a generative model trained over the previously learned data distribution. The notable approaches among these include Generative Adversarial Network (DGR)~\cite{b15} and Generative Auto-encoder (FearNet)~\cite{b16}. DGR employs a deep GAN model to generate the past data that is interleaved with new data to update task solver networks. FearNet employs a dual-memory system inspired by the brain, with a short-term memory system that learns new information quickly for recent recall and a long-term memory system that stores remote memories via a deep auto-encoder network, it also employs a module for selecting the associated memory system to use for prediction.


\subsubsection{Regularization}
The regularization approach consolidates the past knowledge by using additional loss terms that slow down the learning of important weights used in the previously learned task. It reduces the risk that new task information will change previously learned weight. Elastic Weight Consolidation (EWC))~\cite{b17} is one of the most well-known techniques, which uses a Fisher matrix to estimate the importance of weights and create an adapted regularization at the end of each task.
The model gets saturated after several tasks due to excessive regularization, which is a major drawback of this approach. Another famous regularization-based approach is knowledge distillation~\cite{b19},~\cite{b18} that preserves the prediction produced by the model learned on the previous task. This method creates a forward transfer of knowledge from a large network (teacher) to a small network (student) such that the student learns to follow the predictions of the teacher. This forward transfer is advantageous for reducing the size of an extremely large network. 

    
\subsubsection{Architectural}
The architectural approach dynamically changes the model's architecture by isolating the task-specific parameters. The new task is learned by augmenting new modules to the model while the previously learned parameters are kept unchanged. The representative methods include progressive networks (PNN)~\cite{b21} and dynamically expandable networks (DEN)~\cite{b22}. 
PNN keeps a pool of pre-trained models for each seen task from which it extracts useful features by learning lateral connections for the new task model.
DEN uses task relatedness to perform partial retraining of the network trained on old tasks and optimally increases its capacity when needed to consolidate new concepts for new tasks.
The main downside of these methods is the uncontrolled growth of the network parameters.   

\subsection{Scenarios}
\subsubsection{Domain-incremental learning (Domain-IL)}
In this scenario, the task remains the same, but the input distribution changes across sequential tasks. The task identity is unknown at test time, and the model is only needed to solve the current task.
Using a single model, this setting attempts to transfer knowledge from an old task to a new task while maintaining old task performance and achieving reasonable new task performance. The requirement to use a single-headed model ensures that the output space remains consistent. Even if the task structure remains unchanged throughout the learning process, changes in input distribution can cause forgetting. The distributional shift can also be caused due to the dynamic nature of the real world (e.g., new background, noise, and illumination conditions).


\subsubsection{Class-incremental learning (Class-IL)}
This scenario is about multi-class classification. It involves the popular real-world problem of learning new object classes incrementally. 
Each task in this scenario adds new distinct classes, and previous class data will be unavailable for future learning. Its neural network architecture has the same number of output nodes as the total number of classes, uses a single-headed approach, and all class labels are present in the same naming space. In this scenario, a model must be able to infer present and previously observed tasks and classes on its own. 

\subsubsection{Task-incremental learning (Task-IL)}
In this scenario, each task has a distinct output space, that allows learning the task-specific components.
The differences between the output spaces are determined by the output dimensions and their semantic concepts. For example, the old task could be a five-class classification problem, whereas the new task could be a single-value regression problem. For this scenario, a typical neural network has a multi-headed output layer (one head for each task) in which each task has its own output unit while the rest of the network is shared among tasks. To select the corresponding head in the multi-head evaluation, additional information, the task identifier is required, which allows it to produce an output for a specific task.

\begin{figure}
    \centering
    \includegraphics[width=3.0in]{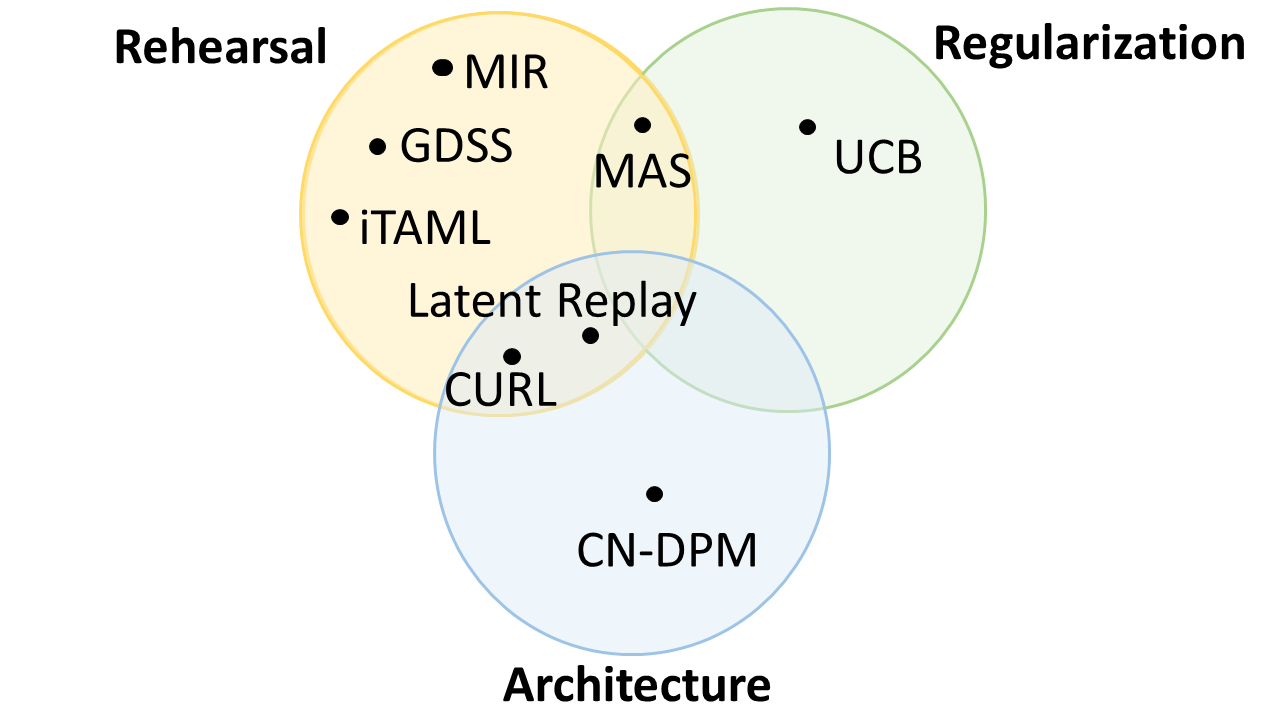}
    \caption{Ven diagram of Online Continual Learning Algorithms. }
    \label{fig4}
\end{figure}

\section{Online Continual Learning Strategies}
\label{section3}
The CL methods mentioned in the previous section require considerable computational and memory resources, which restrict their ability to learn in real-time or with an acceptable latency from data streams. Online Continuous Learning (OCL), on the other hand, applies a set of desiderata to CL systems and their objectives to ensure their ability to continuously learn from practically unlimited sequential data in an online manner. 
OCL algorithms employ task-agnostic techniques that automatically infer the boundaries and labels of the task, allowing fast continuous adaptation with minimal delay and supervision.
In this section, we discuss the recent state-of-the-art OCL strategies shown in Fig.~\ref{fig4}, that are developed by a combination of three basic CL strategies, i.e., regularization, architectural, and rehearsal~\cite{b6}. 
We also summarize the key characteristics of these methods in Table. \ref{table1}. 
\begin{table*}[]
\caption{Comparison Table of Online Continual Learning Algorithms}
\label{table1}
 \renewcommand\arraystretch{1.1}
\resizebox{1\linewidth}{!}{
\begin{tabular}{|l|l|l|l|l|}
\hline
Algorithm                                                                                                            & Learning Strategy                                                   & Network/ML Model                                                                                          & Pros                                                                                                                                                                                                                                                                                                          & Cons                                                                               \\ \hline
\begin{tabular}[c]{@{}l@{}}MAS\cite{b23} \end{tabular}                                                                    & \begin{tabular}[c]{@{}l@{}}Rehearsal \\ Regularization\end{tabular} & AlexNet                                                                                          & \begin{tabular}[c]{@{}l@{}}\tabitem Low memory\\\tabitem Less computations    \\ \tabitem High Stability\end{tabular}                                                                                                                                                                     & --                                                              \\ \hline
Latent Replay\cite{b24}                                                                             & Rehearsal                                                           & MobileNetV1                                                                                      & \begin{tabular}[c]{@{}l@{}}\tabitem Fast learning\\ \tabitem Low memory\\\tabitem Less computations\end{tabular}                                                              & Low accuracy                                                                      \\ \hline
\begin{tabular}[c]{@{}l@{}}CURL\cite{b25}\end{tabular} & \begin{tabular}[c]{@{}l@{}}Rehearsal\\ Architectural\end{tabular}   & \begin{tabular}[c]{@{}l@{}}Guassian mixture \\ models (auto-encoder)\end{tabular}                & \begin{tabular}[c]{@{}l@{}}\tabitem Less computations\\ \tabitem High accuracy\\ \tabitem Fast adaptation\\ \tabitem Unsupervised learning\end{tabular}                                                                                            & --                                                                                 \\ \hline
\begin{tabular}[c]{@{}l@{}}MIR\cite{b26}\end{tabular}                 & Rehearsal                                                           & \begin{tabular}[c]{@{}l@{}}Muti Layer Perceptron \\ (MLP), ResNet18\end{tabular}                 & \begin{tabular}[c]{@{}l@{}}\tabitem High accuracy\end{tabular} & \begin{tabular}[c]{@{}l@{}}High computations\end{tabular}                                                                  \\ \hline
\begin{tabular}[c]{@{}l@{}}GDSS\cite{b27}\end{tabular}          & Rehearsal                                                           & MLP, ResNet18                                                                                    & \begin{tabular}[c]{@{}l@{}} \tabitem High Accuracy \\\end{tabular}                           & \begin{tabular}[c]{@{}l@{}}High computations\end{tabular}           \\ \hline
\begin{tabular}[c]{@{}l@{}}CN-DPM\cite{b28}\end{tabular}         & Architectural                                                       & \begin{tabular}[c]{@{}l@{}}MLP + Variation \\ Auto-Encoder (VAE), \\ ResNet18 + VAE\end{tabular} & \begin{tabular}[c]{@{}l@{}}\tabitem Controlled expansion\\ \tabitem Less network growth \\ \tabitem Low memory\end{tabular}                            & \begin{tabular}[c]{@{}l@{}}Less accuracy\end{tabular}              \\ \hline
\begin{tabular}[c]{@{}l@{}}UCB\cite{b29}\end{tabular}                    & Regularization                                                      & \begin{tabular}[c]{@{}l@{}}Bayesian Neural \\ Network\end{tabular}                               & \begin{tabular}[c]{@{}l@{}}\tabitem Low memory\\\tabitem Less computations \end{tabular}                                                                                                                                                                                                                                                     & \begin{tabular}[c]{@{}l@{}}Lower accuracy \end{tabular} \\ \hline
\begin{tabular}[c]{@{}l@{}}iTAML\cite{b30}\end{tabular}        & Rehearsal                                                           & MLP, ResNet18                                                                                    & \begin{tabular}[c]{@{}l@{}}\tabitem High accuracy\\ \tabitem Fast adaptation\end{tabular}                                                                                                                                                                                       & \begin{tabular}[c]{@{}l@{}}High memory\end{tabular}               \\ \hline
\begin{tabular}[c]{@{}l@{}}OML\cite{b31}\end{tabular}                     & Naive                                                               & Encoder (a deep CNN)                                                                             & \begin{tabular}[c]{@{}l@{}}\tabitem High accuracy\\\tabitem Low memory\\\tabitem Less computations    \\ \tabitem Fast adaptation\\ \tabitem Less forgetting\end{tabular}                                                                         & --                                                                                 \\ \hline
\end{tabular}}
\end{table*}

\subsection{Task-free Memory Aware Synapses} 

Task-free Memory Aware Synapses (MAS)~\cite{b23} incorporates the regularization protocol into online continual learning of DNN over non-stationary streaming data without exploiting specified task boundaries. MAS computes the sensitivity of each parameter related to the previously learned knowledge and avoids updating the important parameters while learning the new information.

In the task-agnostic scenario, this protocol provides a practical and stable solution to decide "when and how to update the important weights".

\begin{figure}
    \centering
    \includegraphics[width=3in]{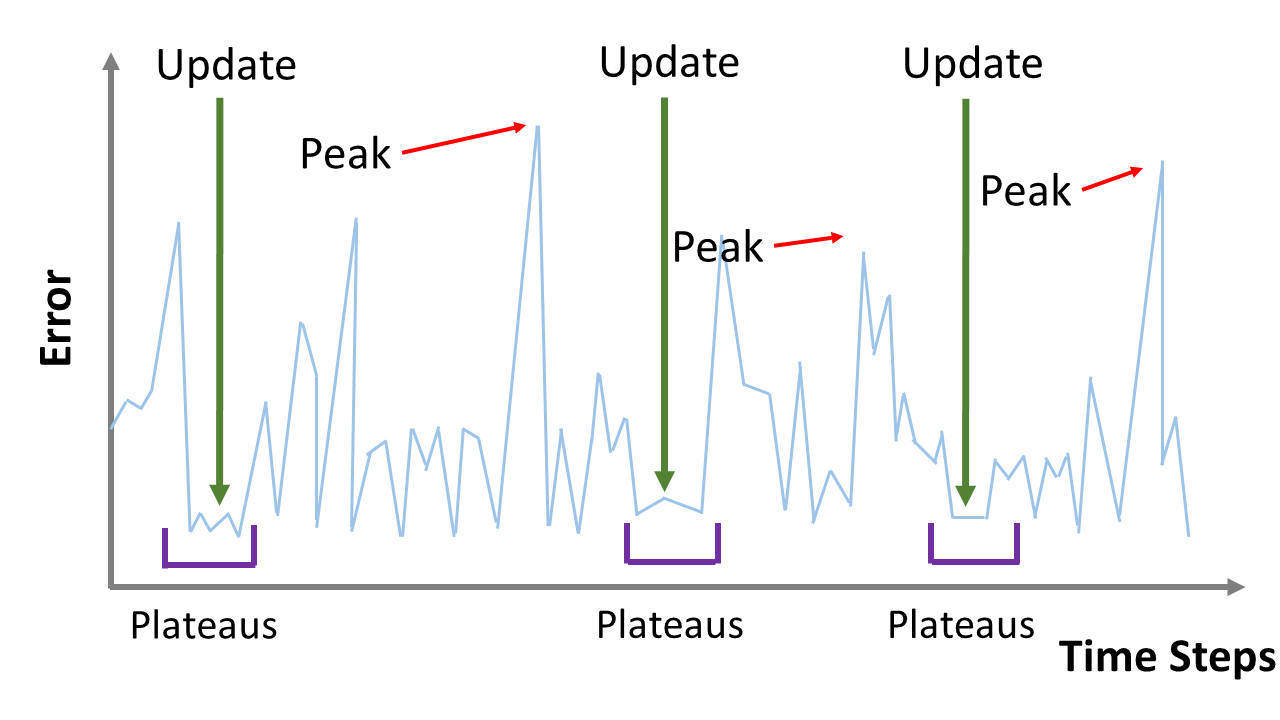}
    \caption{Memory Aware Synapses~\cite{b23}:Online learning process is triggered by analyzing the loss function surface for some defined criterion regions ,i.e., plateaus and peaks.}
    \label{fig5}
\end{figure}

By analyzing the loss surface shown in Fig.~\ref{fig5}, this method derived important information about incoming streaming data. The peaks of the loss surface represent the change in input data distribution, while the stable region named plateaus ensures the convergence of the model. The important weights are computed only during stable regions by using the current streaming data along with some hard examples of previously seen data. As a result, accurate and unbiased important weights are obtained at a varied interval of time. The DNN is only updated at the observed peak of the loss function, thus avoids to re-learn at each time step. 

Hence, this is an effective approach that prevents catastrophic forgetting without the additional memory overhead (like naive rehearsal strategy) and dynamic expansion (like architectural strategy). 

\subsection{Latent Replay}
The latent Replay~\cite{b24} algorithm enables continual learning of DNN on the edge devices. It alters the naive rehearsal strategy of CL in such a way that instead of storing the raw images of previously seen data, it stores the activation volume at some intermediate layer of DNN, thereby improving storage and computational efficiency at the expense of a slight decrease in the accuracy. 

\begin{figure}
    \centering
    \includegraphics[width=3.5in]{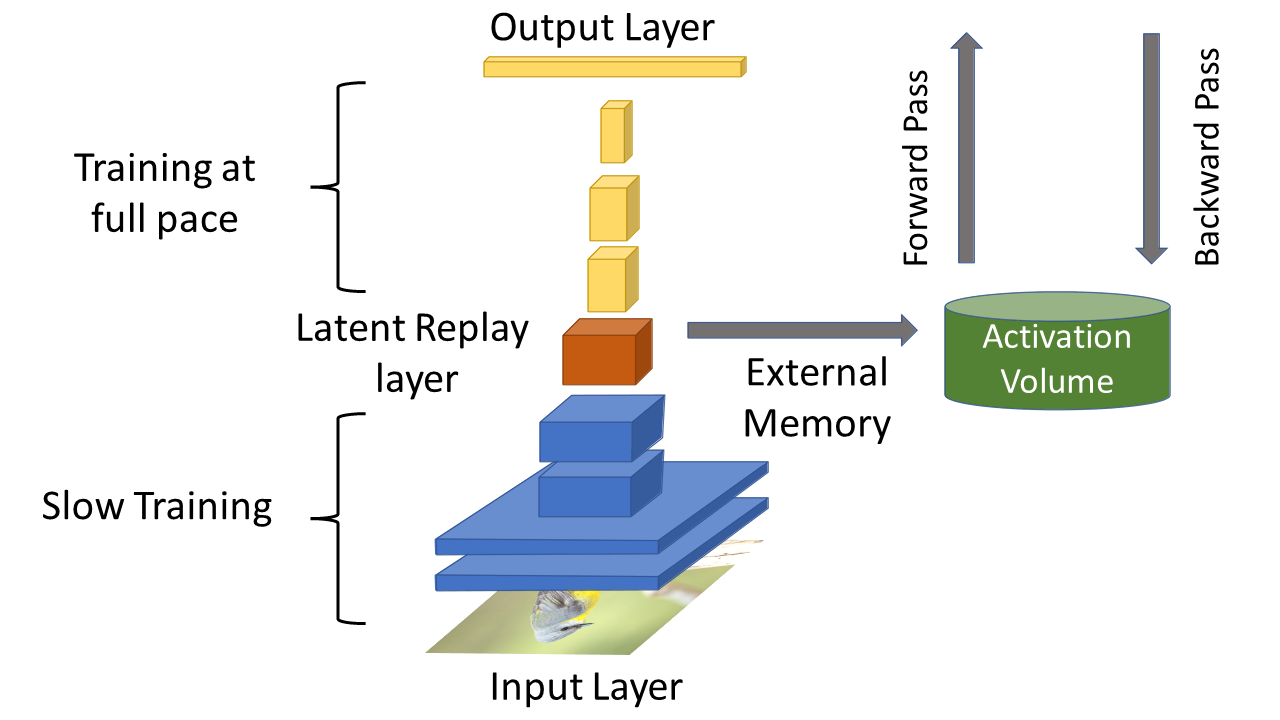}
    \caption{Latent Replay~\cite{b24}: Online training process and network architecture of DNN}
    \label{fig6}
\end{figure}

The latent replay layer would be an intermediate layer of DNN chosen according to the desired accuracy-efficiency trade-off.
The learning mechanism of this algorithm is described in Fig.~\ref{fig6}. During the learning process, the representation at the latent layer should be kept stable and constant, so the training of layers below the latent layer must be slowed down while the above layers are trained at full pace. To incorporate the rehearsal method, the activation volume of the previously seen data is stored in an external buffer. For the forward pass, the concatenation is performed at the latent layer that combines the pattern from external storage with that of the current input data, while the backward pass is stopped before the latent layer. Thus, low computations and memory requirements are achieved for the continuous training of the given DNN.

Hence, the latent replay performs incremental learning for the class incremental scenarios without forgetting previous knowledge. The computation-memory-accuracy trade-off enables the CL for the edge-only device. 

\subsection{Continual Unsupervised Representation Learning}

Continual Unsupervised Representation Learning (CURL)~\cite{b25} approach focuses on task-specific representations learning without knowing the task label and boundaries. This method has the ability to capture all fundamental properties of the data and infer the task ambiguity during online learning. It allows the DNN to dynamically expand its structure in order to capture new concepts over its lifetime and incorporate additional rehearsal-based techniques to deal with catastrophic forgetting.

\begin{figure}
    \centering
    \includegraphics[width=3.5in]{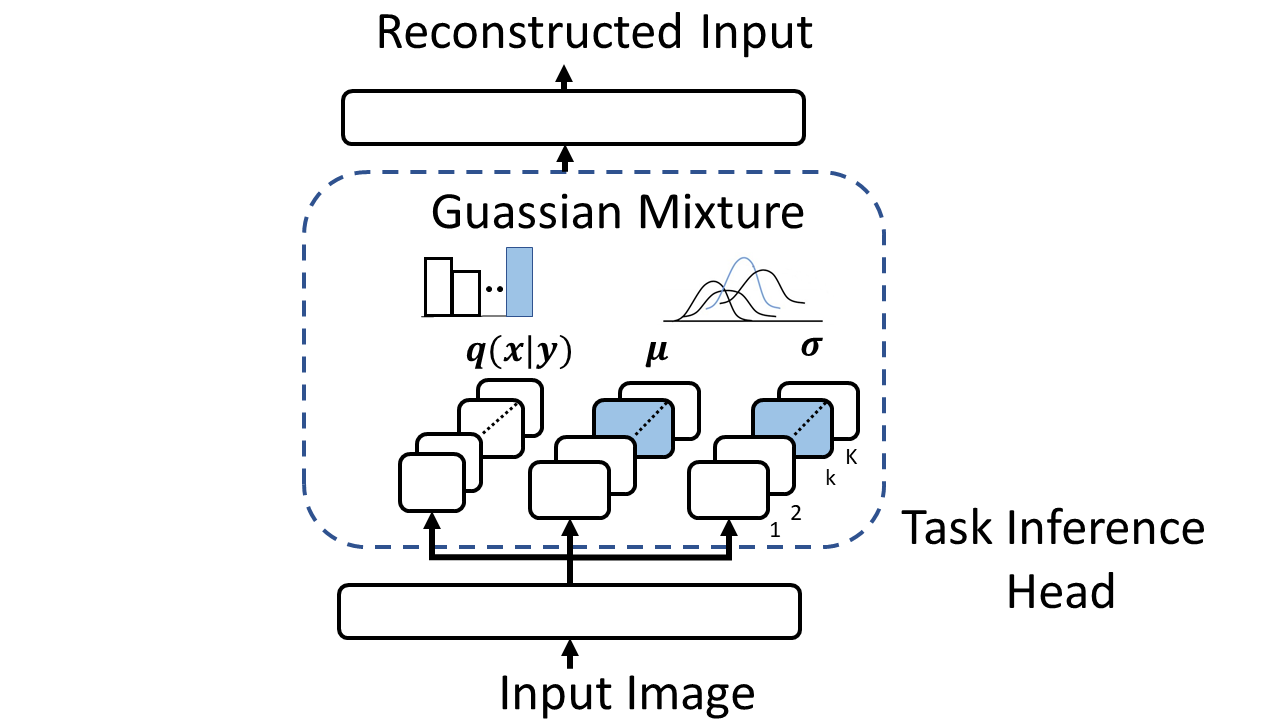}
    \caption{CURL~\cite{b25}: Network architecture and training process of a guassian mixture.}
    \label{fig7}
\end{figure}

The architecture employed for task-specific representation learning is shown in Fig.~\ref{fig7}, in which the input is first mapped to the task-inference head (softmax layer). Each head is composed of Gaussian mixture parameters that allows to learn the task-specific representations. The task inference head encodes the discrete clusters of data while the gaussian mixture encodes both inter and intra-cluster variations. Finally, the decoder maps the gaussian latent space to the reconstructed input. During unsupervised learning, the reconstruction error is minimized that creates accurate clusters at each level. 

The model learns the new task by dynamically expanding its task inference head. During inference, it stores the poorly modeled samples in a separate buffer. When this buffer reaches a critical size, a new component is added. To avoid catastrophic forgetting, the deep generative replay is employed. The update of replay is preferable to be performed before the dynamic expansion; thus, the new knowledge is consolidated more effectively.

Hence, this approach tackles the unsupervised learning issue of CL, in which task labels and boundaries are uncertain and class labels or other external supervision are also not available.


\subsection{Maximally Interfered Retrieval}

Maximally Interfered Retrieval (MIR)~\cite{b26} is a rehearsal-based OCL algorithm that is specially designed to control the sampling of replay memory. This method addresses the solution to the problem that \textit{what sample should be replayed from previously observed data}. The key principle of this method is that instead of choosing the random samples from a replay buffer, it retrieved only those samples that are maximally interfered.

The most interfered samples have identical feature representations but different labels. This method attempts to minimize the loss on the new sample without increasing loss on the most interfered previously learned samples. 

The MIR is evaluated for both experience replay and generative replay settings. It also proposed a hybrid approach for the most challenging dataset of CIFAR-10 in which both replay strategies are not feasible. The offline trained auto-encoder is used in the hybrid approach to compress the incoming samples, minimizing memory capacity space.

\begin{figure}
    \centering
    \includegraphics[width=3.0in]{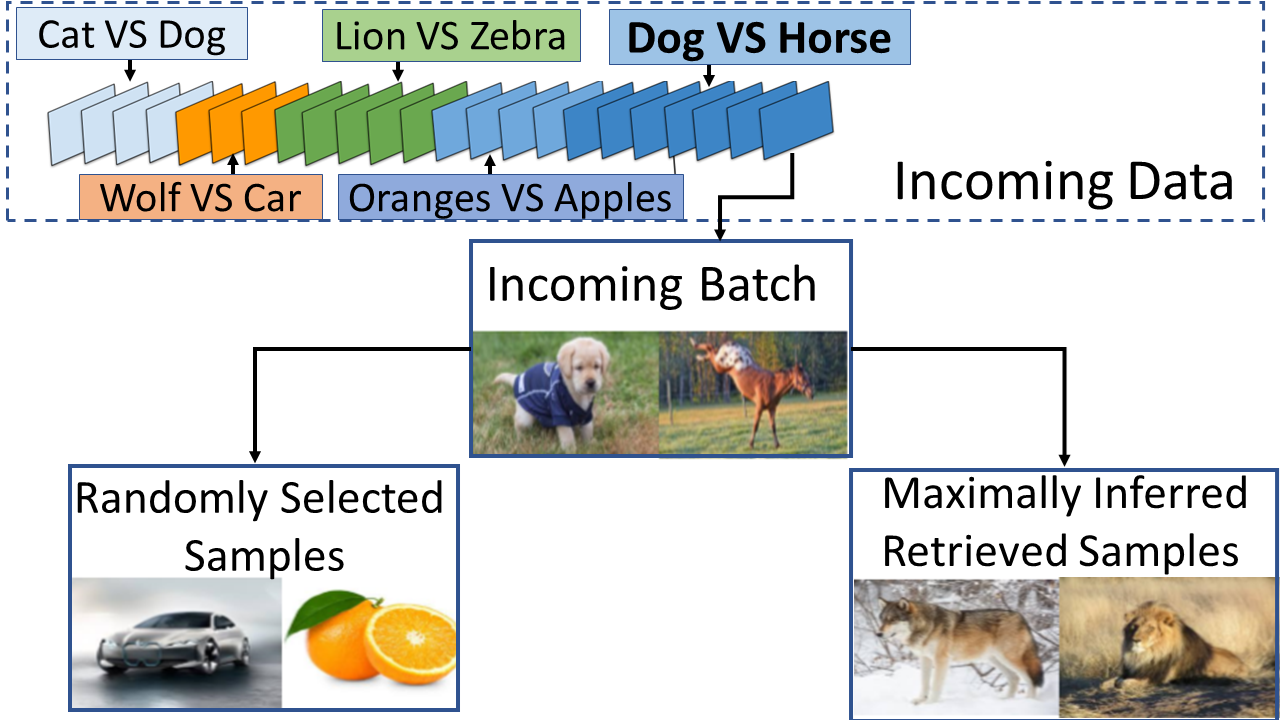}
    \caption{MIR~\cite{b26}: The comparison result between randomly sampled and MIR sampling of rehearsal buffer. The MIR selects those samples that exhibit similar appearance, for example, horse and dog. While the random sampling method selects the highly dissimilar samples.}
    \label{fig8}
\end{figure}

Hence, MIR retrieves the most useful samples from the replay buffer during online learning. As shown in Fig.~\ref{fig8}, the most interfered samples are highly correlated and will cause ambiguity if these samples are exposed to DNN at different updation steps. Thus, MIR produces the most effective batches of training data that yield high accuracy.

\subsection{Gradient based sample selection}

Gradient based sample selection (GDSS)~\cite{b27} is a rehearsal strategy that provides an effective way of populating the rehearsal buffer for the task-agnostic setting. This method selects and stores the subsets of samples from incoming streaming data that would be representative of all previous history.

This technique formulates the problem of storing samples in the replay buffer as a constraint reduction problem (here, the constraint is referred to as the gradient of the parameters produced by the training samples). 

They have proposed two solutions for constraint reduction, namely, inter-quadratic programming and the greedy approach. Inter-quadratic programming is the minimization of the empirical formula of the solid angle between the gradient of each sample in the replay buffer and the gradient of the newly added samples. 

The greedy method is proposed to reduce computational complexity, in which each new sample is assigned a score that is calculated by measuring the cosine similarity between the current samples and a finite number of random samples from the replay buffer. It selects the random samples from the replay buffer that needs to be replaced, compares their score with new samples, and stores the most effective samples with the highest score in the replay buffer.

Hence, using parameter gradient, this approach maximizes the diversity of samples in the replay buffer.

\begin{figure}
    \centering
    \includegraphics[width=3.0in]{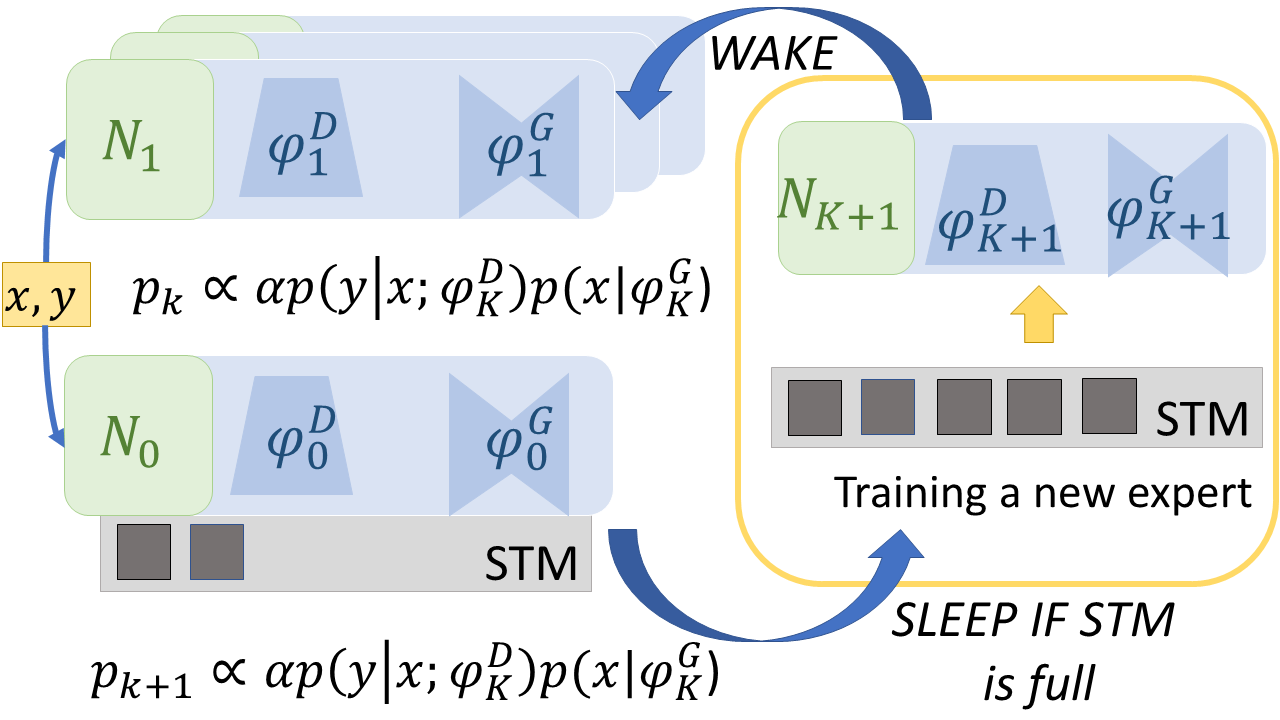}
    \caption{ Neural Dirichlet Process Mixture Model~\cite{b28}: Training Process by using gated experts and STM.}
    \label{fig9}
\end{figure}
\subsection{ Neural Dirichlet Process Mixture Model}
Neural Dirichlet Process Mixture Model (CN-DPM)~\cite{b28} is an architectural expansion strategy that accommodates the new knowledge by expanding the network without knowing task boundaries.  
The CN-DPM network is made up of a group of neural network experts who are responsible for a subset of the data. The number of experts is added by the bayesian non-parametric framework, and able to increase the capacity of the model according to data distributional variation.

As this method concentrates on the realistic condition where the task boundaries are not available, it employs a task inference scheme by using a gating strategy. The gating network is incorporated with each expert in the form of a generative mixture model that is updated at each learning step to avoid catastrophic forgetting. Consequently, it automatically infers the correct expert associated with the current and previous tasks.

The training and prediction procedure of CN-DPM is shown in Fig.~\ref{fig9}. This approach employs a prediction criterion that can be used to train an existing expert. The samples that were rejected are used to build a new expert. A single sample is not preferable for training new experts as it can lead to over-fitting. For this purpose, it employed STM (short-term memory) that stores sufficient data to create a new expert. Until the STM reaches a critical size, the learning phase is referred to as the sleep phase. While during the wake phase, the new expert is trained from data stored in STM and added to the expert pool.

This method is able to substantially control the complexity of the model during dynamic expansion. 

\subsection{Uncertainty-Guided Continual Learning With Bayesian Neural Networks}

Uncertainty-Guided Continual Bayesian Neural Networks (UCB)~\cite{b29} is a regularization technique that aims to retain the previous knowledge by controlling the learning rate of the important parameters related to the previously seen data. 

\begin{figure}
    \centering
    \includegraphics[width=3.0in]{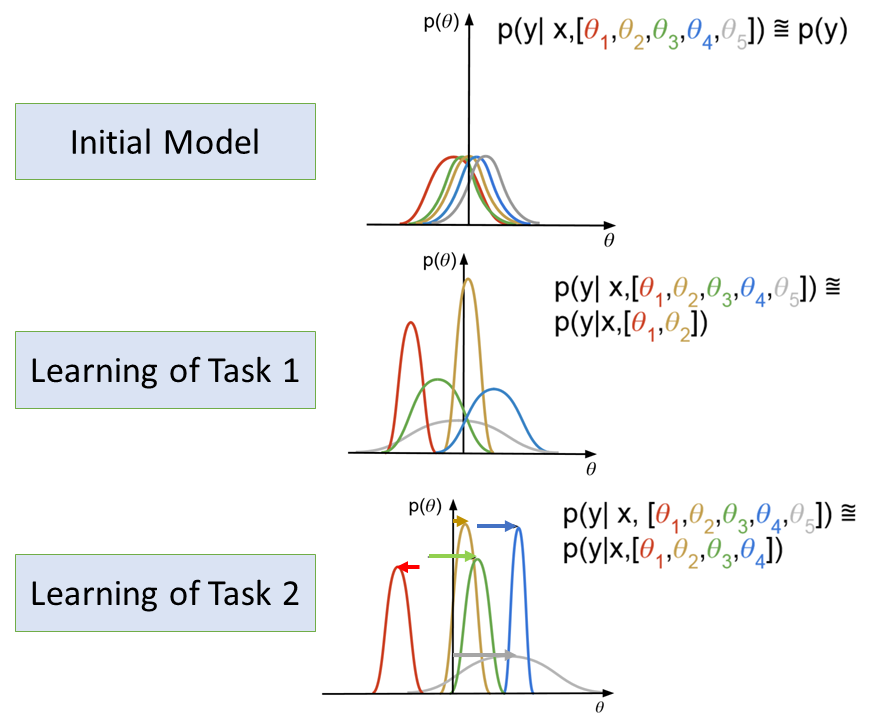}
    \caption{UCB~\cite{b29}: The change in parameter distribution while learning two different task. The parameter with high variance (uncertainty) has high capacity to learn a new task.}
    \label{fig10}
\end{figure}

This method is demonstrated for the Bayesian neural network, where the weights are represented by the mean and variance of the shared latent probability distribution. The importance of each weight has a reciprocal relation to its statistics (mean and variance). The weights with high variance (uncertainty) are more likely to be less important and have a high capacity to learn a new task. This phenomenon is elaborated in Fig.~\ref{fig10}. The learning rate of each parameter is a function of its importance factor. Thus, a higher importance factor will result in a small learning rate.

A variant of UCB with pruning is also developed that merges pruning with the bayesian neural network. The pruning avoids forgetting by using binary masks over the network; thus, a single network has enough capacity to learn all tasks without any forgetting. 
UCB with pruning takes more memory to store the binary mask.
In contrast, the UCB variant without pruning needs no additional memory and allows for slight forgetting, it has a greater capacity to learn new tasks.

\subsection{Incremental Task-Agnostic Meta-learning}
Incremental Task-Agnostic Meta-learning (iTAML)~\cite{b30} employs a meta-learning approach to search the generalized solution that is well-balanced among all tasks. By using the principle of "learning to learn", the meta-learning aims to obtain a generic model that can easily adapt to the new task. Although the key obstacle of the CL methods is a fast adaptation, meta-learning is an ideal approach for this regime. 

In addition, iTAML produces a robust generalized model with the ability to automatically classify the task identity with high precision. It exhibits both task and model agnostic nature (i.e., independent of task information and network architecture being used). 

\begin{figure}
    \centering
    \includegraphics[width=3.0in]{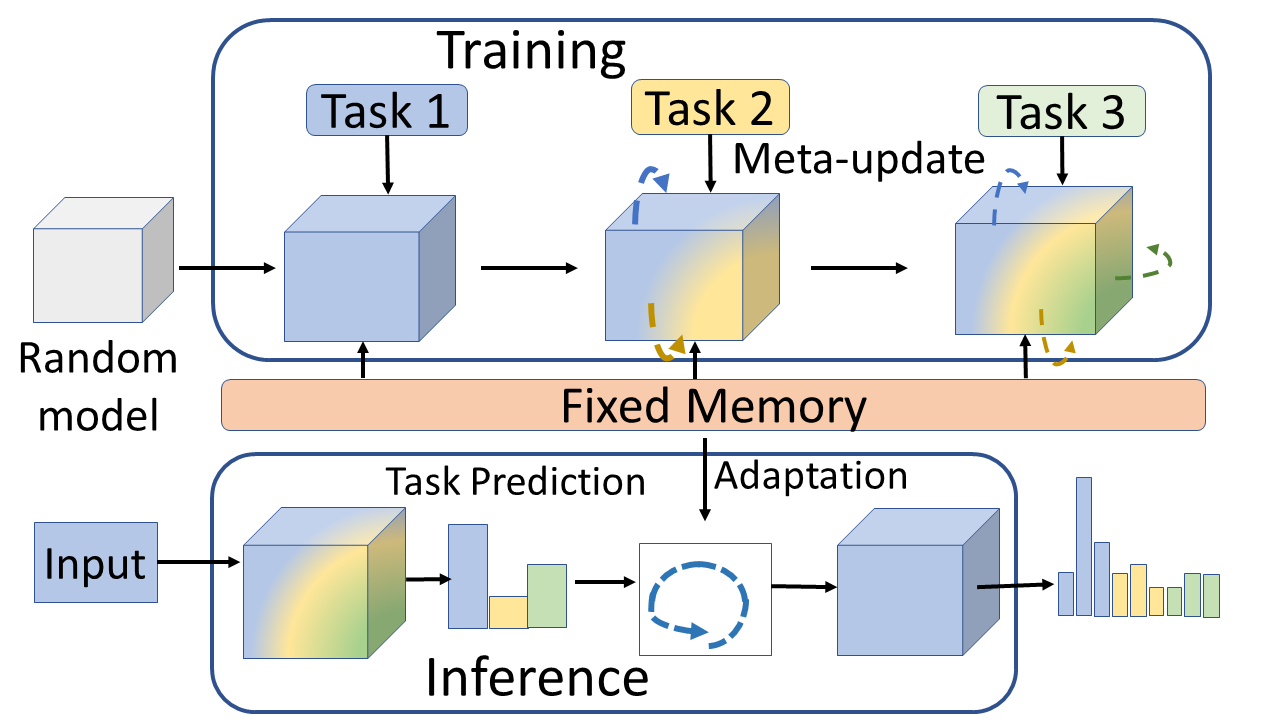}
    \caption{iTAML~\cite{b30}: Training and Inference process}
    \label{fig11}
\end{figure}

The training and inference process involved in iTAML are shown in Fig.~\ref{fig11}. It employed a momentum-based meta-learning approach during training. This approach mitigates the catastrophic forgetting as well as the data imbalancing effect. 
The generalized parameters are first used to identify the task, and then they are adapted to task-specific parameters to accurately predict the class. This method learns a shared feature space that enables fast adaptation toward desired task-specific parameters.   

Hence, this method is able to learn the infinitely large scale data stream with high adaptation and inference speed.  
\subsection{Online Aware Meta-Learning}


Online Aware Meta-Learning  (OML)~\cite{b31} attempts to learn generic representations that are sequentially more efficient for online learning. With the use of meta-objectives, this online updating strategy is more powerful than other OCL approaches. The generic representations have enough capacity to learn without catastrophic interference and accelerate future learning. 

\begin{figure}
    \centering
    \includegraphics[width=2.75in]{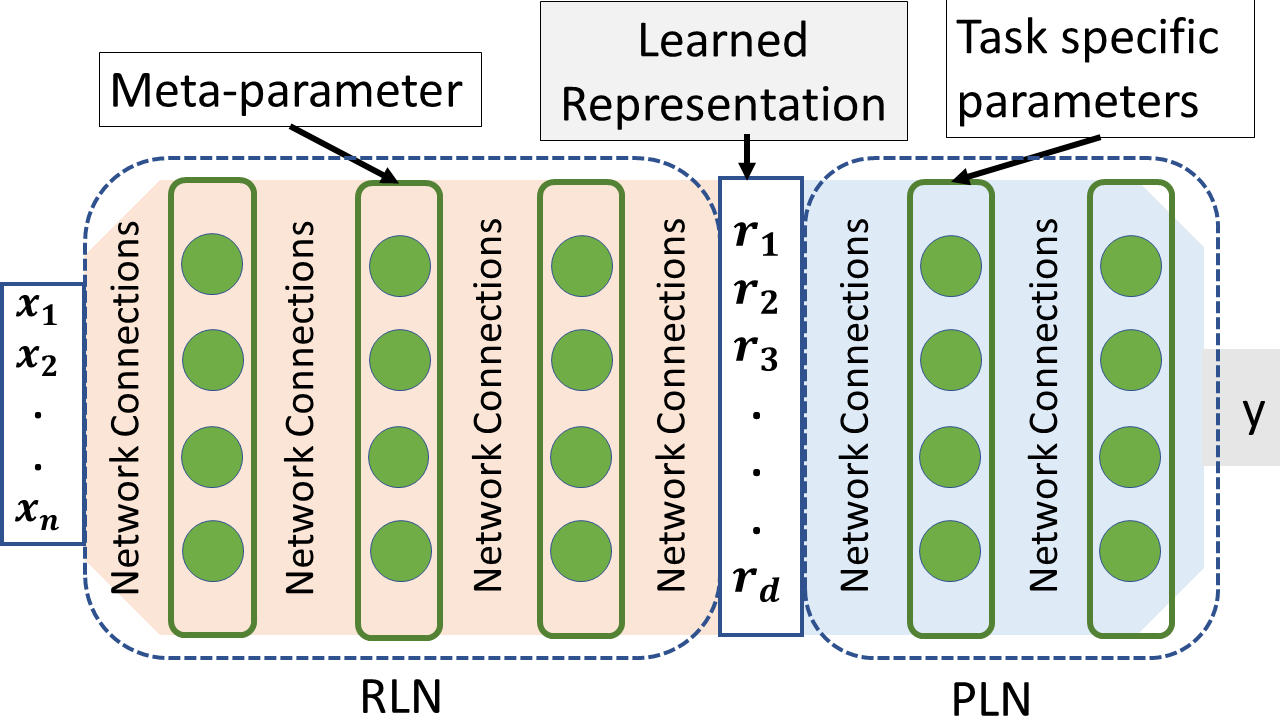}
    \caption{OML~\cite{b31}: The network architecture and training process.}
    \label{fig12}
\end{figure}

During the training, the whole network architecture shown in Fig.~\ref{fig12} is assumed to be composed of two different parts, i.e., representation learning network (RLP) and prediction learning network (PLN). The online updates for continual task-specific learning only modify the PLN portion, while RLN is updated with meta-objective and retains the generalized knowledge during online stochastic updates. 

By learning online sequentially, the objective of OML is to find a generic parameter vector that is effective for all distributions. The representations of distinct classes have different manifolds and intersections. Positive generalization is possible through parallel manifolds, making online updates more efficient. While the representations that generate such manifolds are unlikely to arise naturally, Instead, OML has the ability to identify these manifolds and achieve a highly generalized solution. The main aim of OML is to optimize the generic representations by taking the effects of OCL into account. 

\section{Continual Learning for Real-World Autonomous Agents}
\label{section4}

In this section, we first explain the importance of CL for autonomous real-world systems and then highlight the key challenges encountered by CL algorithms under real-world scenarios.

\subsection{Scope of Continual Learning for Real-world Autonomous Systems}
The autonomous agents always interact with real-world environments, where they experience non-stationary changing data distributions. In such scenarios, the frozen AI models cannot generate correct predictions as they are not trained for all possible scenarios. An adaptive model is required that learns and reacts to the non-stationary environment in real-time. 
In addition, these embedded platforms suffer from many restrictions in terms of power or memory, and that is precisely what CL aims to optimize. On the other hand, these agents have rich knowledge about their experiences. They monitor their connection with the environment, which can help them understand the principle of causality and extract the information from various sensor types (images, sound, depth, etc.). This rich knowledge allows the machines to generate powerful representations that are essential for CL algorithms.

Many CL approaches are not directly linked to autonomous real-world systems; instead, they concentrate on image processing experiments or virtual environments. In the next section, we will highlight the challenges that make CL difficult to apply in real-world environments. 
\subsection{Challenges for Continual Learning}
\subsubsection{Hardware}
The main challenge while conducting real-world CL experiments is the hardware. These agents are fragile and compact, the learning failures will affect its infrastructure, causing long delays in experiments. Moreover, the autonomous agents are an embedded platform, so they have limited memory and computation resources. Besides, these agents must exhibit energy autonomy, which allows them to complete the long experiments of CL without manual recharging or power failure. 

\subsubsection{Stability}
CL learns different experiences sequentially over time. It retains the knowledge of each experience through certain memory handling techniques. During sequential learning, if only one experience is corrupted, then the whole learning system can collapse. As the previous data is not available, so it is very difficult to restore the actual problem. Corruption of one learning experience will lead to memory degradation and then to the deterioration of the model while learning subsequent tasks.
 
\subsubsection{Data Sampling}
The autonomous agents must have the ability to gather useful training data while interacting with an unknown and unlimited environment~\cite{b32}. Data serves as the base for exploring and understanding the world. This problem is usually addressed by RL algorithms where agents generate the goals and explore the environment~\cite{b35}. The self-supervision also enables an effective exploration of the surrounding. Curiosity and self-supervision allow to search the environment and gather data from all task~\cite{b33}\cite{b34}. Data aggregation is a big challenge for lifelong learning as redundant data is highly undesirable for CL strategies. 

\subsubsection{Data Labelling}
The annotation or labeling of online training data is highly tedious for autonomous systems. External expert supervision is mostly employed for safety-critical conditions. Likewise, a reward function used in RL also helps the autonomous agent to automatically understand the environment~\cite{b37}. Furthermore, they also have the ability of self-supervision in which they exploit the sensor's data to define some abstract rules about the environment~\cite{b36}.    
\subsection{Learning Paradigms for Continual Learning in Real-world Autonomous System:}
Continuous learning has been extensively studied with regard to three primary machine learning paradigms in real-world autonomous systems. These paradigms are defined based on the kind of supervision signal.
\subsubsection{Supervised Learning}
Supervised learning is the process of learning from labeled training data that contains complete knowledge about input and output of the system ~\cite{russell2009artificial}. 
Supervised CL allows to learn from sequential streaming data in order to map data to labels for the entire sequence. The use of CL in supervised learning reduced the complexity of the learning algorithm.


\subsubsection{Unsupervised Learning}
Unsupervised learning includes algorithms that do not require labels. The most common example is generative models that are trained to reproduce the input data distribution. 
In the CL setting, the generative model is updated when the distributional shift occurs, resulting in the final output being generated from the entire input distribution. The most common use of these models has been observed in generative replay CL strategy.

Unsupervised CL may be useful in autonomous systems for creating increasingly powerful representations over time, which can then be fine-tuned with an external feedback signal from the environment. The main goal of unsupervised learning is to develop a suitable surrogate and meaningful self-supervised learning signals, to learn robust and adaptive representations.
 
\subsubsection{Reinforcement Learning}
Reinforcement Learning (RL) uses a reward function as a label to train an agent that performs the sequence of actions in a specific environment. RL complex environments do not provide access to all data at once, so it can be viewed as a CL scenario. To assume i.i.d. data distribution, RL uses several important components of CL models, such as the ability to learn multiple agents in parallel or the use of a replay memory ( CL rehearsal strategy). Furthermore, the Fisher Matrix used in a popular stable RL method, the TRPO algorithm, improves the learning process, similar to the CL strategy (EWC).
The challenges of CL are the severe limitations of reinforcement learning algorithms. Consequently, improving CL will also improve reinforcement learning performance. 
\section {Applications of Continual Learning in Real-World Use Cases}
\label{section5}
\subsection{Self-Driving Cars}

The self-driving car uses a deep learning-based decision-making system that maps on-board sensory data (cameras, lidar, inertial sensors, and so on) to control output. It always interacts with evolving real-world non-stationary environments. Moreover, safety and comfort is also an important aspect of an autonomous vehicle. The frozen models are not sufficient to deal with such situations as they are not trained for all possible scenarios. Therefore, they require an efficient self-adaptive method that detects and adapts the novel driving scenarios on the fly.

The different experimental frameworks of CL in the field of self-driving cars are explained in this section. A brief summary of these frameworks/solutions is also presented in Table~\ref{my-label_s_1}. 


\begin{sidewaystable}
\sidewaystablefn%
\begin{center}
\begin{minipage}{\textheight}
 \caption{Summary Table of Real time Continual learning applications in Self-driving cars}
  \label{my-label_s_1}
  \centering

 \renewcommand\arraystretch{0}
  \resizebox{1\linewidth}{!}{
  \begin{tabular}{p{2cm}p{2cm}p{2cm}p{10cm}p{2cm}p{2cm}}

    \hline
    \textbf{CL Strategy} 
    &\textbf{Learning} & \textbf{Application} 
    &\textbf{Description}&
    \textbf{Motivation}&
    \textbf{Limitation}
\\  
\hline
   \multirow{8}{*}{Rehearsal}& \multirow{2}{*}{Supervised}&Steering Control\cite{b39}& Real-time learning is performed to adapt to the unseen road situation on the fly. The highly biased input road data is adapted online by using the divided buffer technique that stores and replays each driving task separately. & Stable \& fast adaptation
   &--
   \\
   \cline{3-6}
   && Controlling vehicle dynamics\cite{b47}& Online adaptation of a neural network-based controller is performed for the most difficult scenarios where both the input and target distributions are non-stationary. & Less computations & Less energy efficient
\\
\cline{2-6}
   &\multirow{6}{*}{Reinforcement}&Vehicle Trajectory Prediction\cite{b50} 
   &The online adaptation of the parameter sharing generative adversarial learning model is performed to adapt the trajectory prediction of a specific driver.& high accuracy&--
   \\ 
   \cline{3-6} 
   &&Uncertainty prediction\cite{b54}& 
   The detection and aggregation of the most useful data at the sub-optimal states are performed through the Monte Carlo dropout that estimates the uncertainty of the output control.
   &Efficient training data sampling &--
   \\ 
   \cline{3-6}
   &&Path Tracking Controller\cite{b48}& 
   The proximal Policy RL optimization algorithm is used to train the neural network that tunes the adaptive weight of the controller for smooth and efficient driving control.
   &fast convergence&--
   \\ 
   \cline{3-6}
   &&Learn to drive\cite{b41}&A continuous, model-free RL deep learning algorithm is used to learn a full-sized real-world car in real-time. &Fast Adaptation &-- 
   \\ 
   \cline{3-6}
   &&Learn to drive\cite{b43}&A deterministic finite state machine is used to adaptively restrict the action space of the vehicle, it enables to learn policy in realistic scenario. & Collision free learning& Slow learning speed
   \\ 
   \cline{3-6}
   &&Meta Policy\cite{b42}& Using the heirarchical deep RL process, a meta-policy network for novel task is learned by exploiting a mixture of previously learned policies. &quick convergence&--
   \\ 
   \cline{1-6}
   Architectural & 
   Unsupervised&Abnormality detection\cite{b46} & Guassian Process along with kalman filter is used to measure the abnormality based on the trajectory error prediction. The predicted abnormality is learned incrementally by incorporating new banks of the filter. &Deal with sparse inputs observations.&--
   
   \\ 
   \hline
\end{tabular}
}
\end{minipage}
\end{center}
\end{sidewaystable}

\begin{sidewaystable}
\sidewaystablefn%
\begin{center}
\begin{minipage}{\textheight}
\begin{flushleft}
\textcolor{black}
{\footnotesize\textbf{Table \ref{my-label_s_1} Continued.}} 
\end{flushleft}
  \label{my-label_s_2}
  \centering
  \renewcommand\arraystretch{0}
  \vskip 0.2 cm

  \resizebox{1\linewidth}{!}{
  \begin{tabular}{p{2cm}p{2cm}p{2cm}p{10cm}p{2cm}p{2cm}}

    \hline
    \textbf{CL Strategy} 
    &\textbf{Learning} & \textbf{Application} 
    &\textbf{Description}&
    \textbf{Motivation}&
    \textbf{Limitation}
    \\ 
    \hline
    \multirow{3}{*}{Architectural}& \multirow{2}{*}{Unsupervised}&
    Abnormality detection\cite{b45} &
   DBN predictive model is used to measure the abnormality based on the trajectory error prediction.
   The incremental learning is performed through free energy minimization that controls the complexity of the predictive model while attaining high accuracy. & Low increase in network complexity&--\\
   \cline{3-6}
    &&Padestrians trajectory prediction\cite{b51}&
    Similaririty based incremental learning is performed to predict and
learn the pedestrian trajectory.The base model is expanded by computing pair wise similarity between pre-trained and newly acquired trajectory results. & Less increase in model complexity&-- 
    \\ \cline{2-6}
    &
    Semi-supervised
    &Classifica- tion\cite{b44}& 
    The  human understandable fuzzy rules are utilized to deal with non-stationarity of input data distribution. 
    & Fast adaptation& Large memory 
    \\ \cline{1-6}
    \multirow{4}{*}{Naive}&
    Supervised \& Reinforcement &
    Lane Following\cite{b40}&
    The online visual perception-action multi-modal learning is performed to adapt the new road condition in real-time.
    &Fast adaptation&--
    \\ \cline{2-6}
    &
    Self-Supervised
    &Connected Vehicle\cite{b52}   &
    V2V communication are used to aggregate the labelled online training data about unexpected events. Active learning based approach selects the most accurate and reliable data from connected vehicles. 
    & High quality  online training data& Large network load 
    \\ \cline{2-6}
    &
    Supervised
    &Vehicle trajectory prediction\cite{b49} &
    A particle-ﬁlter-based parameter online adaptation algorithm is used to adapt the pre-trained policy network toward the predicted target vehicle online.
    & High accuracy
    &--
    \\ \cline{2-6}
    &
    Supervised
    & Driver behaviour monitoring\cite{b53}&
    A cloud-based continuous learning is performed on the pre-trained driving monitoring model that learns a driver's behaviour while the vehicle's control is transferring between the driver and the self-driving AI.
    &Computation \& energy efficient system.&--
    \\ \cline{1-6}
    Offline Learning &
    Supervised
    & Platooning\cite{b55}&
    This method collects the additional training examples from situations where the autonomous vehicle has made mistakes, this method reduces the number of training examples needed to define robust decision boundaries, and is trained offline to solve the problem of catastrophic forgetting.
    & High accuracy. 
    & High computations
\\  \cline{1-6}

\end{tabular}
}
\end{minipage}
\end{center}
\end{sidewaystable}

\subsubsection{Lane keeping}
The visual perception-based lane-keeping system is mostly employed for autonomous vehicles, as they have a better ability to understand the environment like humans. The input of such a system is road images acquired from the in-vehicle camera, while the output is the steering control value. The online learning of the visual lane-keeping system enables the autonomous car to adapt to the diverse road conditions that are not involved during its training phase. To drive onto a new road, a few minutes of real-time learning is required to adapt the features of the unseen road. The different incremental learning methods for visual lane-keeping are discussed below: 
\begin{itemize}
    \item Online Learning of Convolutional Neural Network:
    
 The lane-keeping in autonomous vehicles can be performed by using Convolutional Neural Network (CNN) based-control system~\cite{b39}. The CNN is able to learn the unseen road condition when the vehicle is driven by the human driver.  The steps involved during its online adaptation are shown in Fig.~\ref{fig0}. During the training phase, the CNN weights are updated through back-propagation by minimizing loss function between the predicted and desired steering control values.   

This method addresses two common problems that arise during online learning. The first problem occurs when the model is modified with a small amount of training data and is affected by noise. The second issue arises when the same type of training data that comes in succession is used to update the model, causing bias towards one kind of data.

    The effect of noise is resolved by using mini-batch learning, Adam optimizer, which is a weight-updating algorithm. The biasness in training data is reduced by using the divided buffer strategy. Instead of storing the most recent data in the buffer, this method stores the data of each maneuver (left turn, right turn, and straight) in a separate buffer, as shown in Fig.~\ref{fig13}. Thus, the divided buffering method avoids the biasness and creates a stable, fast, and efficient real-time learning system.
        \begin{figure}[ht!]
    \centering
    \includegraphics[width=0.7\linewidth]{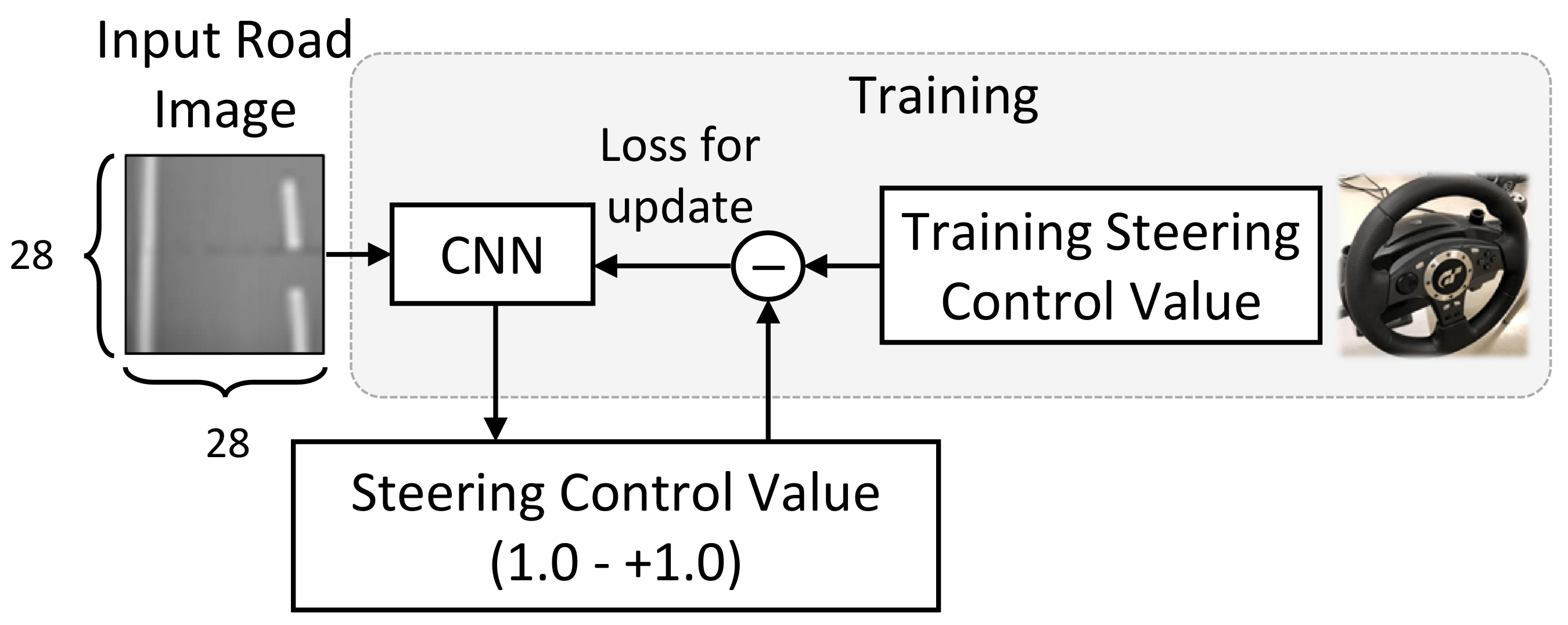}
    \caption{Online learning of CNN based steering control system~\cite{b39}}
    \label{fig0}
    \end{figure}

    \begin{figure}[ht!]
    \centering
    \includegraphics[width=0.7\linewidth]{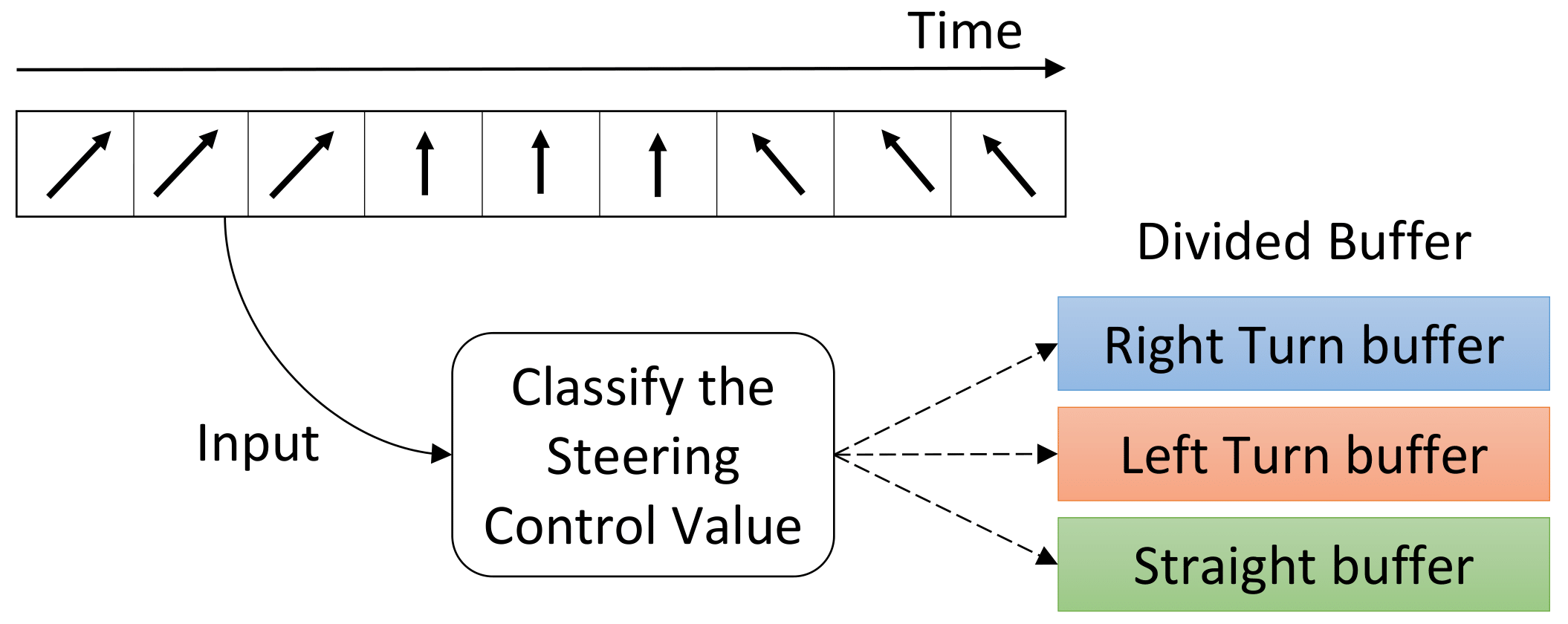}
    \caption{Divided buffer strategy~\cite{b39}}
    \label{fig13}
    \end{figure}

    \item Symbiotic learning
    
In Symbiotic Learning~\cite{b40}, the hebbian learning is extended for multi-modal channel modeling technique to perform action-perception of the lane following system, as shown in Fig.~\ref{fig14}. This method proposes a unique approach to real-time learning that uses supervised learning (by demonstration), instantaneous reinforcement learning, and unsupervised learning (self-reinforcement learning) interchangeably.

During its real-time adaptation, the vehicle is initially learned through demonstration provided by the human driver. After a few minutes of training, the vehicle switches to an autonomous mode, where it further improves the performance by using internal (predictive coding) and external (human reward) feedback. Hence, this method makes use of a visual constancy assumption about the road to allow the system to generate performance feedback on its own, allowing for self-reinforcement learning. 

    
    
    \begin{figure}[ht!]
    \centering
    \includegraphics[width=0.7\linewidth]{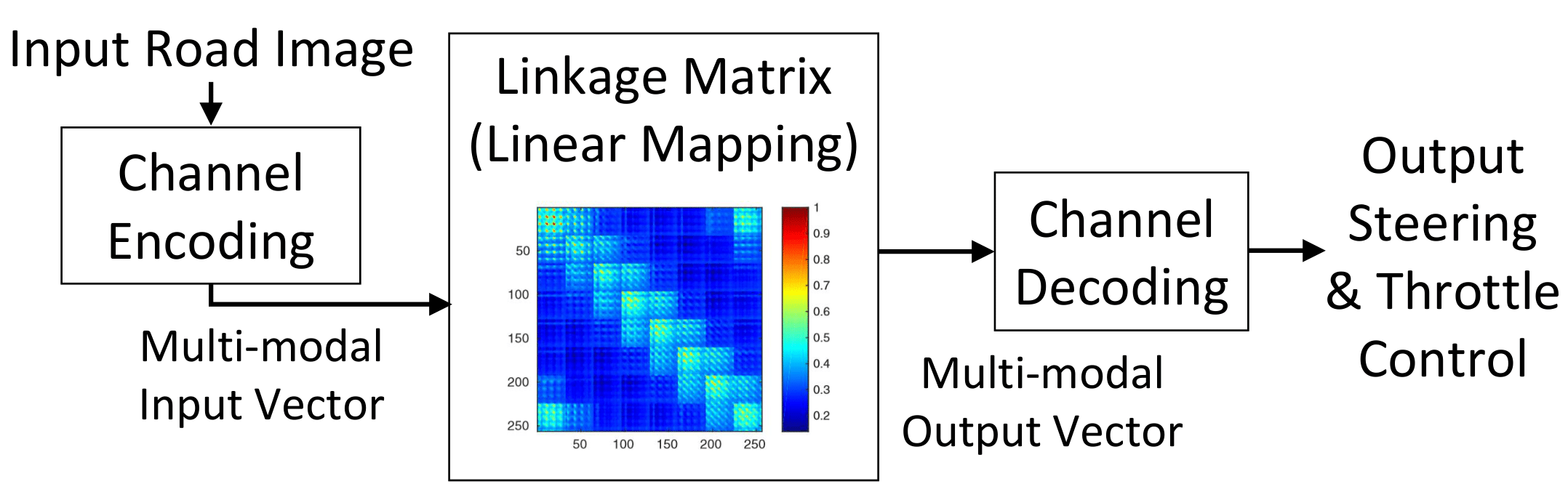}
    \caption{Symbiotic learning~\cite{b40}: The multi-modal channel modelling architecture that maps the road image to the throttle and steering control value. The multi-modal input vector is mapped to multi-modal output vector through linear transformation.}
    \label{fig14}
    \end{figure}
    
    \item Driving as Markov Decision Process:
    
    The continuous learning of lane following on non-stationary data can be performed by using the model-free deep reinforcement learning algorithm~\cite{b41}, which enables the vehicle to explore all the possible states and learn the human level autonomy. 
    
    This method formulates driving as an markov decision problem in which the reward function is defined as the distance traveled without infractions. The deep deterministic policy gradient (DDPG) is used to learn the policy network for lane following. The network is composed of actor-critic pairs, as shown in Fig.~\ref{fig15}, with the critic estimating the cumulative reward function for each state-action pair, while the actor estimating the optimal policy. This method performs random exploration and learns the policy network with state-action pairs that attain maximum cumulative reward without any infractions. 
    
In this method, all explorations and learning are carried out on a full-sized vehicle. With less computational complexity, the vehicle can learn the continuous state-action space in real-time setting.
    
    \begin{figure}[ht!]
    \centering
    \includegraphics[width=3.0in]{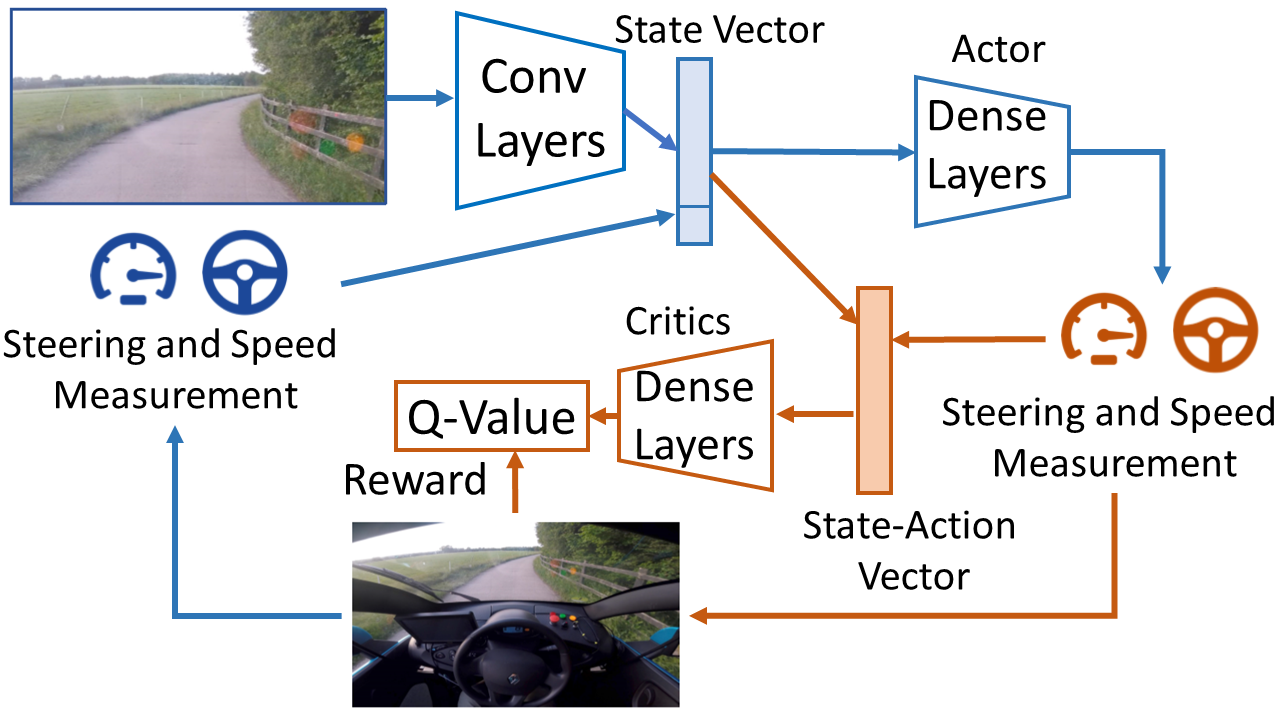}
    \caption{Markov decision process~\cite{b41}: The reinforcement learning strategy of full-sized real-world vehicle.}
    \label{fig15}
    \end{figure}

    \item Meta-policies learning  through hierarchical deep reinforcement learning:
    
    In~\cite{b42}, the control policy of a new task is considered as a shared or meta-policy of previously learned policies.
    
    This method uses hierarchical deep reinforcement learning to learn the meta-policy network for new tasks by using previously learned policy networks. The meta-policy can be modeled through the multi-layer perceptron (MLP) when the MDP is fully observable and no history is involved. While in the case of partially observable MDP, it employs RNN to model the effect of temporal states.
    
    This method consumes less exploration, and the learning of the meta-policy network will converge within a few iterations and attain high reward value as compared to policy networks learned from other conventional RL methods. 

    \item Deep reinforcement Learning(DRL) and Deep Finite State Machine (DFSM):

For a realistic environment, the convergence of reinforcement learning in autonomous driving is extremely difficult. To address this issue, a method combining deep RL and a DFSM~\cite{b43} is proposed, which adaptively restricts the agent's action space based on its current driving situation.
    
The design architecture, as shown in  Fig.~\ref{fig16}, uses the traditional Deep RL algorithm in which the state-action pair is fed to the neural network that estimates their corresponding values, and the maximizer is used to extract the best action with high state-action value. The DFSM is incorporated by a navigator that restricts the state-action pair. Thus, the state-action value is not calculated for all the set of actions, which reduces the computations and accelerates the training process.
    
    \begin{figure}[ht!]
    \centering
    \includegraphics[width=3.0in]{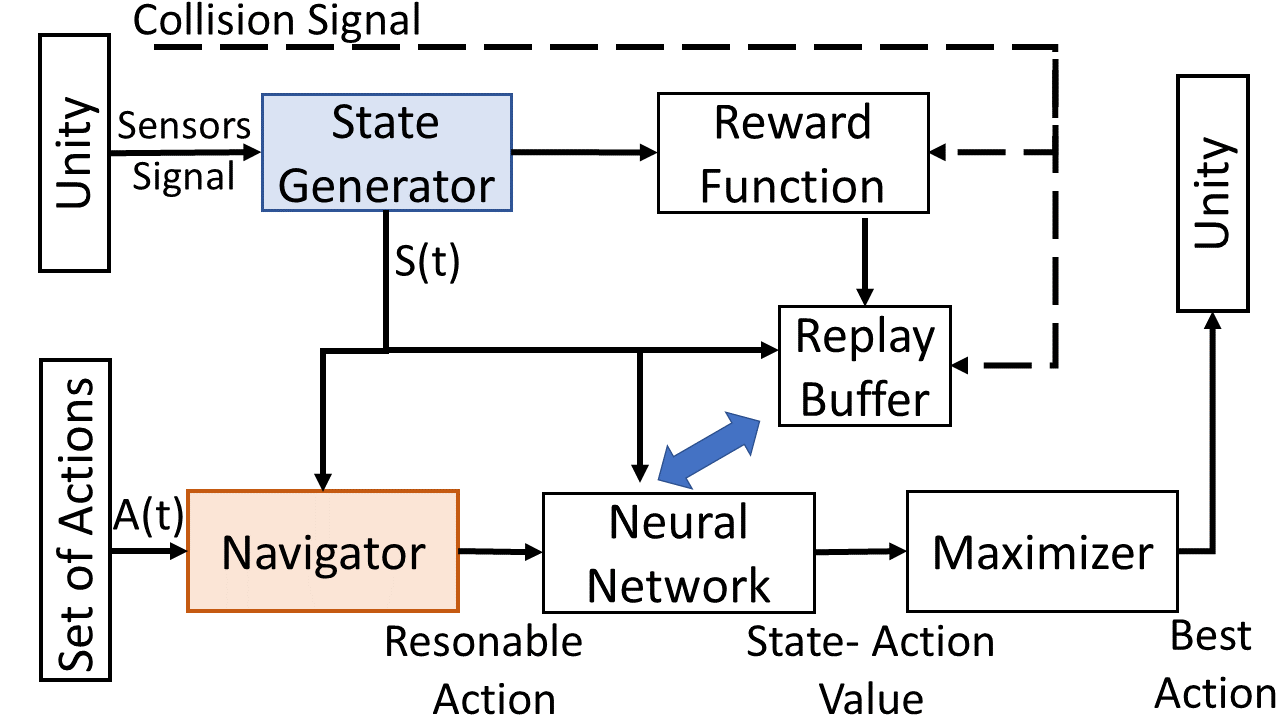}
    \caption{DRL and DFSM~\cite{b43}: The data flow diagram between the driving environment and learning modules.}
    \label{fig16}
    \end{figure}
    \end{itemize}
    
\subsubsection{Scene Recognition}
The classification of unseen scenes is highly critical for autonomous driving, as it requires a highly accurate and reliable system that guaranteed safety under diverse non-stationary settings. To address these challenges, interpretable and explainable incremental learning algorithms are developed that allow the detection and learning of unseen scenes without the need for external supervision.
\begin{itemize}

\item Deep Rule-based classification: 

Deep Rule-based (DRB)~\cite{b44} is a semi-supervised non-parametric approach based on human-understandable (IF..THEN) fuzzy rules, as shown in Fig.~\ref{fig17}. These rules are self-updated and self-evolved  to deal with the non-stationarity of input data. DRB is evaluated for highly challenging classification tasks of different lighting conditions. 

The architecture of DRB is composed of a feature extractor (pre-trained CNN) and a fuzzy rule-based (FBR) layer. The FBR layer contains fuzzy rule subsystems for each class, which are divided into several prototypes to represent local data density peaks.  The density-based nature of these prototypes allows for a more detailed analysis of input data distribution.

The DRB model uses fewer computation resources and much faster convergence speed as compared to conventional deep neural networks. It has high tolerance toward incorrect pseudo-labels of semi-supervised learning due to its density-based nature. 

\begin{figure}[ht!]
    \centering
    \includegraphics[width=3.0in]{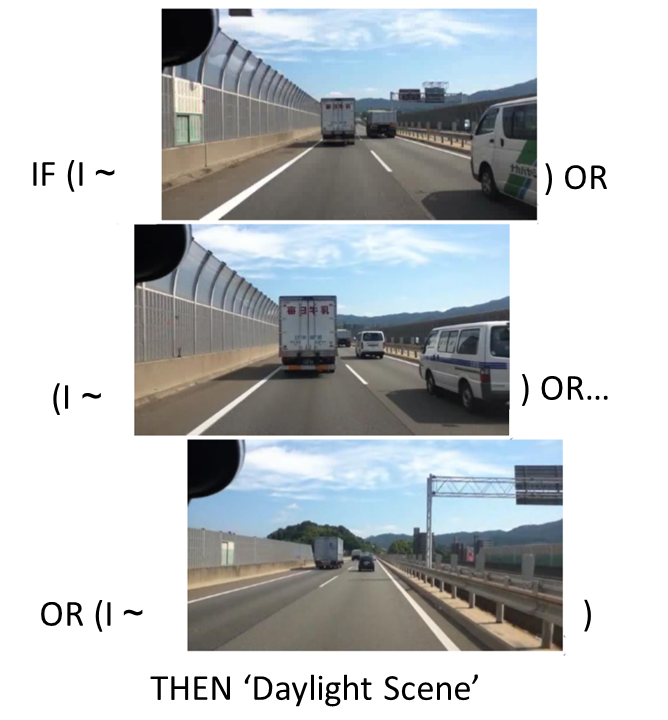}
    \caption{Human understandable fuzzy rule (Example)~\cite{b44}}
    \label{fig17}
\end{figure}
\end{itemize}
\subsubsection{Abnormality detection}
The self-awareness capability enables the autonomous agent to detect its states and impact on the surrounding environment. These systems can be realized by training a predictive model through its own multi-sensory data. As the real-world data is highly non-stationary, dynamic models must be able to detect and incrementally learn the novel states.

For this purpose, the Dynamic Bayesian Networks (DBN) are mostly considered to model these abnormalities. The complexity control of these models during each incremental step is the major research parameter, few DBN-based incremental algorithms are developed to address this issue.  

\begin{itemize}
 \item Free Energy Minimization:

Free energy minimization~\cite{b45} is a process of minimizing the error between the sensor's input and the predicted output of the prediction model. The advantage of energy-based optimization is to keep the previous knowledge at a low cost of model complexity. 

This method employs the clustering-based approach GNG which splits the input data into multiple clusters. Each cluster corresponds to a dynamical model that predicts future observations of the sensory data. Based on the prediction error, it detects the abnormality and the new DBN is created to incorporate the detected novelty. The overall learning process is depicted in Fig.~\ref{fig18}.

This method was evaluated on the ICAB benchmark dataset shown in Fig.~\ref{fig19}, in which three different tasks were performed, i.e., perimeter monitoring (PM), PM under obstacle avoidance, and PM during u-turn. The first task was used to train the initial predictive model while the anomalies were tested for the former two tasks.


\begin{figure}[ht!]
    \centering
    \includegraphics[width=0.7\linewidth]{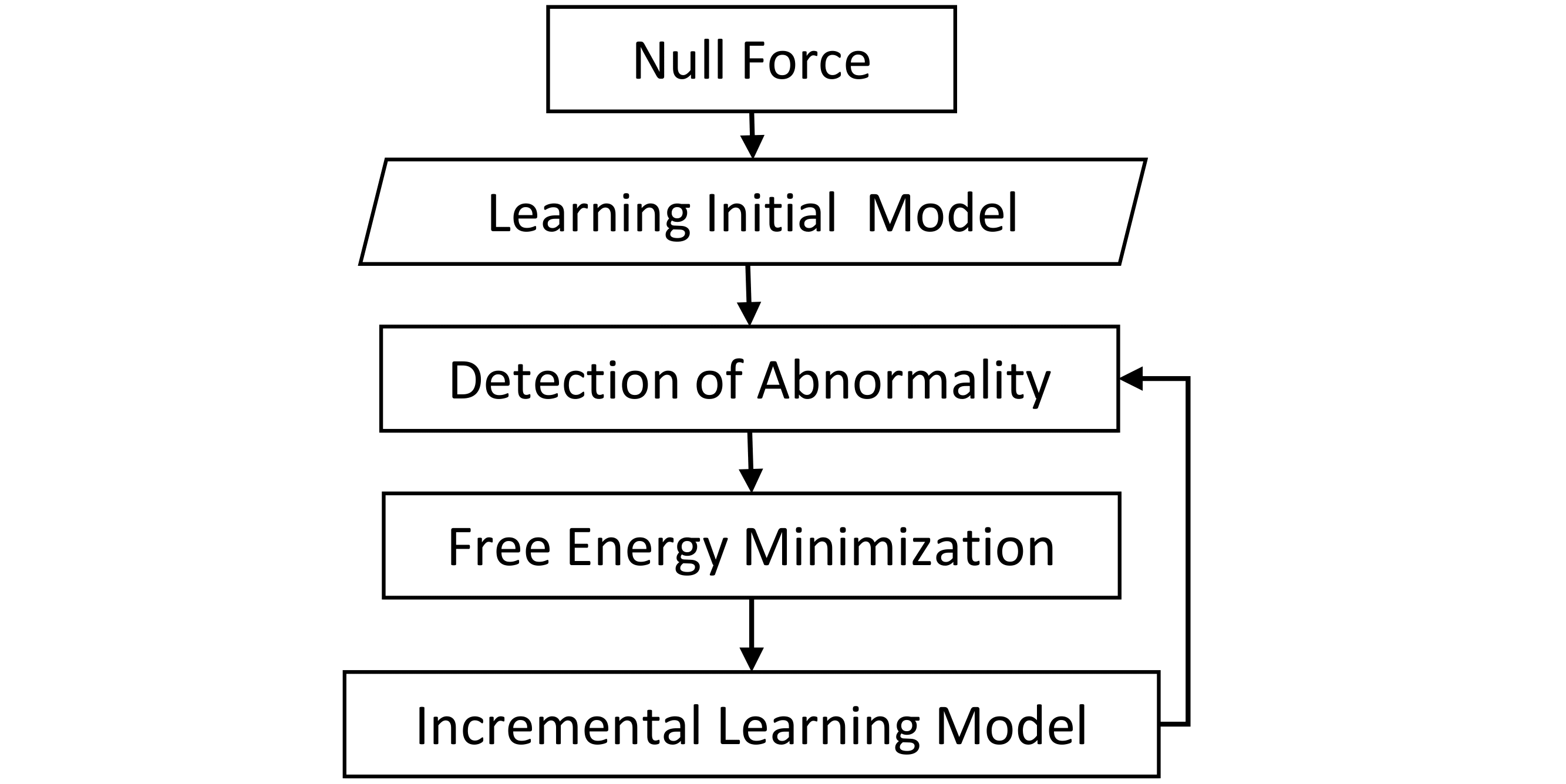}
    \caption{Free energy minimization~\cite{b46}: The adaptive incremental learning process to detect and learn the abnormality.}
    \label{fig18}
\end{figure}


\begin{figure}[ht!]
    \centering
    \includegraphics[width=1\linewidth]{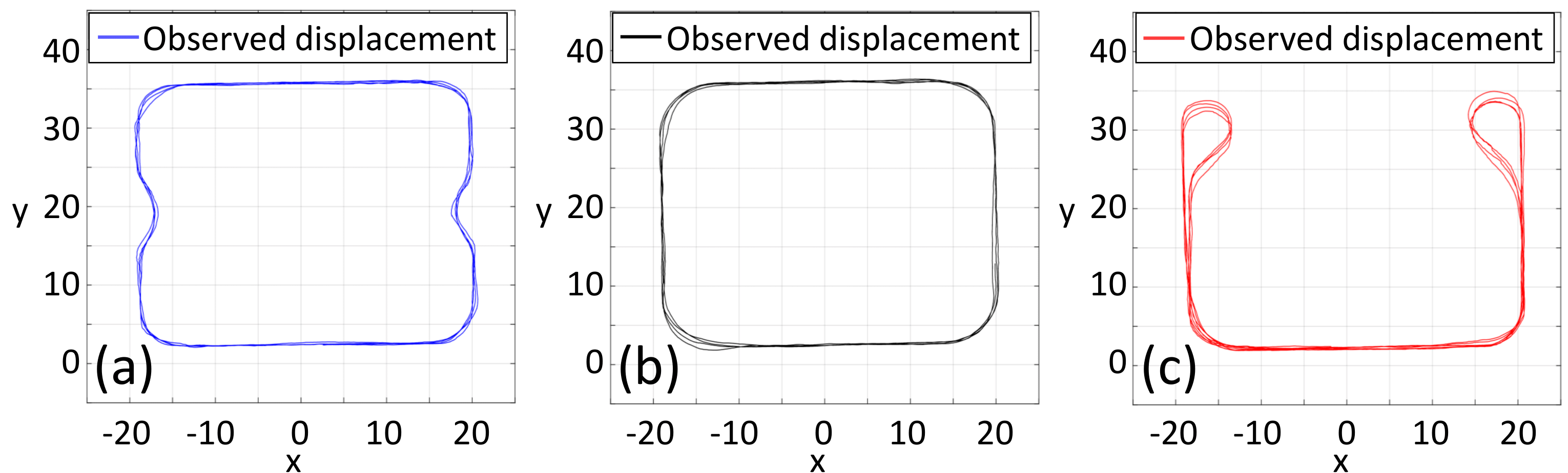}
    \caption{Abnormality detection~\cite{b45}\cite{b46}: ICAB dataset acquired for three different scenarios (a) Perimeter monitoring (b) Perimeter monitoring under obstacle avoidance (c) Perimeter monitoring in presence of u-turn}
    \label{fig19}
    \label{fig:first_sub}
    \label{fig:second_sub}
    \label{fig:third_sub}
\end{figure}

\item Guassian Process:
 
Gaussian process (GP) regression~\cite{b46} allows the approximate estimation of expected vehicle motion over a large area using sparse observed data.

The data flow of the system is briefly shown in Fig.~\ref{fig20}. The input to the system is the position of the autonomous vehicle, which is segmented into a finite spatial zone of motion using the GP regression. 
These spatial zones are modeled by a set of Kalman filters that perform the prediction of future instances. Based on the prediction error, the abnormality is identified and learn incrementally by incorporating new banks of the filter. 

This method is also evaluated on the iCAB dataset shown in Fig.~\ref{fig19} that performed the perimeter measurement for the vehicle-pedestrian scenario.

\begin{figure}[ht!]
    \centering
    \includegraphics[width=3.5in]{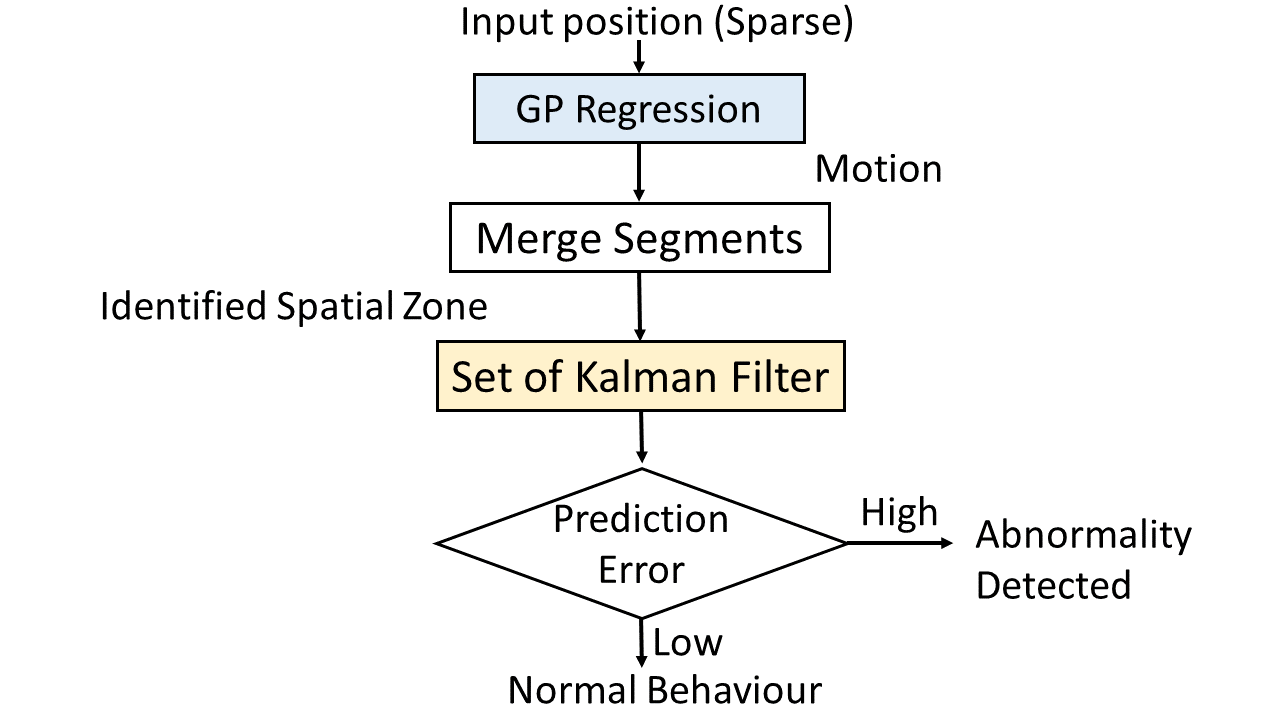}
    \caption{The dataflow of abnormality measurement by incorporating Guassian Process with Bayesian network~\cite{b46}.}
    \label{fig20}
\end{figure}
\end{itemize}

\subsubsection{Controllers}
Vehicle controllers widely use the neural network model to map extremely non-linear vehicle dynamics. The change in dynamics usually occurs due to external (road condition) and internal uncertainties (weights, suspension, stiffness). Such neural network-based control systems must be updated online to deal with uncertain conditions.

 
 \begin{itemize}
 \item Locally Weighted Regression Pseudo-Rehearsal for Vehicle Dynamics(LWPR2):

 In LWPR2, neural network modeled vehicle dynamics is used by the MPC control law application~\cite{b47}. In the case of the MPC controller, the input distribution remains constant while the output target distribution changes. The adaptation process involved in this technique is briefly explained in Fig.~\ref{fig21}. For input-output mapping, this method employs the LWPR technique, which is updated online to alter target mapping. During online learning, the pseudo-data is combined with the incoming online data to update the neural network. The pseudo-rehearsal is used to produce the artificial input data points while their pseudo-labels are produced from the LWPR module. 
 
 Consequently, the non-stationarity of input-output mapping is easily handled with LWPR2. This method produces a model that can be safely usable by the MPC controller. Instead of directly uses LWPR in the MPC application that requires high computation per prediction, this method incorporates the LWPR in pseudo-rehearsal technique with a much-reduced computational cost.
    \begin{figure}[ht!]
    \centering
    \includegraphics[width=3.5in]{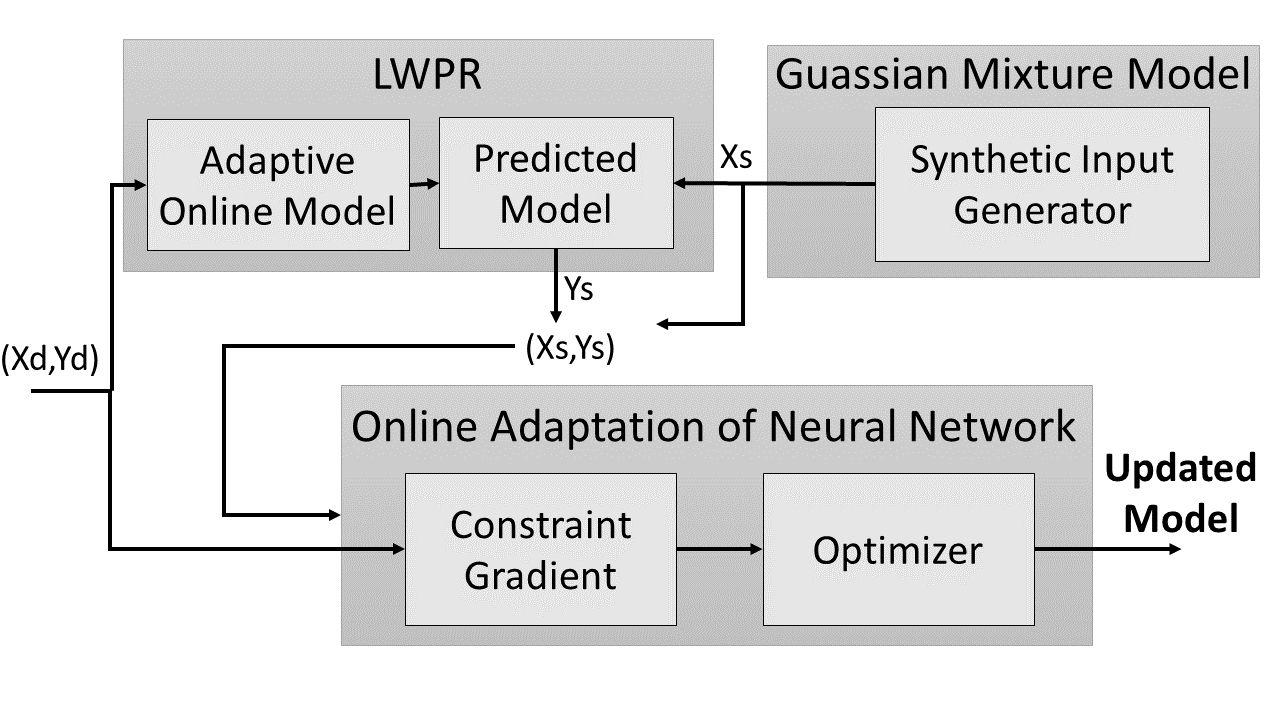}
    \caption{The online adaptation process involved in LWPR2~\cite{b47}. }
    \label{fig21}
    \end{figure}
    \item  Reinforcement Learning Based Adaptive Path Tracking Control System:
    
    In~\cite{b48}, the neural network is used to adaptively select the optimal weights of the conventional controllers, ensuring high smoothness and low path tracking error. 
    
    The reinforcement learning algorithm named PPO ((Proximal Policy Optimization) is used to train the network. This method has performed fast convergence through a well-designed reward function that maintains a balance between tracking error and smoothness. During the learning process, the state and reward transitions are computed and stored in a replay buffer, which is used for online learning of the RL model.
    
    The architecture of the discussed control system is demonstrated in Fig.~\ref{fig22}. First, the data transition module computes the predicted error for finite future time instances. Due to the predictive capability of the RL model, it takes the error as input and computes the optimal weights for path tracking controllers. As a result, the controller with adaptive weights performs better than a pre-defined fixed weight controller.

\begin{figure}[ht!]
    \centering
    \includegraphics[width=3.0in]{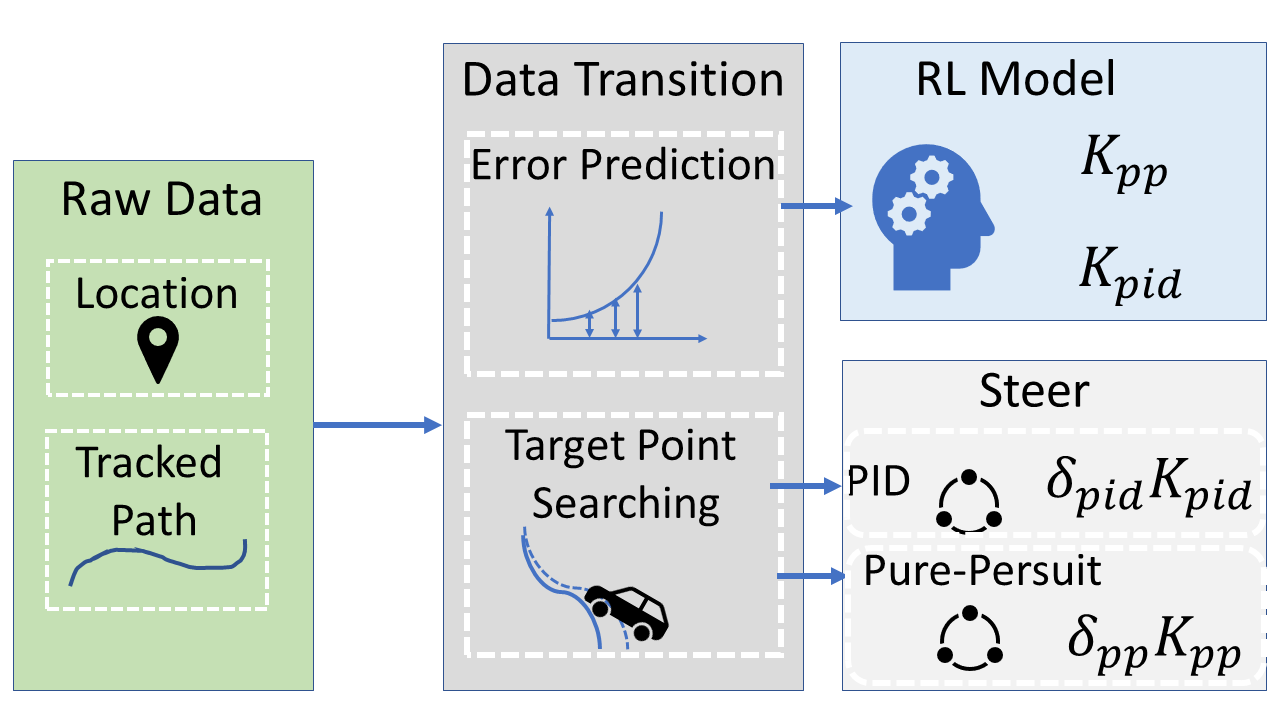}
    \caption{Architecture of adaptive path tracking controller~\cite{b48}.}
    \label{fig22}
\end{figure}
 \end{itemize}
\subsubsection{Vehicle Tracking}
The ability to predict the other vehicle's future trajectory is important for safe and reliable autonomous driving. This prediction is not deterministic because human driving behavior is uncertain. Under such circumstances, the probabilistic distribution of trajectories would be estimated. 

The generalized offlined trained trajectory prediction model is not able to correctly predict the trajectory of a single driver, it needs to be tailored online to a specific expected target.

\begin{itemize}
\item  Bayesian Recurrent Neural Network:

The Bayesian Recurrent Neural Network (BRNN)~\cite{b49} allows a wide range of feasible trajectory distributions. The BRNN model is consist of a policy and a system's dynamic network, as shown in Fig.~\ref{fig23}. The system model is assumed to be known prior while the policy is adopted online. The BRNN is initially trained by a gradient-based black box divergence minimization. The online adaptation is performed through the particle filter-based adaptation technique. 

Hence, this method predicts feasible trajectories over a long time horizon. The adapted BRNN has a high likelihood value of trajectory than the offline trained model.


\begin{figure}[ht!]
    \centering
    \includegraphics[width=0.7\linewidth]{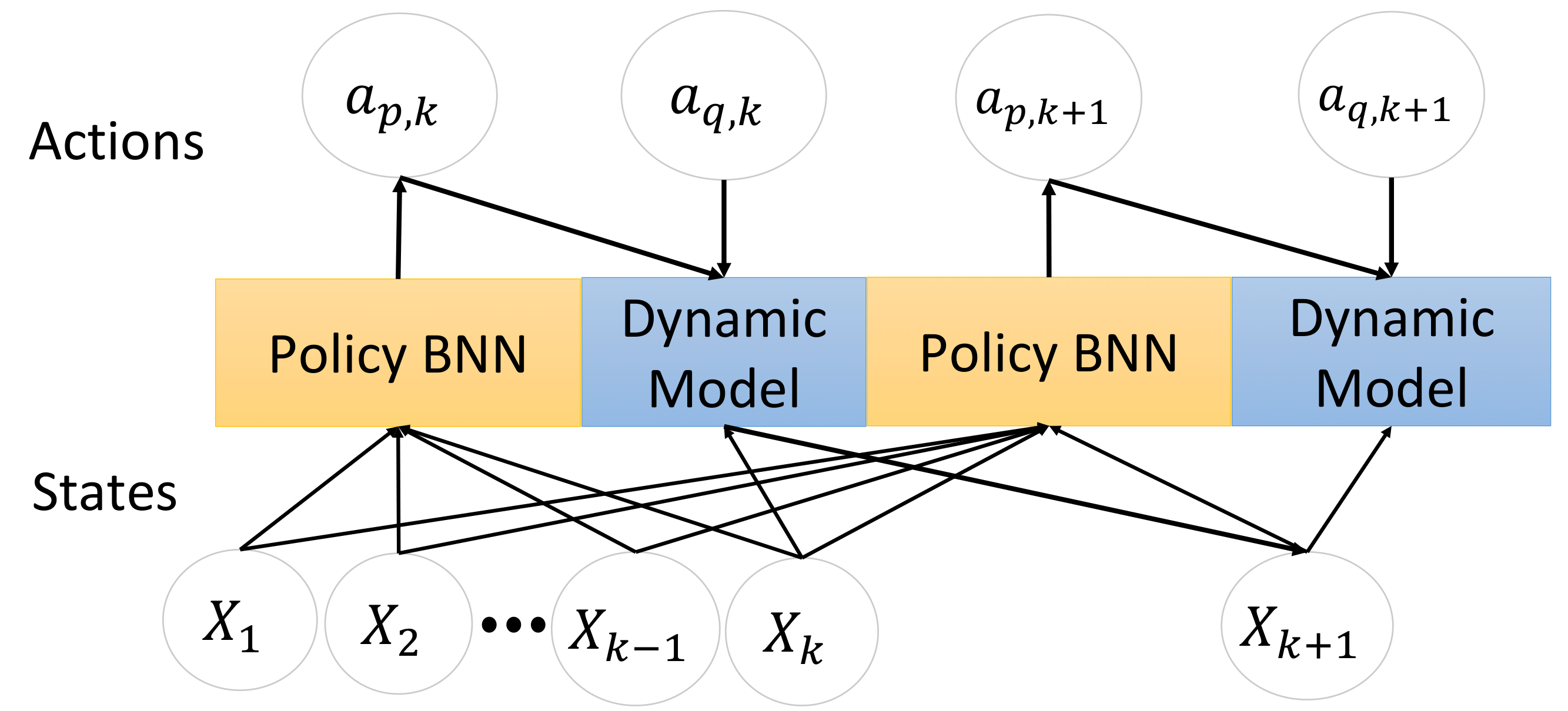}
    \caption{The structure of bayesian recurrent network~\cite{b49}.}
    \label{fig23}
\end{figure}

\begin{figure}[ht!]
    \centering
    \includegraphics[width=0.6\linewidth]{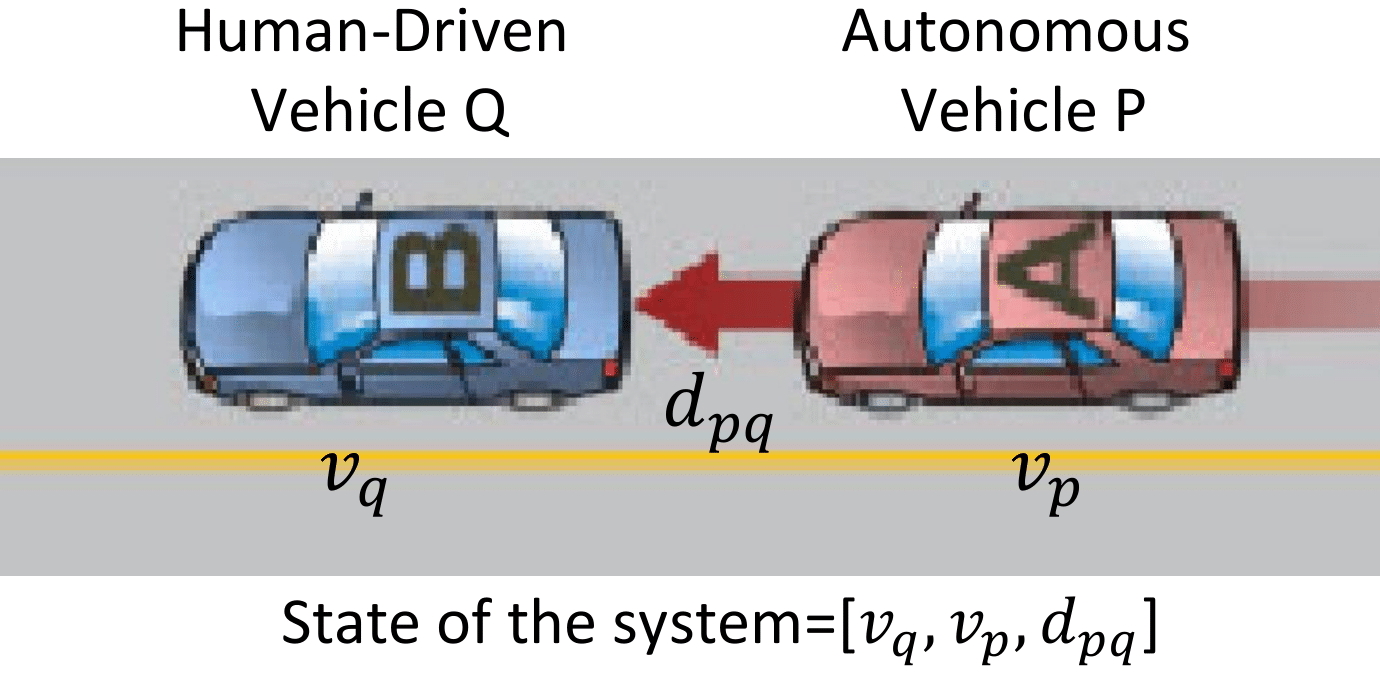}
    \caption{Vehicle Tracking System}
    \label{fig24}
\end{figure}

\item Generative Prediction Network
    
Parameter sharing generative adversarial imitation learning (PS-GAIL)~\cite{b50} learns the interactive model and effectively averages out various driving behaviors.
   
    To adapt the model to a specific driver, the online adaptation of PS-GAIL is performed with a recursive least square parameter adaptation algorithm (RLS-PAA). 
The architecture of PS-Gail is shown in Fig.~\ref{fig25:first}.    
    During its online adaptation, the features extractor is taken from the offline trained model and remained static during adaptation, while the parameters between the last hidden layer and output layer are updated through the recursive least square parameter adaptation algorithm (RLS-PAA). Its one-step learning method uses a linear function, resulting in fewer computations.
    The multi-step prediction through the RNN feedback loop is also demonstrated in Fig.~\ref{fig25:second}. 
    
\begin{figure}[ht!]
\centering
\subfigure[Offline Training of policy network through PS-GAIL]{%
\label{fig25:first}%
\includegraphics[height=1.7in]{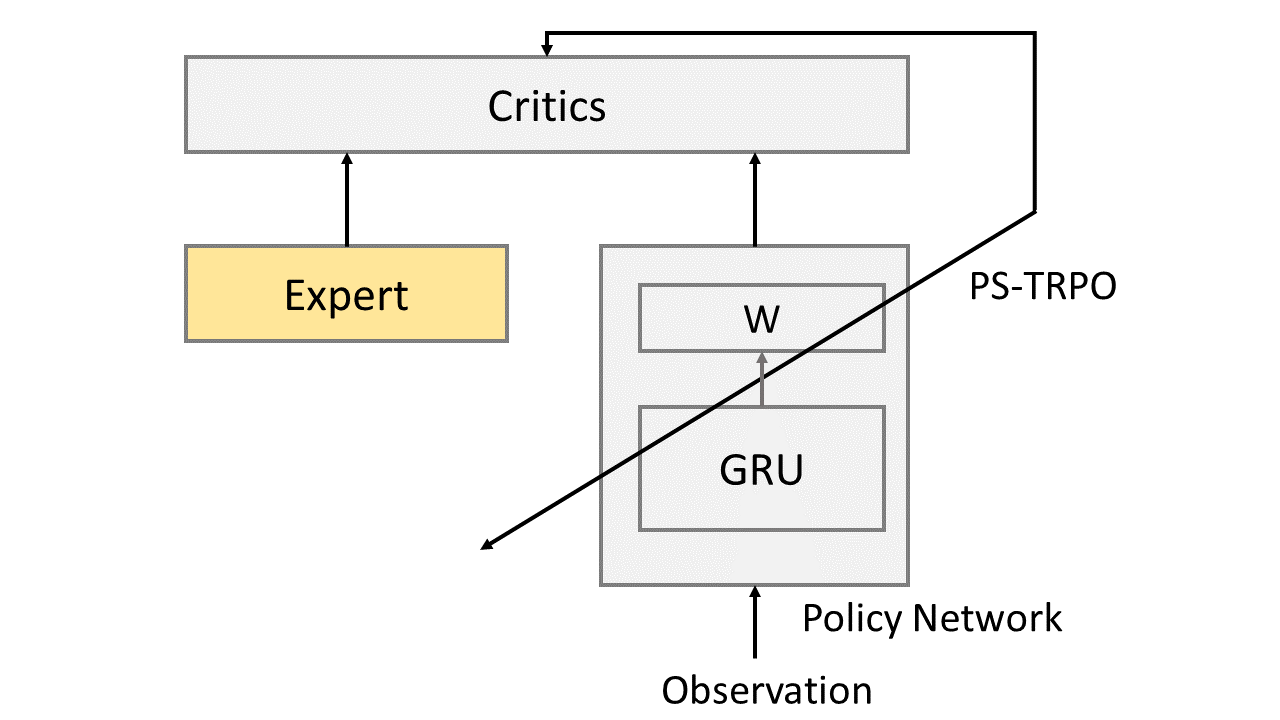}}%
\qquad
\subfigure[Online Learning of the model by using RLS-PAA]{%
\label{fig25:second}%
\includegraphics[height=1.7in]{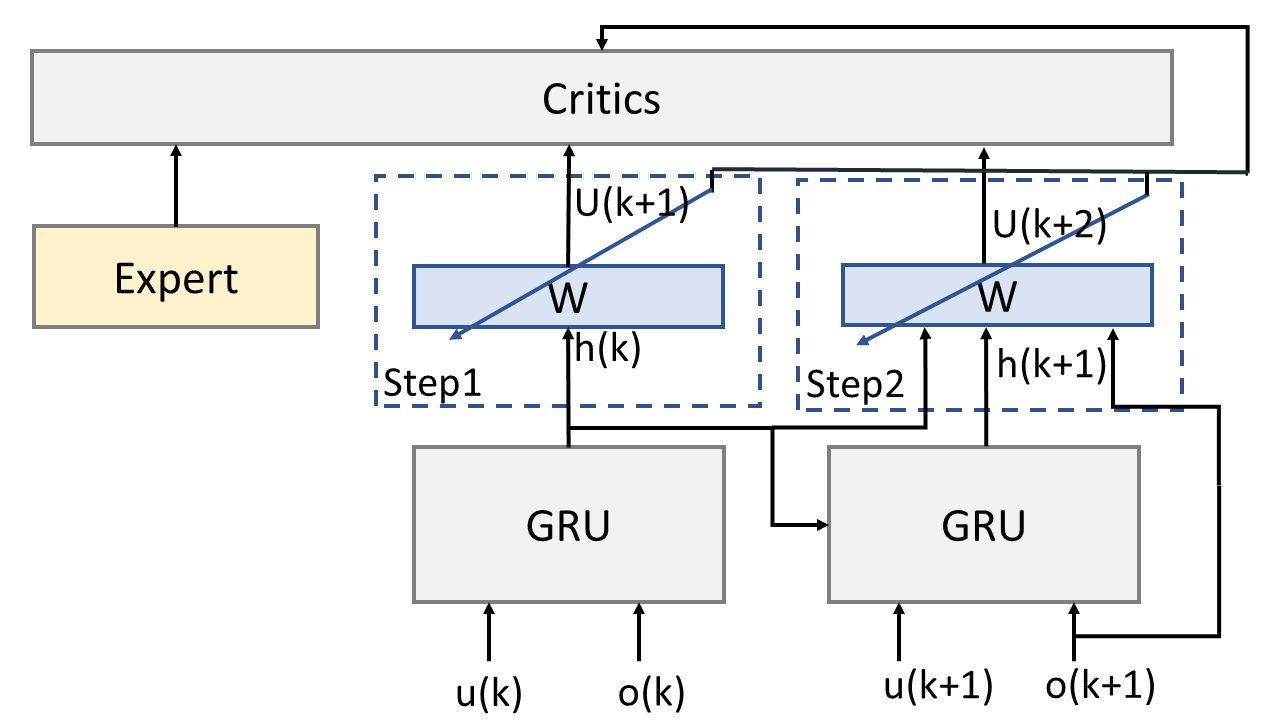}}%
\caption{Generative prediction network~\cite{b50}: The offline training of policy network is performed through PS-GAIL. The online learning is performed by incorporating RLS-PAA with PS-GAIL.The policy network is composed of GRU and fully connected last layer. }
\label{fig25}%
\end{figure}

\end{itemize}
\subsubsection{Pedestrian Tracking}

Autonomous driving in an urban environment necessitates to interact with pedestrians and correctly anticipating their motion. This is highly challenging task because 'rules' are less clear or can be violated. 

For this purpose, the Similarity-Based Incremental Learning Algorithm (SILA)~\cite{b51} is developed that predicts and learns the pedestrian trajectory in urban intersections. This method primarily focused on constraining the size of the model during incremental learning, resulting in a slower model growth rate and less learning time.

\begin{figure}[ht!]
\centering
\includegraphics[width=0.6\linewidth]{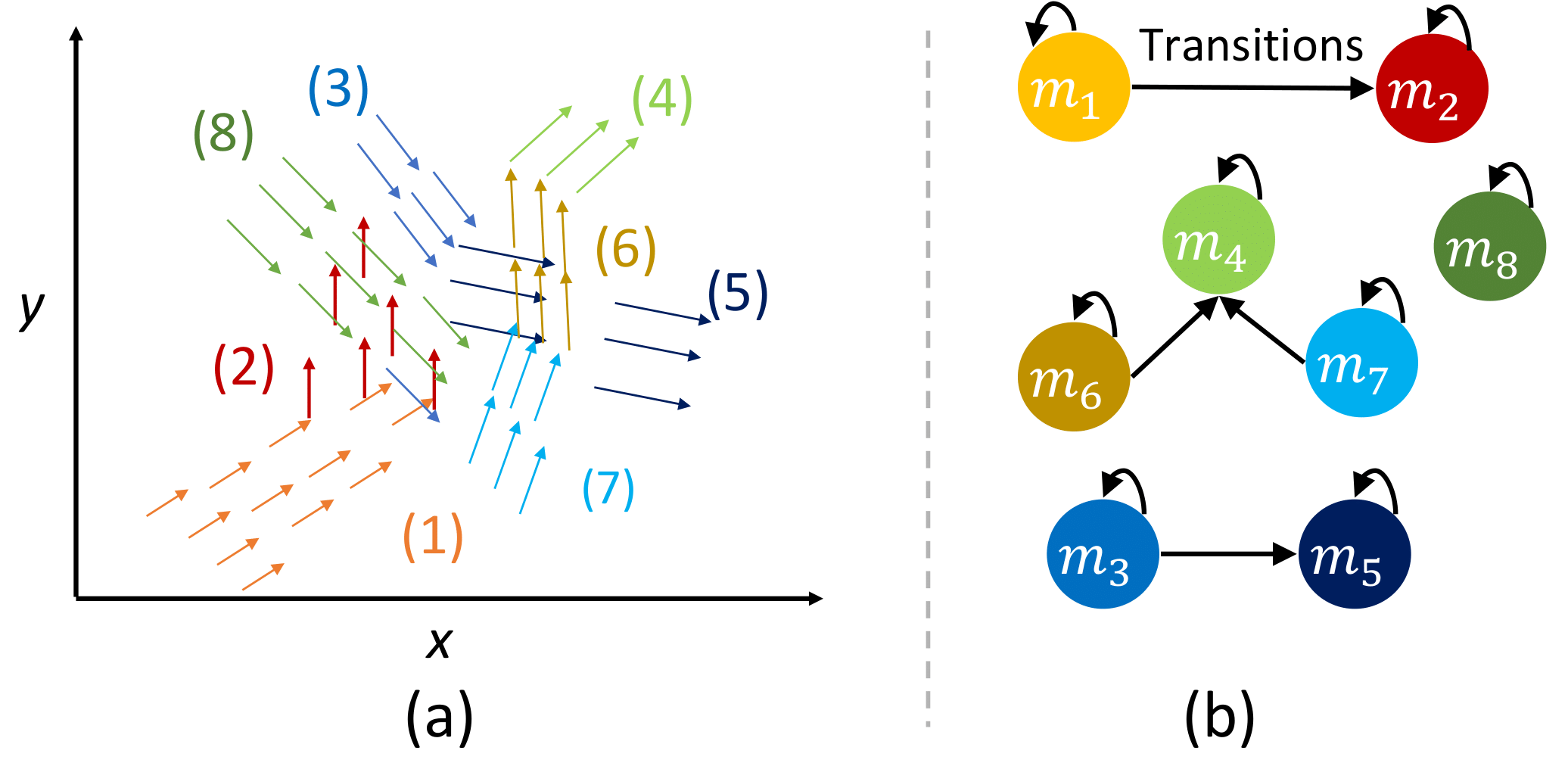}%
\caption{SILA~\cite{b51}(a) Motion Primitives (b) Motion Primitive Graphs}
\label{fig26}
\label{fig26:first}%
\label{fig26:second}%
\end{figure}

In SILA, the trajectories are modeled as motion primitives as shown in Fig.~\ref{fig26:first}. The Gaussian process is used to model the transition between these primitive. The motion primitives and transition are learned from features that are transferable over the environment. In each training cycle, the new motion primitive graph is created that consists of motion primitive and transitions.

\begin{figure}
    \centering
    \includegraphics[width=3.5in]{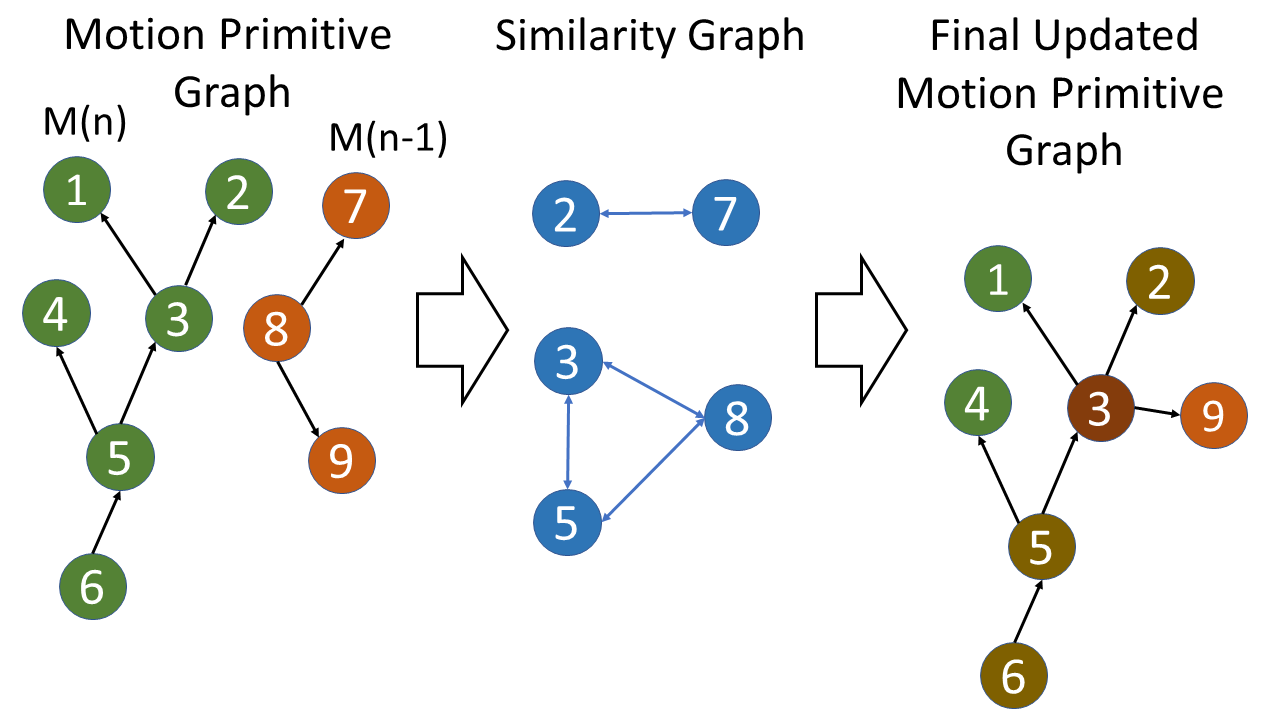}
    \caption{Similarity Based Incremental Learning Algorithm~\cite{b51}.}
    \label{fig27}
\end{figure}

During incremental training, the pair-wise similarity between the pre-trained and new motion primitive graph model is computed. The similar primitives are merged while novel primitives are added to the pre-trained motion primitive graph, illustrated in Fig.~\ref{fig27}. As a result, the size of the model grows at a slow rate.

\subsubsection{Connected Vehicle}

The classification of unexpected events is highly demanding for autonomous driving. In an online learning framework, the labeled data of unknown events can be effectively collected through vehicle-to-vehicle (V2V) communication. An active learning-based method~\cite{b52} is developed to acquire the most diverse labeled training data with high quality from connected vehicles.

\begin{figure}
    \centering
    \includegraphics[width=0.7\linewidth]{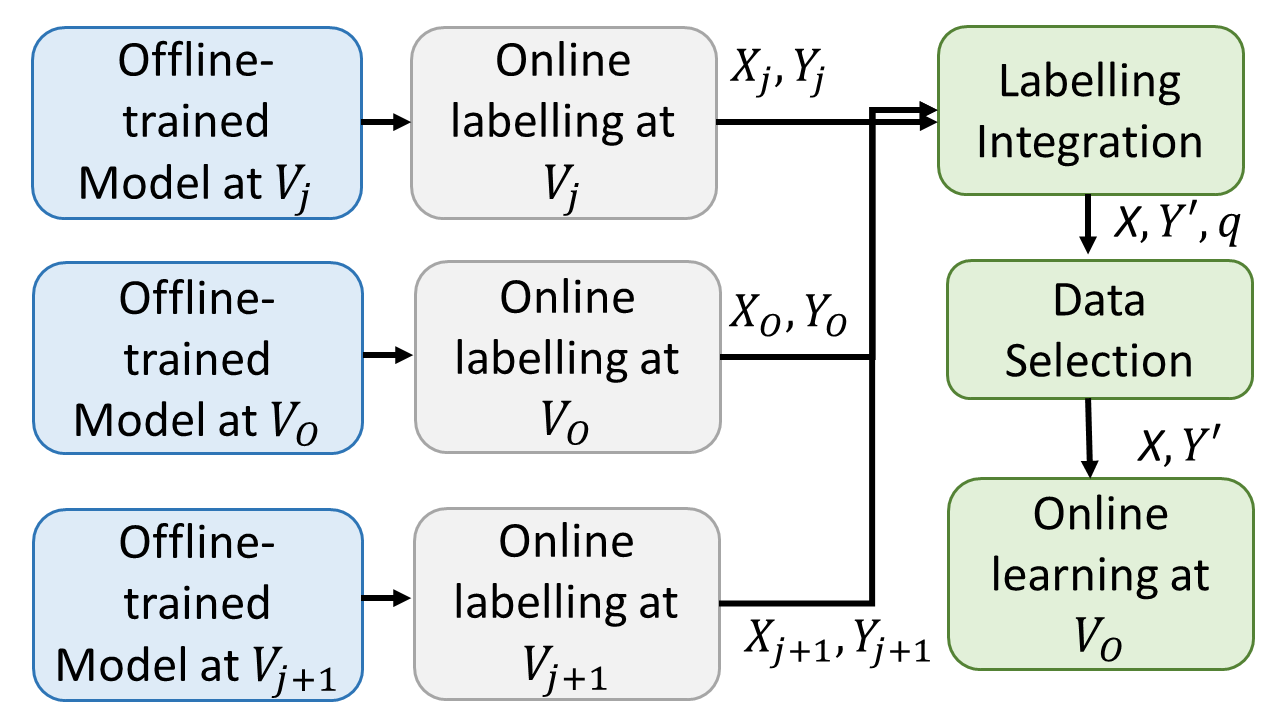}
    \caption{Active learning based framework for V2V communication~\cite{b52}.}
    \label{fig28}
\end{figure}

The data flow of the active learning system is briefly discussed in Fig.~\ref{fig28}. Each vehicle in V2V communication has its own pre-trained model, that generates the online labels. These labels can be integrated into the ego vehicle by using three different methods of label integration which involve majority voting, weighted majority voting, and weighted average.

The V2V communication allows three modes of data exchange i.e. label mode, data mode, and sample mode.  In label mode, it exchanges the label only. In data mode, it only transfers the data, while, in sample mode, it exchanges both data and labels. 
There is a trade-off between accuracy and network load bandwidth. This active learning method achieves high accuracy for sample and data mode, while these modes require high network load. 

Thus, this method provides an effective result in the situation where the ego vehicle produces bad quality signal due to the aging of sensors.

\subsubsection{Driver-monitoring}

In the semi-autonomous vehicle, the driver monitoring system is an essential component as it controls the transfer between the driver and the autonomous system. Under uncertain conditions, the confidence level of the autonomous system is low, the control needs to be hand-over back to the driver. But, before this switching, the system needs to ensure that driver is ready to take control.

The driver monitoring system should have the ability to incrementally learn the anomaly data and also needs to adapt online to a specific driver's behavior. For this purpose, a "Driving Not-Driving detection system "~\cite{b53} is developed that monitors the driver behavior in real-time and adapts to novel scenarios. 

\begin{figure}[ht!]
    \centering
    \includegraphics[width=0.55\linewidth]{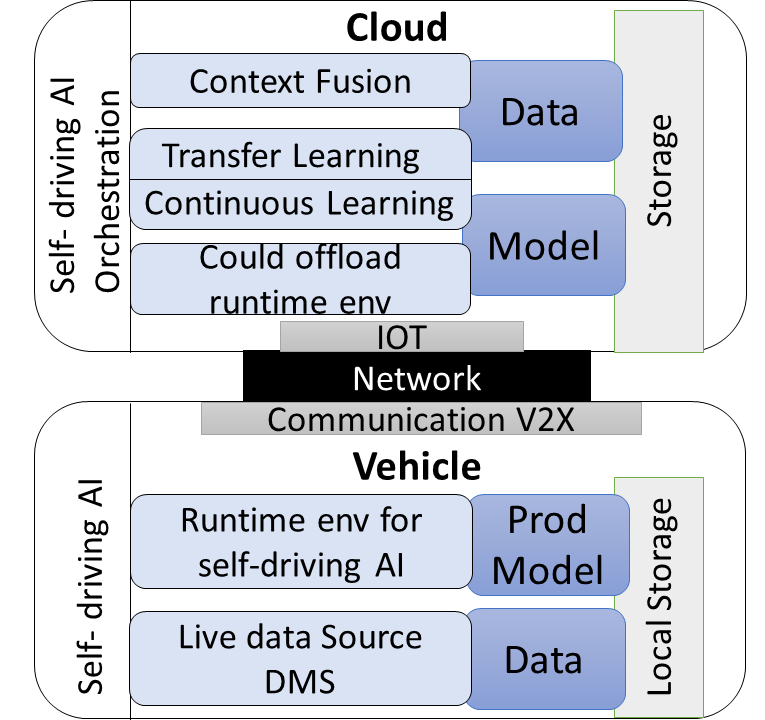}
    \caption{The framework of cloud-based driving monitoring framework. The architecture of self-driving orchestration at cloud side and self-driving AI at vehicle side~\cite{b53}}
    \label{fig29}
\end{figure}

This approach employs the cloud-based framework, shown in Fig.~\ref{fig29}, which consists of two separate modules, i.e., vehicle side and cloud side module, both are connected through the network. The vehicle-side module is composed of local storage that stores the live data temporarily until the data is offloaded to the cloud side. The continuous learning process in the vehicle side triggers the re-training at the cloud side by observing the performance drop due to anomalous data.
Consequently, the cloud-based AI system leverages the performance of incremental learning,  allowing all computational complexity to be offloaded from the vehicle to the cloud.  

\subsubsection{Uncertainity Detection and Data Aggregation}

The uncertain conditions in autonomous vehicles lead to infractions. The data acquired at the uncertain states is highly useful to define robust decision boundaries for deep learning systems.
An autonomous agent must be able to detect sub-optimal states and aggregate the most useful training data. 
\begin{itemize}
    \item  Uncertainty-Aware Data Aggregation for Deep Imitation Learning (UAIL):
    
    UAIL~\cite{b54} gathers training data by estimating the output's uncertainty at a sub-optimal state. Monte Carlo Dropout is used for uncertainty estimation, in which the output distribution is computed using multiple dropout masks at each level, then the statistics of this distribution are used to calculate the uncertainty score.
    When the vehicle enters the sub-optimal state, it switches the control to the driver and collects the unknown training data. 

UAIL acquires the most useful training data and attains a high success rate. 
It predicts the infraction before many time steps and avoids the series of bad actions through human intervention.

\begin{figure}[ht!]
    \centering
    \includegraphics[width=3.5in]{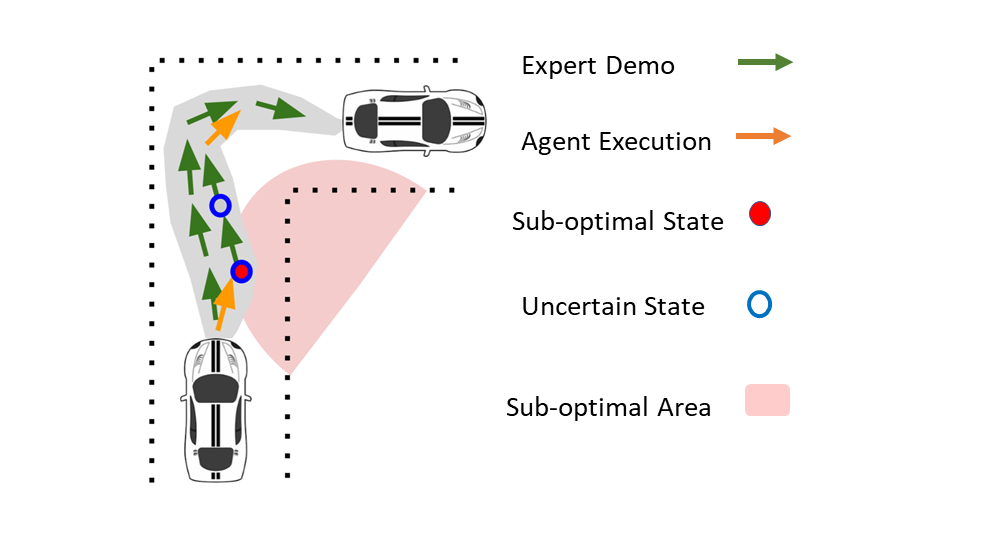}
    \caption{The vehicle avoids the sub-optimal and uncertain zones by UAIL~\cite{b54} }
    \label{fig30}
\end{figure}
\item Correction Based Incremental Learning (CBIL):
 
CBIL~\cite{b55} enables the autonomous vehicle to adapt to the novel scenario by using the least amount of training data. This algorithm gathers the most useful novel data and creates an accurate deep learning model than conventional supervised learning. 

When the autonomous vehicle encounters an unseen condition, the CBIL collects the most uncertain hard training examples (where it commits the mistakes) in an online setting. To avoid catastrophic forgetting, all of the collected data is subjected to offline training. This method can be preferably employed in an autonomous vehicle development process where the online data aggregation and testing are performed in fleed-wide periodic schedules. In practice, the autonomous vehicle is mostly trained for the data acquired under normal conditions. It does not cover the adverse driving scenarios like sudden emergency stops and drifting the road off. For such situations, CBIL is used to gather these hard examples through the safety driver intervention.

Hence, this method greatly improves MTTF (mean time to failure) because it concentrates on examples that are highly sensitive to the decision boundary. It is more scalable than supervised learning as it uses a very less amount of training data.
\end{itemize}

\subsection{Unmanned Aerial Vehicles}
UAVs have been widely used in many industrial applications because of their high speed and portability. 
Deep learning-based control systems are mostly used in UAVs due to their high speed and generalization performance. However, the off-line trained DNN cannot generalize well to the unseen conditions since it has been trained for limited training data. To operate the UAV in an uncertain environment, online adaptation is required.


The CL-based approaches used in UAV applications are discussed in this section. A brief description of these approaches is also explained in Table.~\ref{my-label_u}.

\begin{sidewaystable}
\sidewaystablefn%
\begin{center}
\begin{minipage}{\textheight}
 \caption{Summary Table of Real-time Continual learning applications in UAVs}
  \label{my-label_u}
  \renewcommand\arraystretch{0}
  \resizebox{1\linewidth}{!}{
  \begin{tabular}{p{2cm}p{2cm}p{2cm}p{10cm}p{2cm}p{2cm}}

    \hline
    \textbf{CL Strategy}  
    &\textbf{Learning} & \textbf{Application} 
    &\textbf{Description}&
    \textbf{Motivation}&
    \textbf{Limitation}\\
    \hline
    \multirow{3}{*}{Naive} & Supervised& Trajectory tracking\cite{b57} & The online learning of a pre-trained DNN-based controller is performed to learn unseen trajectories in real-time. 
    & Fast adaptation & High computation\\ 
    \cline{3-6}
    &&Trajectory tracking under varied dynamics\cite{b58} &The online learning of a pre-trained deep fuzzy neural network-based controller improves the control of non-linear system under diverse and varied operating conditions (i.e., different payloads, height, and speed).&Fast adaptation&High computation\\ 
    \cline{2-6}  
    &Self-supervised&Object tracking\cite{b59}&The online fine-tunning of a CNN-based tracking system is performed to adapt the changed appearance of the tracked target.&Gather useful training data &--\\
    \cline{1-6}
    Architectural& Supervised & System Identific- ation\cite{b60}& The online learning of recurrent fuzzy neural network is performed to deal with the uncertainty of real-time time series. 
    &Fast adaptation&--\\ 
    \cline{1-6}
    Rehearsal&Reinforcement&Search and Rescue\cite{b61}& A variants of Deep Q-learning algorithm with a double state input data (the raw image and a map containing positional information) is used to continuously improve the understanding of a UAV while exploring a partially observable environment.
    & Low computational and memory cost&--\\ 
    \cline{1-6}

    \end{tabular}}
\end{minipage}
\end{center}
\end{sidewaystable}

\subsubsection{Unseen Trajectory Tracking}
\begin{figure}%
\centering
\subfigure[]{%
\label{fig32:first}%
\includegraphics[height=1.2in]{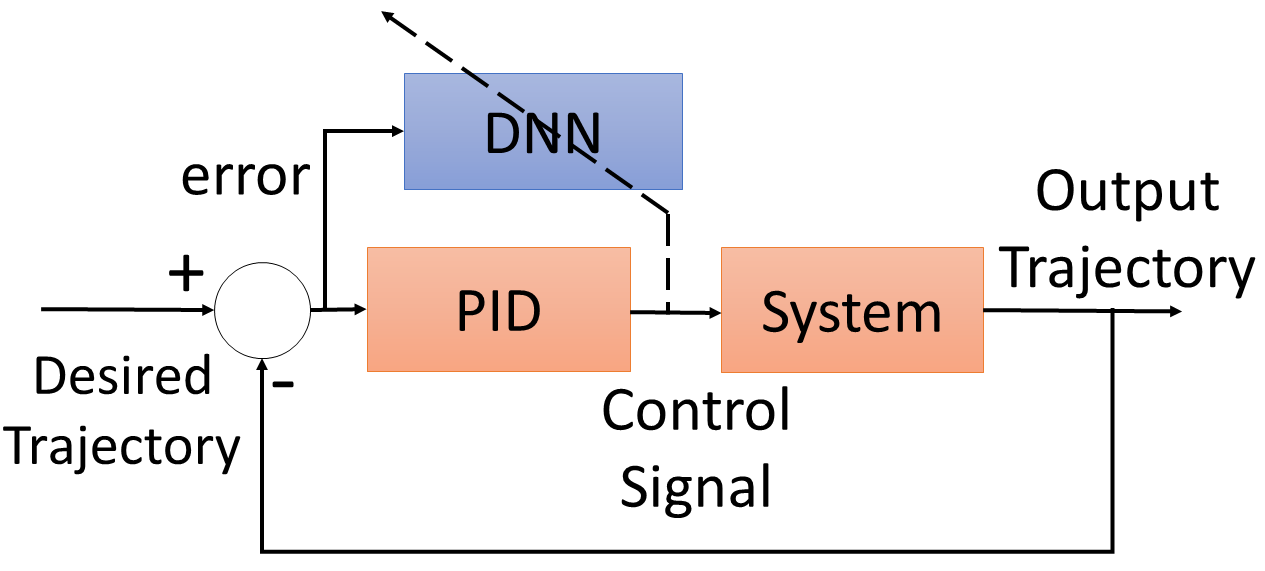}}%
\qquad
\subfigure[]{%
\label{fig32:second}%
\includegraphics[height=1.2in]{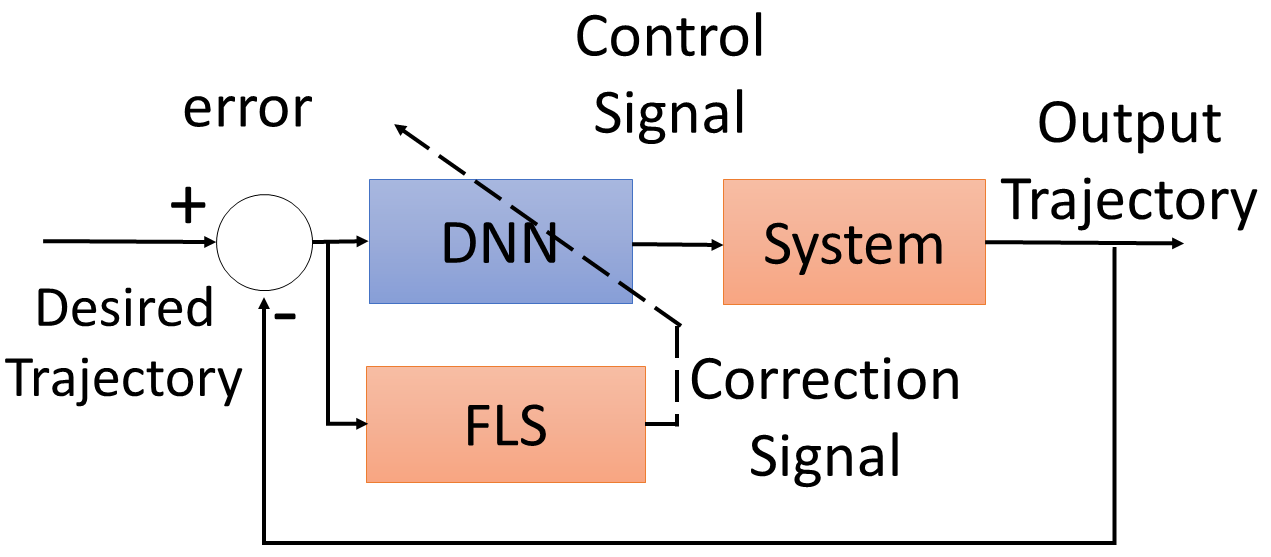}}%
\caption{(a) Offline training of DNN based controller by PID controller (b) Online training of DNN by FLS~\cite{b57}}
\label{fig32}%
\end{figure}

In trajectory tracking, the online learning of the DNN-based controller enables UAVs to deal with unknown trajectories~\cite{b57}. The DNN-controller maps the error signals between the desired and observed trajectory to the control output of the system. It was initially trained offline on the data acquired from the PID controller for a limited set of trajectories (shown in Fig.~\ref{fig32:first}). Then, the pre-trained model is used to control the seen and unseen trajectories of UAVs in real-time. Whenever an unseen trajectory is provided, the DNN-controller starts updating its weights by using the online target information provided from the fuzzy logic rule-based system (shown in Fig.~\ref{fig32:second}). The fuzzy rules, stated in Table.~\ref{my-label1}, are defined by expert knowledge and incorporated in the online learning system at a much low computational cost.



\subsubsection{Control of unknown system dynamics}
The non-linear dynamics of UAVs highly vary during their diverse operating conditions (i.e., different payloads, height, and speed). To deal with these drastic changes, Deep Fuzzy Neural Network (DFNN)-based controller~\cite{b58} is developed to control the UAV in real-time. DFNN estimates the inverse dynamical model of the system, where it combines the deep neural network and fuzzy logic into a single framework. 

\begin{figure}
    \centering
    \includegraphics[width=0.75\linewidth]{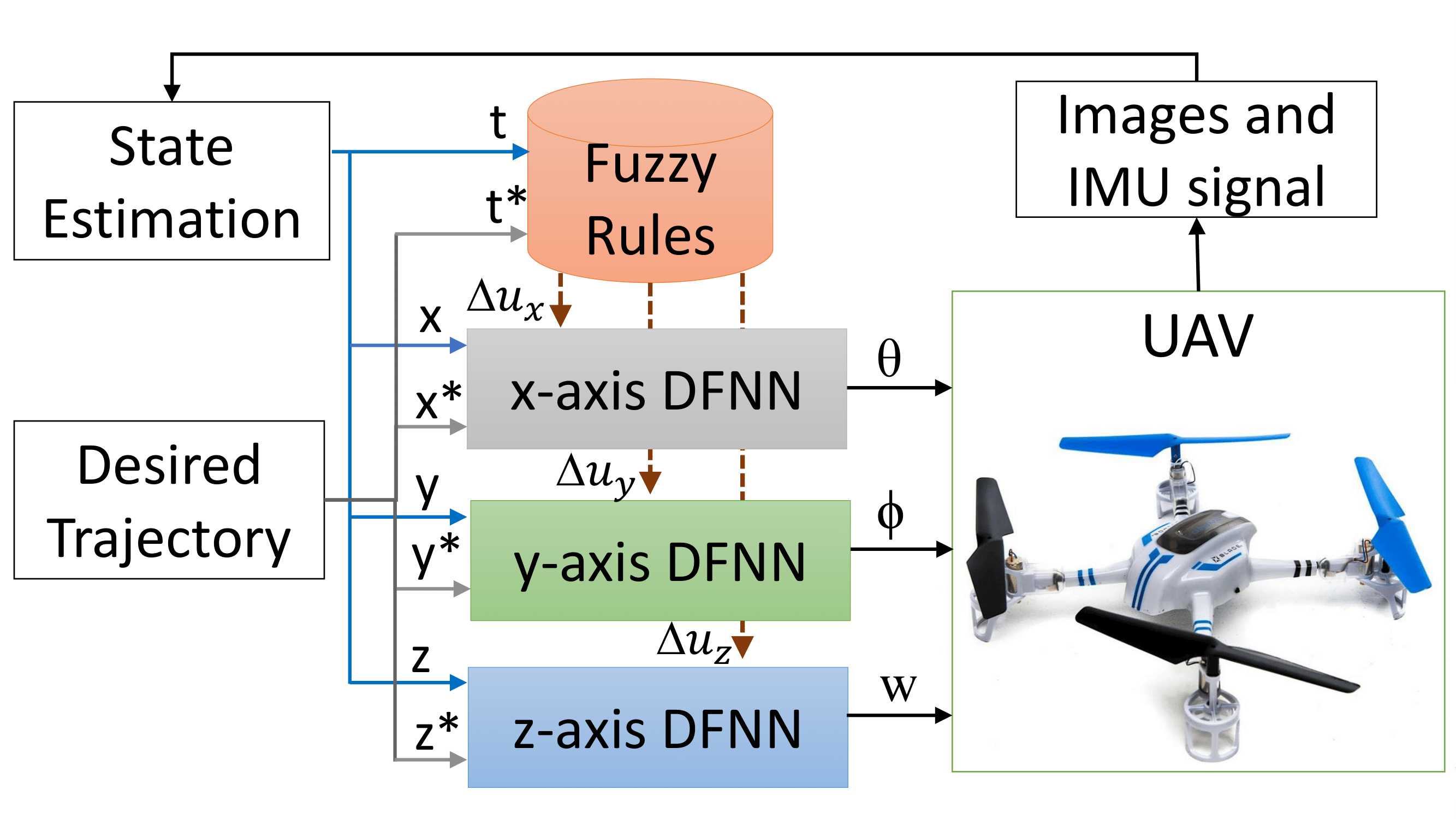}
    \caption{The online learning of DFNN-based controller. The DFNN controls the UAV and also improve its perfronmance by employing expert knowledge encoded in fuzzy rule base~\cite{b58}}
    \label{fig33}
\end{figure}

This method employs three parallel DFNN (one for each axis) as shown in Fig.~\ref{fig33}. The learning of DFNN is performed in two phases (1) Online learning (2) Offline learning. The offline learning of DFNN is performed on the data collected from the conventional controller. During online learning, the pre-trained DFNN controls UAV and improves performance by adapting the unknown dynamics. The online labels are provided by expert knowledge that is encoded in the form of fuzzy rules. FLS monitors the behavior of DFNN and, it corrects the action of DFNN depending on the variations in the system dynamics. 
 
DFNN is a computationally suitable choice for real-time applications as it exhibits faster response time and reduces tracking errors effectively.

\begin{table*}[]
 \caption{Fuzzy rule base for updation of control signal~\cite{b57}}
\label{my-label1}
 \renewcommand\arraystretch{1.1}
  \resizebox{1\linewidth}{!}{
\begin{tabular}{p{2cm}p{3cm}p{3cm}p{3cm}p{3cm}}
\hline
error& \multicolumn{3}{c}{Derivative of error}\\
\cline{2-4}
 &Negative &Zero &Positive    \\ \hline 
Negative &R1:Big decrease &R2:Small decrease  & R3:No change \\ \hline 
Zero &R4:Big decrease& R5:No change  & R6:Big increase \\ \hline 
Positive & R7:No change &R8:Small increase  &R9:Big increase \\ \hline 
\end{tabular}}
\end{table*}

\subsubsection{Online System Identification from datastreams}

The UAVs require a highly efficient and accurate classification system, which can perform online learning at much low computation cost. Such system can be realized through an evolving intelligence system (EIS) that controls the complexity of the network and exhibits single-pass online learning.

In~\cite{b59}, a variant of EIS, i.e., Metacognitive Scaffolding Interval Type 2 Recurrent Fuzzy Neural Network (McSIT2RFNN), is used to identify and learn the non-stationary time-series data stream in real-time. McSIT2RFNN contains an internal memory structure in the form of recurrent connections that deal with temporal dynamics. It also exhibits a single fuzzy rule layer that classifies the incoming data and able to detect the uncertainties of the system. This system properly addresses the issues of network complexity, as it adds the new rule for each novel concept and eliminates the inactive rules. Moreover, this method is sample efficient, and selects the most appropriate and uncertain training data.
 
This method does not require offline pre-training. It can perform online learning from scratch. Due to single-pass learning, it requires fewer computing and memory resources, which makes it an ideal online learning solution for real-time applications. 
 
\subsubsection{Deep Reinforcement Learning for Navigation and Exploration}

The exploration of the unseen environment during search and rescue operations is highly difficult for UAVs. As it follows the shortest path for reaching the target location and, it also requires avoiding the obstacles and cluttered areas during the flight.

\begin{figure}
    \centering
    \includegraphics[width=3.0in]{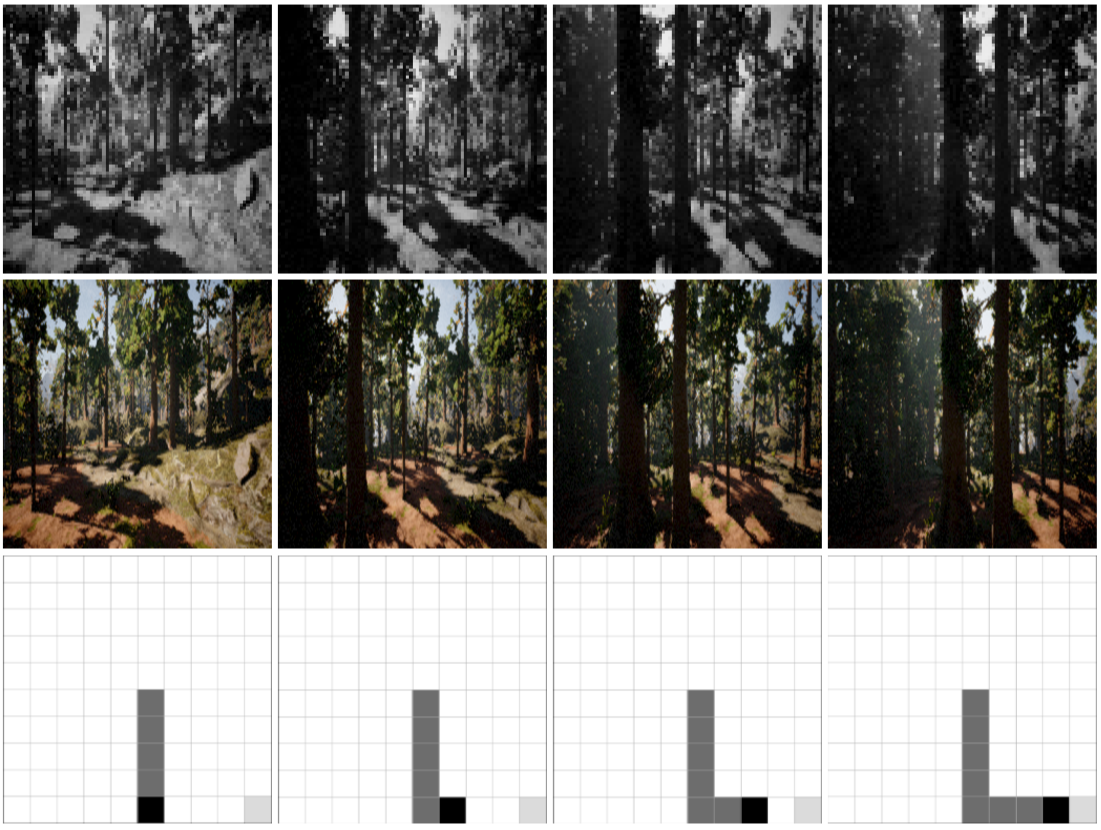}
    \caption{Double State input used in online reinforcement learning. The top row is the input raw image used by network, the middle row shows the RGB version of raw images and bottom row is the decision map of the mission that serves as second input of the system~\cite{b60}.}
    \label{fig34}
\end{figure}

For the exploration of a partially observable environment,
a deep learning-based method is developed named Extended
Double Deep Q-Network (EDDQN)~\cite{b60} that utilizes a double
state-input strategy (shown in Fig.~\ref{fig34}), combining the
acquired knowledge from the raw images with the map
containing positional information. The map helps to estimate
the target location, while the images identify the cluttered areas.

This approach will boost the system's overall performance over multiple flights.  During the SAR mission, it was evaluated on diverse unseen weather conditions (light, heavy snow, dust, and fog)  where it successfully adapted novel scenarios.


\subsubsection{Object Tracking}

Visual tracking is an essential task in the field of UAVs. The tracking system needs to be updated online for the changing visual appearance of the tracked target under different environmental conditions.

To adapt changing appearances, an online CNN-based tracking method~\cite{b61} is proposed that learns robust features by incorporating the locally connected layers.  Moreover, this framework also employs focal loss that enables to learn hard examples and avoids the class imbalance issue.

\begin{figure}
    \centering
    \includegraphics[width=3.0in]{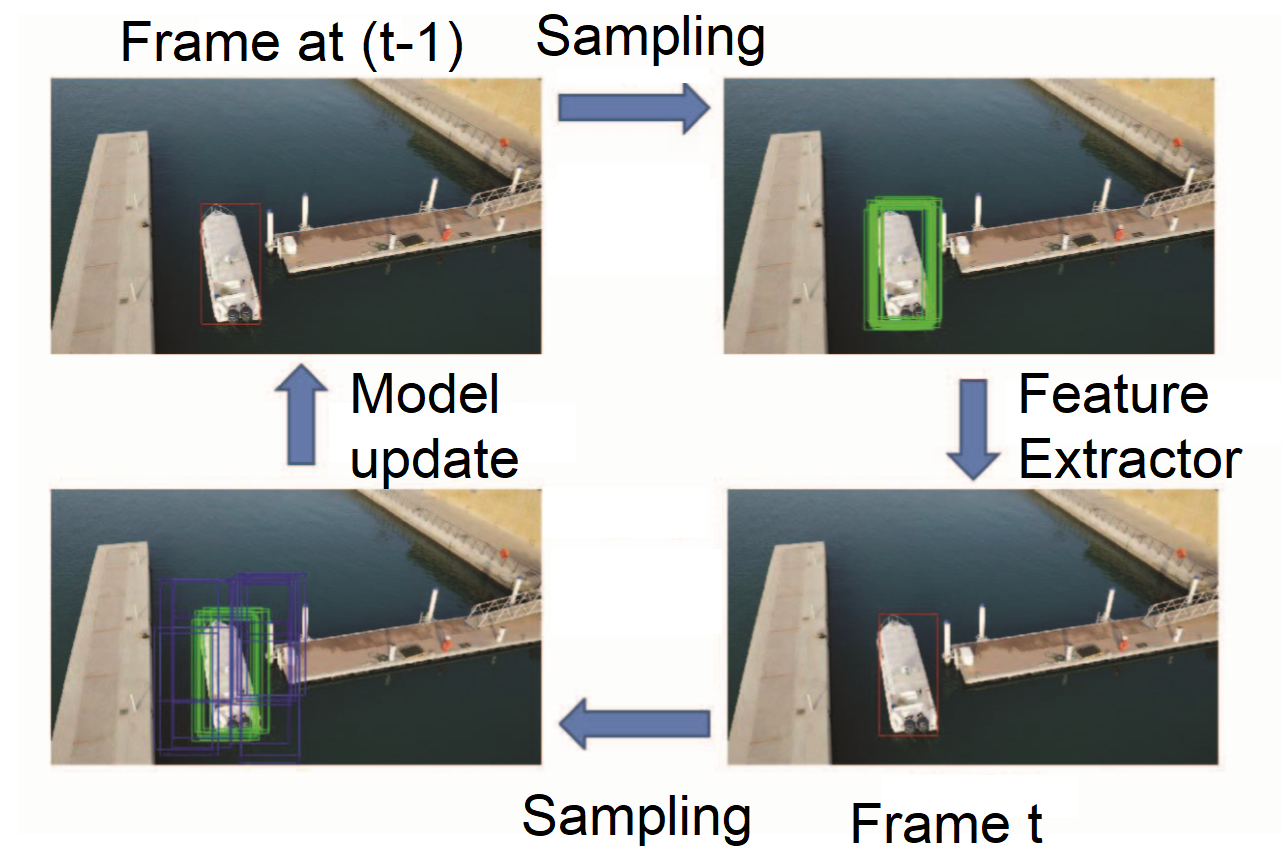}
    \caption{Online object tracking and its adaptation process~\cite{b61}.}
    \label{fig36}
\end{figure}

 During the online tracking phase, the offline pre-trained network is used to detect the target online and produce the features that separate the foreground object from the background. In the subsequent online learning phase, fine-tuning is performed to adapt to the changed appearances. The online fine-tuning process is briefly explained in Fig.~\ref{fig36}. Samples (green rectangles) are collected around the target position in a single time frame. To extract features, the CNN model is used, and the best candidates with the highest positive scores are found. New training examples are extracted around the target location (green rectangles and blue rectangles) in the next time frame, and the model is modified.

This method improves the tracking performance and is also beneficial for large-scale classification.

\subsection{Urban Robots}

In urban robots, the deep learning models enable a wide range of perception (i.e., visual, material, manipulation, traversability), so they must possess the ability to continuously improve performance in the changing environment. 

This section discusses few use cases of life-long learning in the urban robots framework. A short description of these techniques is also discussed in Table~\ref{my-label_r_1}. 

\begin{sidewaystable}
\sidewaystablefn%
\begin{center}
\begin{minipage}{\textheight}
 \caption{Summary Table of Real-time Continual learning applications in Urban Robots}
  \label{my-label_r_1}
  \centering
  \renewcommand\arraystretch{0}
  \resizebox{1\linewidth}{!}{
  \begin{tabular}{p{2cm}p{2cm}p{2cm}p{10cm}p{2cm}p{2cm}}

    \hline
    \textbf{CL Strategy} 
    &\textbf{Learning} & \textbf{Application} 
    &\textbf{Description}&
    \textbf{Motivation}&
    \textbf{Limitation}
    \\ 
    \hline
    \multirow{2}{*}{Rehearsal}&
    Supervised&
    Under Water Vehicle \cite{b62}&
    The online adaptation of the SVM-based controller is performed to the varying system dynamics. 
    &Less computations
    &--
    \\ \cline{2-6}
    & 
    Reinforcement &
    Object Grasping System\cite{b65}&
    The continuous adaptation for vision-based robotic manipulation is accomplished by fine-tuning of policy network via off-policy reinforcement learning methods.
    &Less exploration, high accuracy&Prone to over-fitting
    \\ \cline{1-6}
    Rehearsal \& Regularization \& Architectural & 
    Supervised& 
     Open World Recognition\cite{b66} &An end-to-end deep learning-based object recognition framework detects and learns  the unknown objects by incorporating web-based supervision. 
    & Webly supervision &Less accurate 
    \\ \cline{1-6}
    Regularization&
    Supervised
    &Material Percep- tion\cite{b67}&
     Life long cross-modal learning is performed to learn the new tasks incrementally about material perception.
    & High accuracy&-- 
    \\ \cline{1-6}
    \multirow{3}{*}{Architectural} 
    &Supervised 
    &  Object and motion recognition\cite{b69} & 
    Human-robot interaction enables the robot  to incrementally learn the new object and its associated motion on-the-fly. 
    &High accuracy& High learning time
    \\ \cline{2-6}
    & 
    Self-Supervised
    & Action classification in human\cite{b64} & Online learning of human activity classification is performed to adapt the average user model to specific user behaviour.
    & Less computation & --
    \\ \cline{2-6}
    & 
    Self-supervised 
    &
     Sound Source Localiza- tion\cite{b68} & 
    Incremental learning of sound source localization model under multi-room environment is performed by using own positive experiences.
    & Fast adaptation&--
    \\ \cline{1-6}
\multirow{2}{*}{Naive}& 
Self-supervised (tracking error)
&  Neural adaptive controller\cite{b63} & The neural network-based dynamic control of flexible joints quickly adapt the uncertain dynamics by using different learning rates for hidden and output layers.
& Fast adaptation&--
\\ \cline{2-6}
&
Self-supervised 
& Navigation\cite{b70} & 
    An end-to-end learning-based mobile robot navigation system uses self-supervised off-policy data of real-world environments, and learns an efficient policy network without human supervision. &Fast adaptation&--\\ \cline{1-6}
    
\end{tabular}}
\end{minipage}
\end{center}
\end{sidewaystable}

\subsubsection{Control Systems:}
Deep learning-based controllers are mostly employed to model the robot dynamics for real-world applications. 
These control systems must be updated online to maintain high performance under varied operating conditions.

\begin{itemize}
\item Autonomous Underwater Vehicles Dynamic Control:

The marine robot dynamic's model interacts with a highly varying external environment due to different payloads and disturbances. To deal with such scenarios, the incremental support vector regression (IncSVR) algorithm~\cite{b62} is proposed, which performs online Autonomous Underwater Vehicles (AUV) adaptation. IncSVR effectively models the input-output relationship of AUV dynamics and learns the prediction model using only support vectors (input samples that are robust to decision boundaries), resulting in lower computational complexity.

Fig.~\ref{fig37} illustrates the online learning process of the SVR-based control system. The sample selector selects the support vectors (SVs) based on the prediction error and correlation with the current vector, thus avoids the redundant training data. The batch sampler concatenates the newly extracted SV with the vectors from the forgetting buffer and performs incremental learning. To prevent outlier samples from being stored in the forgetting buffer, the resulting support vector is again passed through an outlier filter.

IncSVR significantly reduces forgetting while maintaining high accuracy. It consumes less memory and computations as it uses a small set of training samples, i.e., support vectors. 

\begin{figure}[ht!]
    \centering
    \includegraphics[width=0.75\linewidth]{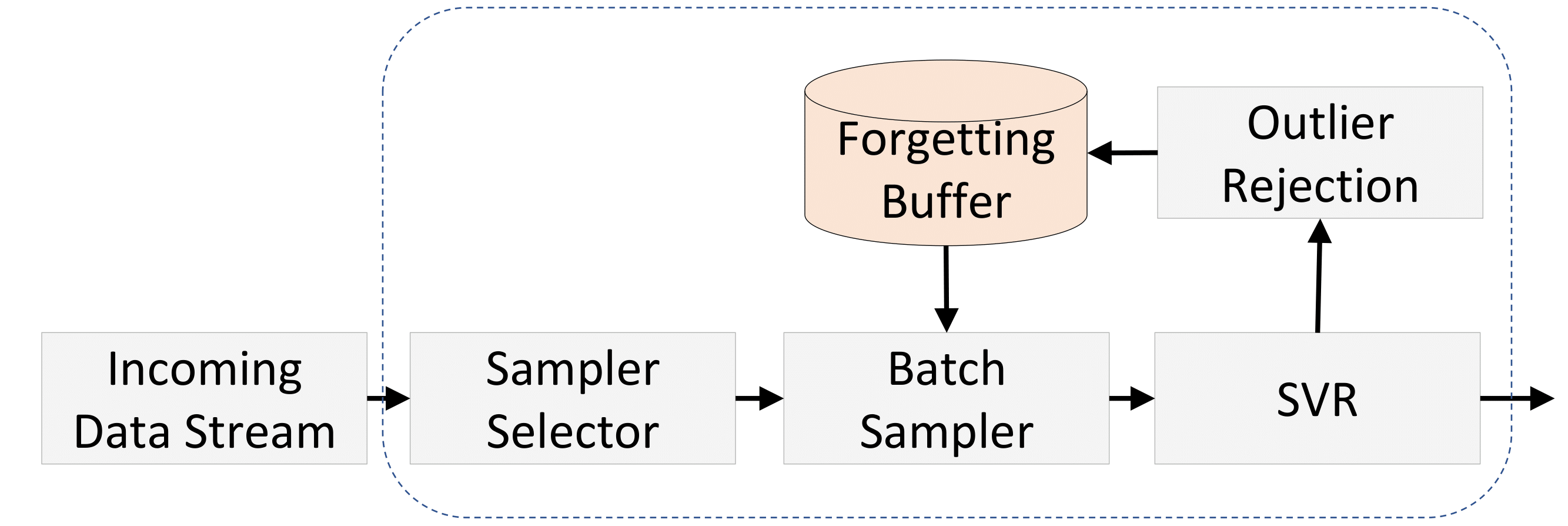}
    \caption{Online training of SVR-based control system for AUV dynamics~\cite{b62}}
    \label{fig37}
\end{figure}
\item Adaptive Trajectory Tracking Control for Flexible-Joint of manipulator:

The control design of a robot manipulator is highly complicated, and tracking accuracy is minimal. Adaptive controllers are commonly employed to reducing trajectory tracking error~\cite{b63}.

Fig.~\ref{fig38} depicts the architecture of an adaptive controller. The multi-layer perceptron neural network, which describes the linear performance and non-linear basis of robot dynamics, is used to model the unknown dynamics of the flexible-joint robot. The weights of the output layer are updated based on tracking error, allowing it quickly adapts to the changes of the system dynamics. The further update in the internal weights also reduces the tracking error. Internal and outer weights have different update rates; the internal weights are updated at a slower rate than outer weights, resulting in fewer computations and faster tracking error convergence.
 \begin{figure}[ht!]
    \centering
    \includegraphics[width=0.8\linewidth]{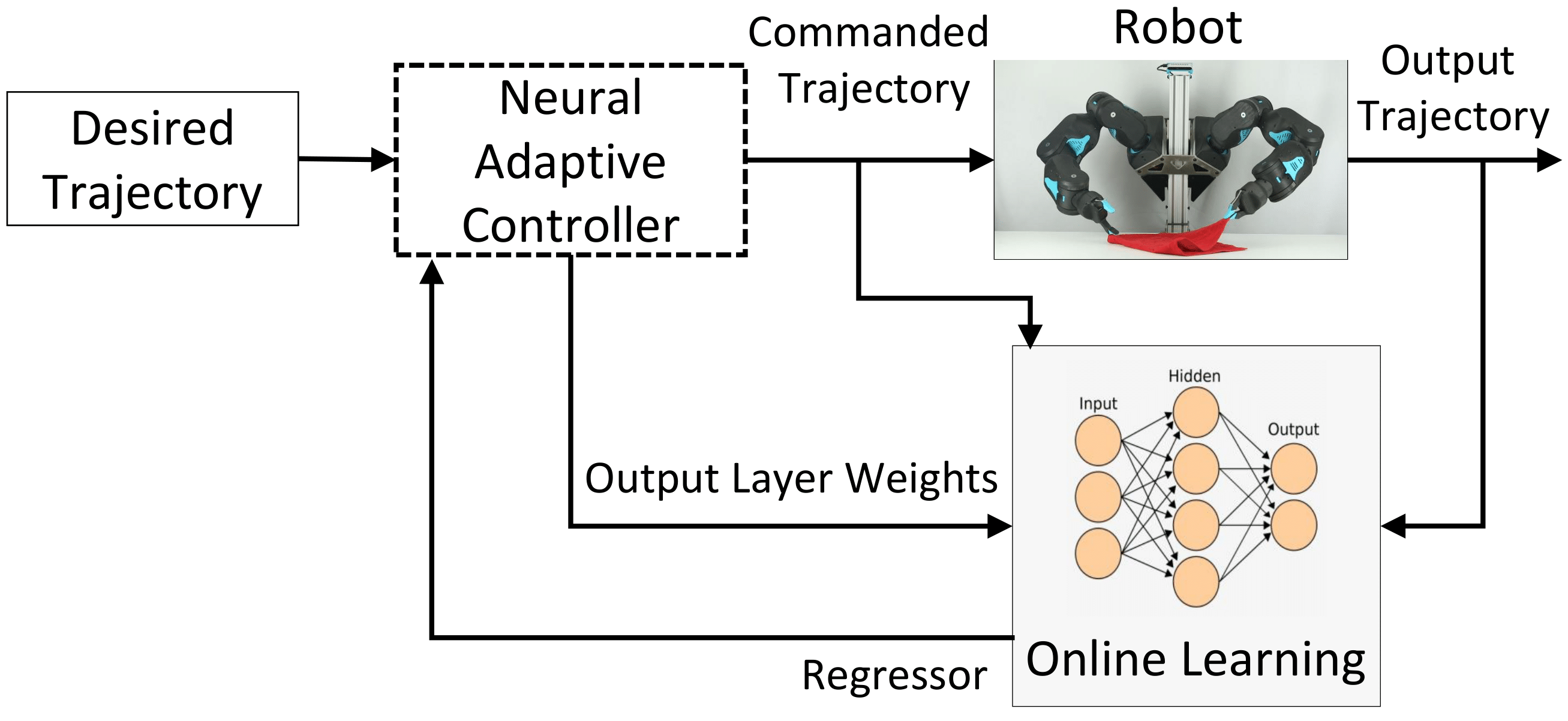}
    \caption{The incremental learning-based control system architecture of a flexible joint robot~\cite{b63}.}
    \label{fig38}
\end{figure}
\end{itemize}
\subsubsection{Vision-Based Manipulation}
 When vision-based object grasping manipulation policy is subjected to drastic changes (i.e., background, lightning condition, gripper shape, and new objects), its efficiency suffers greatly. To adapt to these major changes, an end-to-end fine-tuning-based reinforcement learning approach~\cite{b65} is proposed. 

\begin{figure}[ht!]
    \centering
    \includegraphics[width=3.5in]{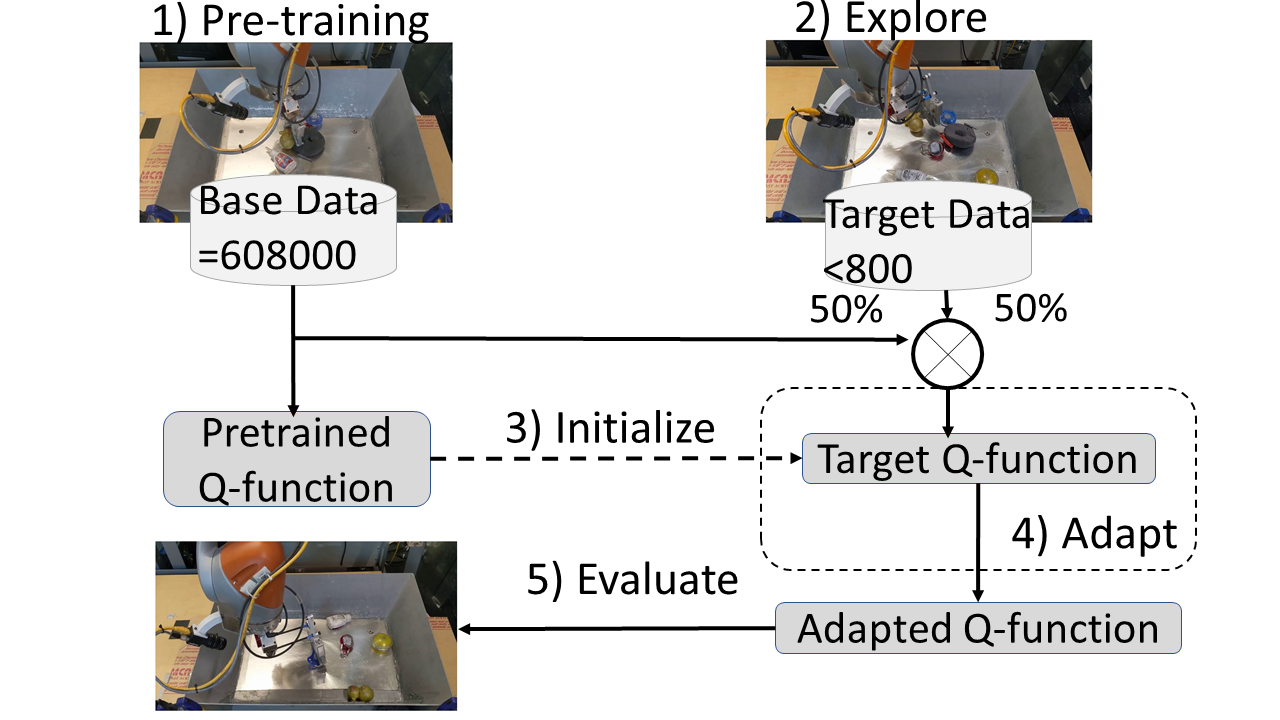}
    \caption{The schematic diagram of fine-tuning process for vision-based manipulation~\cite{b65}.}
    \label{fig39}
\end{figure}

Fig.~\ref{fig39} depicts the data flow of continuous adaptation by fine-tuning. For exploration, a pre-trained policy network is used to gather a new training dataset related to the target task. The new policy network is initialized with the parameters of the pre-trained policy network. Its fine-tuning is performed on training examples sampled from base task data and new task data with equal probability. The new policy network's learning is an offline process that does not require any interaction with the real-world environment.
The combination of fine-tuning and reinforcement learning substantially improves the performance of a vision-based manipulator over novel tasks. 
\subsubsection{Web Aided Deep Open World Recognition(OWR)}

Within the visual perception system, the robots always interact with novel objects. Conventionally, visual recognition is considered as a closed world problem (in which it remembers a limited set of objects and associates every new object to the limited known classes). While the deep OWR method~\cite{b66} formulates the visual perception as the open-world recognition system in which the robots can detect the unknown classes and learn them incrementally. This approach also solves the main problem of incremental learning (i.e., the non-availability of labels and annotated training data) by combining the oracle (human) with web-based monitoring. 

\begin{figure}[ht!]
    \centering
    \includegraphics[width=3.5in]{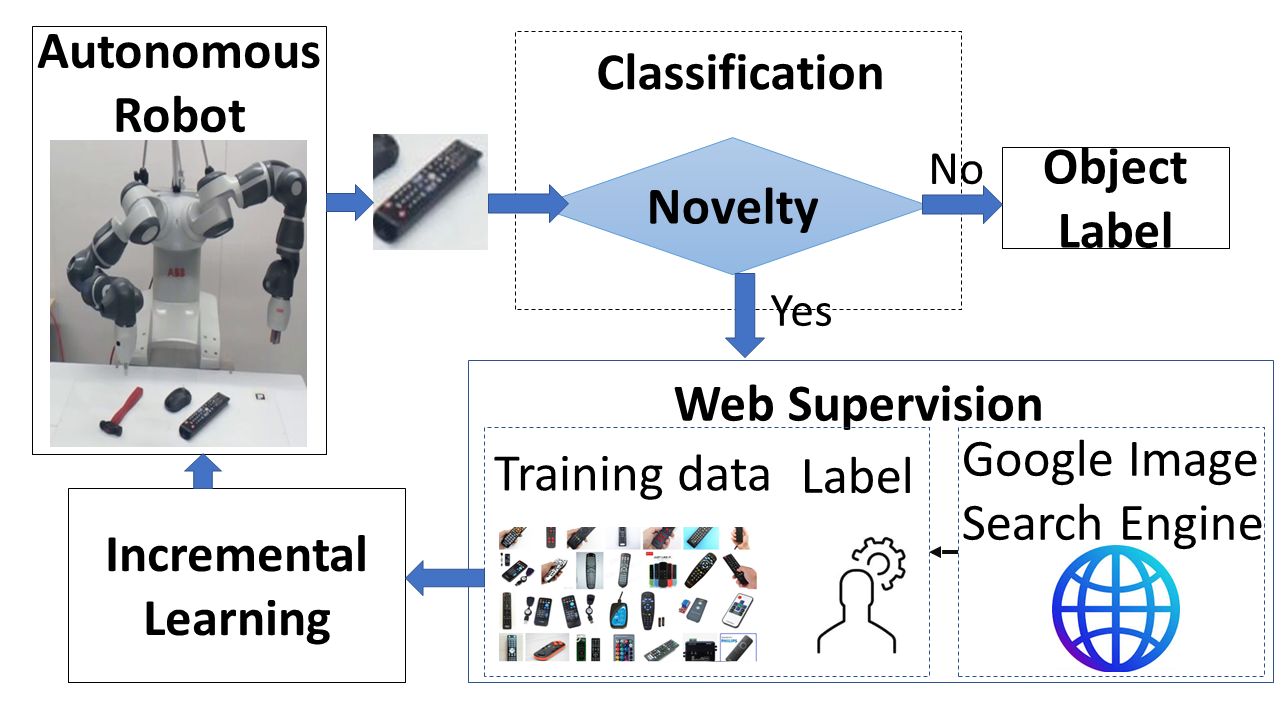}
    \caption{Open world recognition mechanism to incrementally learn unknown object from web-supervision~\cite{b66}}
    \label{fig40}
\end{figure}

The methodology of Deep OWR is discussed in Fig.~\ref{fig40}. DeepOWR allows the robot to recognize unseen objects and incorporates web mining that provides the corresponding label and additional training data. For incremental learning, the deep learning representation is incorporated with the non-parametric nearest non-outlier method, and trained in an end-to-end fashion. To avoid catastrophic forgetting, the rehearsal buffer along with regularization (distillation loss) is used to maintain the previous knowledge.

Hence, deep OWR gives a highly efficient open-world recognition system by incorporating a web mining approach.

\subsubsection{Visual-Tactile Cross-Modal Learning for Robotic Material Perception}

Material perception cannot be realized solely through tactile information in novel and unstructured circumstances. The ability to realize human-level material perception is enabled by cross-modality based on visual-tactile data. To incrementally improve the ability of robotic cross-modal material perception, a unique approach of life-long cross-modal learning~\cite{b67} is developed.

\begin{figure}[ht!]
    \centering
    \includegraphics[width=3.5in]{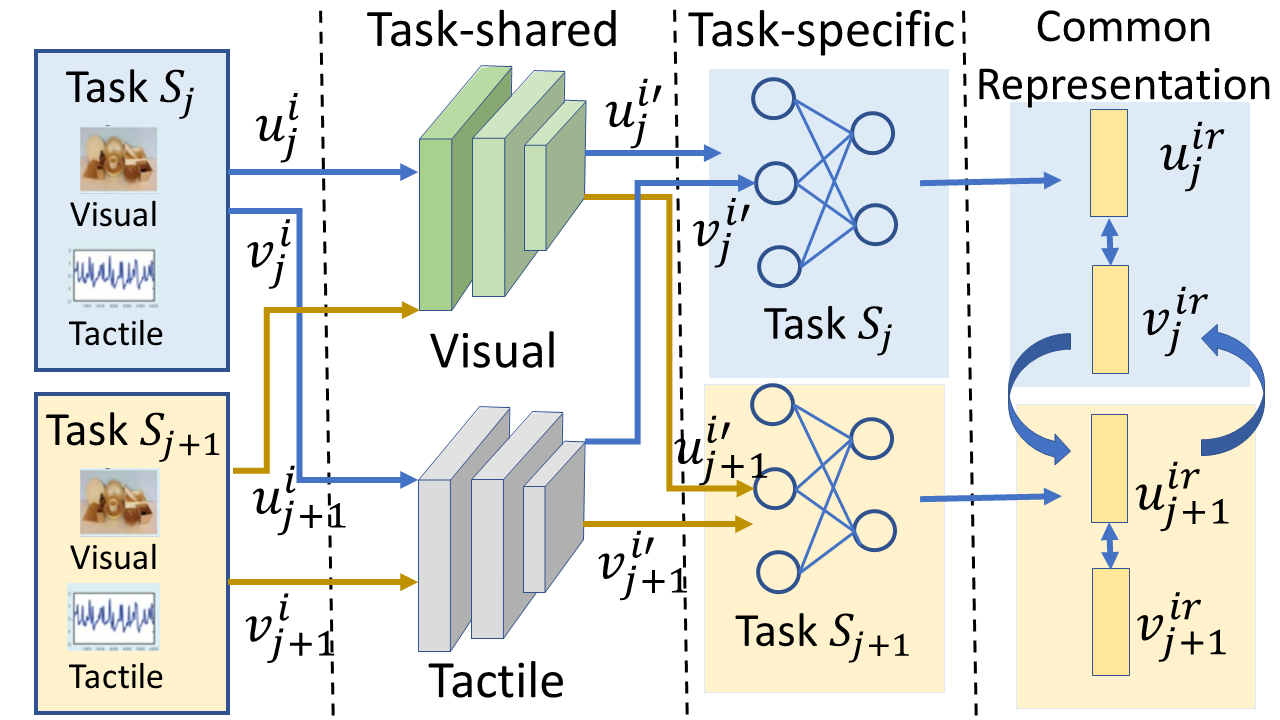}
    \caption{The learning process and architecture of life-long visual-tactile cross-modal system~\cite{b67}}
    \label{fig41}
\end{figure}

The network architecture used to learn heterogenous cross-modality is shown in Fig.~\ref{fig41}. It is composed of task-shared and task-specific sub-networks. The purpose of the task-shared network is to leverage the previously learned knowledge to learn the new task effectively; moreover, it also helps to learn the cross-modal correlation over different tasks. 

The optimization of the task-shared network is composed of inter-task and intra-task cross-modal correlation learning. For intra-task learning, it employed the end-to-end adversarial learning technique to learn the correlated representations between heterogeneous data. For inter-task learning, the distillation loss is used to avoid catastrophic forgetting. 
Finally, the learned task-shared network's representations are fed into the task-specific network, which produces a common representation that differentiates the specific material classes. 
Consequently, lifelong cross-modal correlation learning enables the robot to learn the material properties in an unfamiliar environment.

\subsubsection{Localization of human sound source in Complex Indoor Environment}

The sound source localization for the untrained and unknown environment is highly challenging for mobile robots. Specifically, across the multiple room environment, the localization system needs to ranks the room according to the likelihood of the sound source and incrementally learn new experiences over time.   

\begin{figure}[ht!]
    \centering
    \includegraphics[width=3.5in]{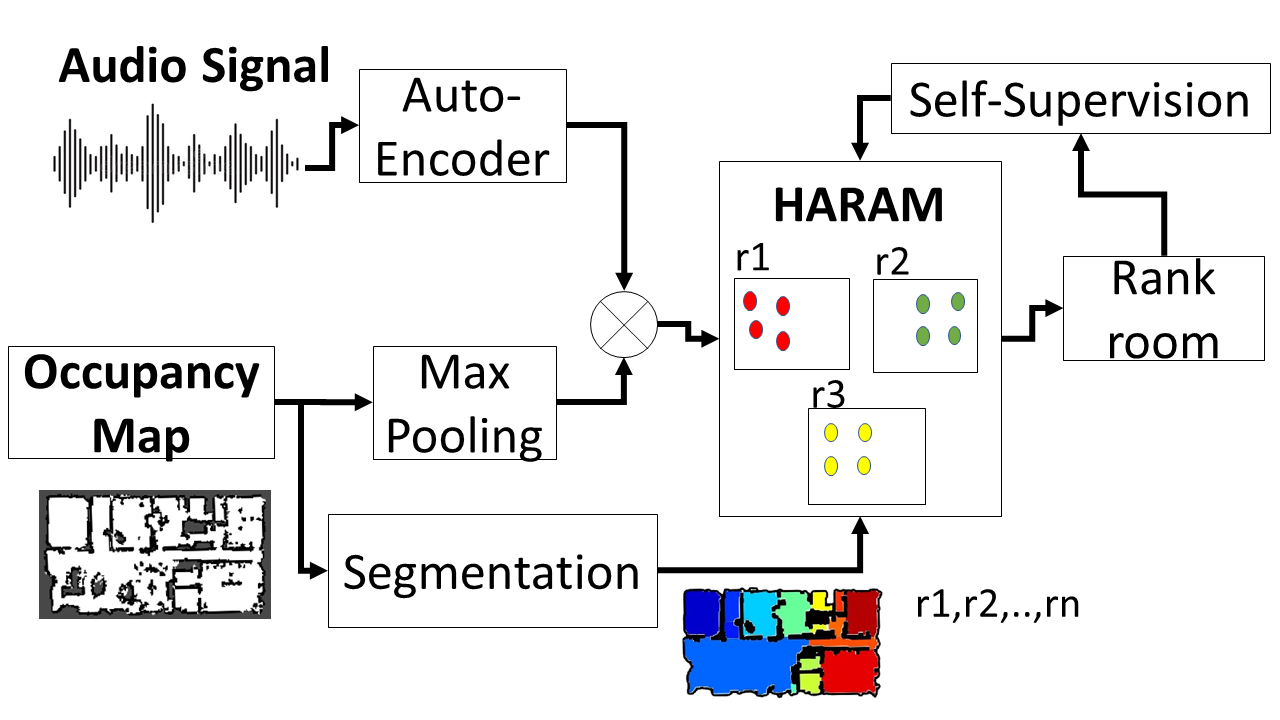}
    \caption{Self-supervised incremental learning scheme in which audio signal along with occupancy map is used to rank rooms~\cite{b68}}
    \label{fig42}
\end{figure}

In~\cite{b68}, an incremental learning framework is proposed that predicts the rank of room to explore the sound source and incrementally learn the positive experiences, as shown in Fig.~\ref{fig42}. 
It employed the Hierarchical Adaptive Resonance Associative Map (HARAM) model that utilizes the occupancy map (segmented) to identify the total number of candidate rooms for exploration (the number of rooms must be equivalent to the number of labels). This model can perform incremental learning from scratch without any prior training, and it further improves performance by collecting more real-time data. For the first sample, the system randomly guesses the exploration. Then afterward, it follows the rank provided by the learning model and verifies the predicted rank through human pose detection. The negative and positive examples are collected during exploration and incrementally learn the model to improve the performance of the system.


\subsubsection{Online Object and Task Learning via Human Robot Interaction}

The human-robot interaction~\cite{b69} provides a substantial way to incrementally learn new knowledge for the development of the robotic system. 

\begin{figure}[ht!]
    \centering
    \includegraphics[width=3.5in]{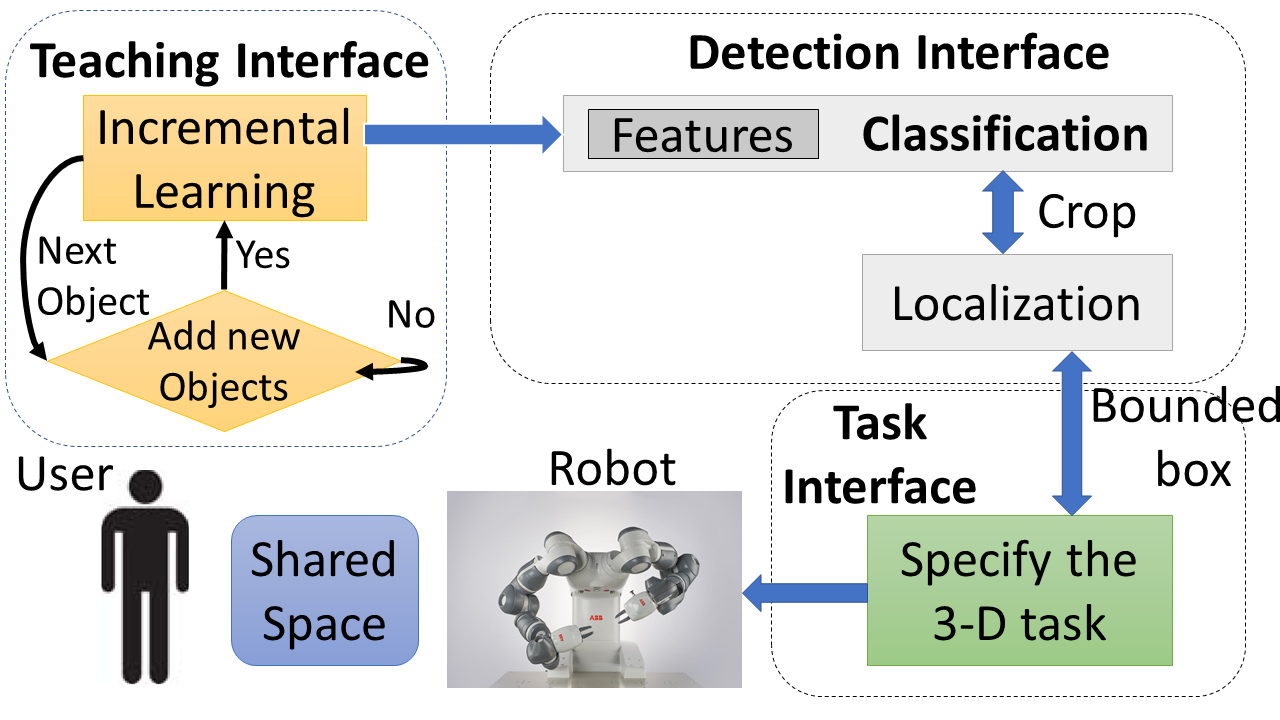}
    \caption{User interfaces during human robot interaction~\cite{b69}}
    \label{fig43}
\end{figure}

This approach discusses a robotic system having a user interfaces that allows to incrementally learn the system for the novel objects and their associated motion in real-time. The user interface provided by the system is described in Fig.~\ref{fig43}. The teaching interface allows the user to add or remove a new class.
The detection interface detects and displays the name and location of the new object on the screen. The task interface allows to add 3D motion (this can be done by drawing 2D trajectory on a touch screen) and associate that motion with one of the detected objects. Finally, the robot is able to pick the object and perform motion learned from the user. 
The human-robot interaction is beneficial for the acquisition of reliable and diverse training data about the novel objects. 


\subsubsection{An Autonomous Self-Supervised Learning-Based Navigation System}
The navigation in mobile robots is purely based on the geometry of the environment. 
While self-supervised system~\cite{b70} formulates the navigation beyond the geometric problem and allows the robot to learn from its own experiences. It employed an end-to-end learning scheme that performs supervised learning on the data gathered from off-policy exploration in a real-world environment. This method still can improve performance for the novel scenarios by collecting more data and fine-tune the pre-trained model on the new data.

The self-supervised reinforcement learning-based method enables the robot to learn from its own collision-free experiences. The random control policy is employed to gather online data from a real-world environment. The labeling of raw data is performed in a self-supervised manner such that it uses the observations from LIDAR and IMU under three major scenarios of navigation, i.e., collision, bumpiness, and position. 

This method outperforms the other navigation policies as it can generalize over a novel environment and improves performance by gathering a small amount of training data.

\subsubsection{Personalized Online Learning of Whole-Body Motion Classification}
The online action classification is necessary for the control of wearable healthcare devices. The average user classification model is not able to accurately predict individual motion behavior. The system utility would be enhanced by maximizing the system-user collaboration. It is achieved by developing a personalized model that can be adapt to the personal behavior of the user in real-time. 

\begin{figure}[ht!]
    \centering
    \includegraphics[width=3.5in]{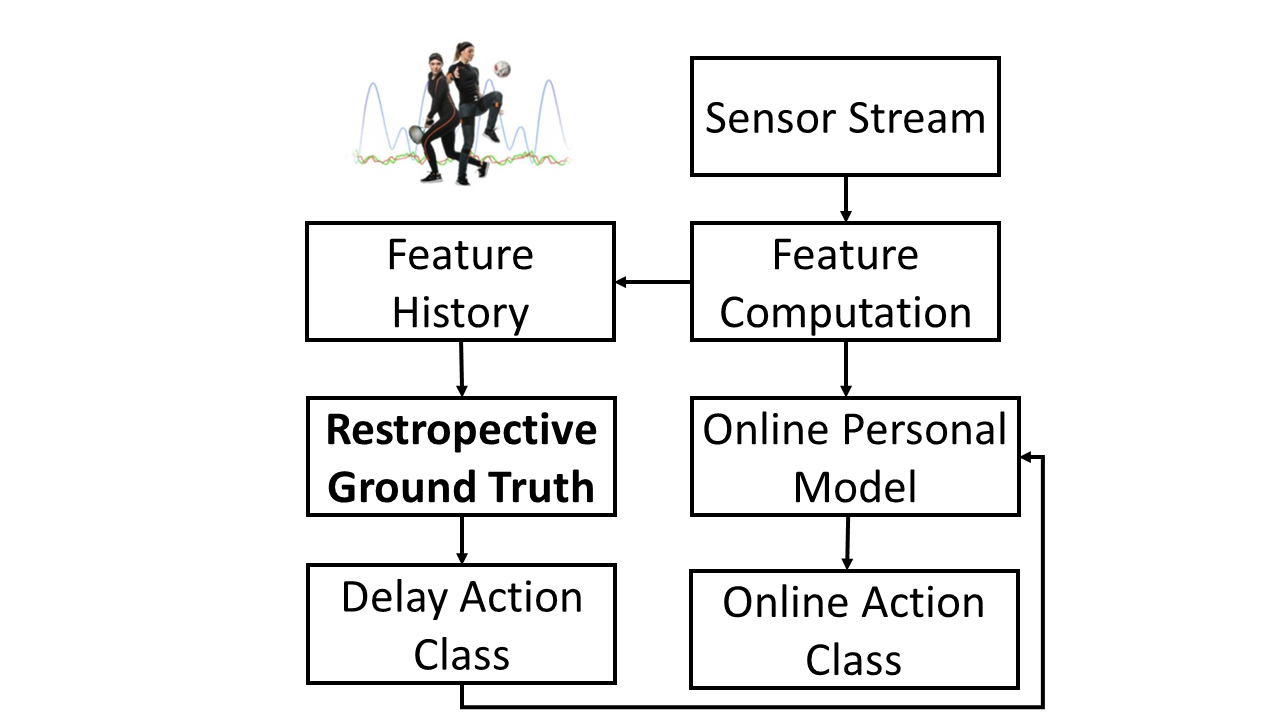}
    \caption{The online learning architecture of Whole-Body motion classifier. A seperate model is used to retreive the ground truth, it stores the some previous history and produce the delayed action as ground truth~\cite{b64}}
    \label{fig44}
\end{figure}

For personalization, an online learning algorithm~\cite{b64} is developed that allows updating the average model, as described in Fig.~\ref{fig44}.
This approach is sample efficient as it allows to learn on the samples that have a high testing error. To acquire the online training data, the IMU sensors are employed to accurately observe the posture and position of the body, and the annotated data is obtained retrospectively. 

The personalized model outperforms the average user model. Moreover, it utilizes fewer computations and attains high accuracy. 

\section{Other Applications}
\label{section6}
\subsection{Anomaly Detection Using Deep Learning Real-Time Video Surveillance} 

The incremental Spatio-temporal learner (ISTL) algorithm~\cite{b71} is developed to adapt the non-stationary anomalous behavior in real-time. This approach employs the auto-encoder to reconstruct the spatio-temporal data, which detects the anomalous data based on the reconstruction error.

This system continuously learns the evolving normal behavior through the use of active learning (fuzzy aggregation). It selects the most appropriate sample for continual learning based on the prediction error. Moreover, human feedback is also used to confirm the presence of anomaly in real-world scenario. Hence, this method provides a robust anomaly detection system that adapts the evolving normal behavior on the fly.  

 \begin{figure}[ht!]
    \centering
    \includegraphics[width=3.5in]{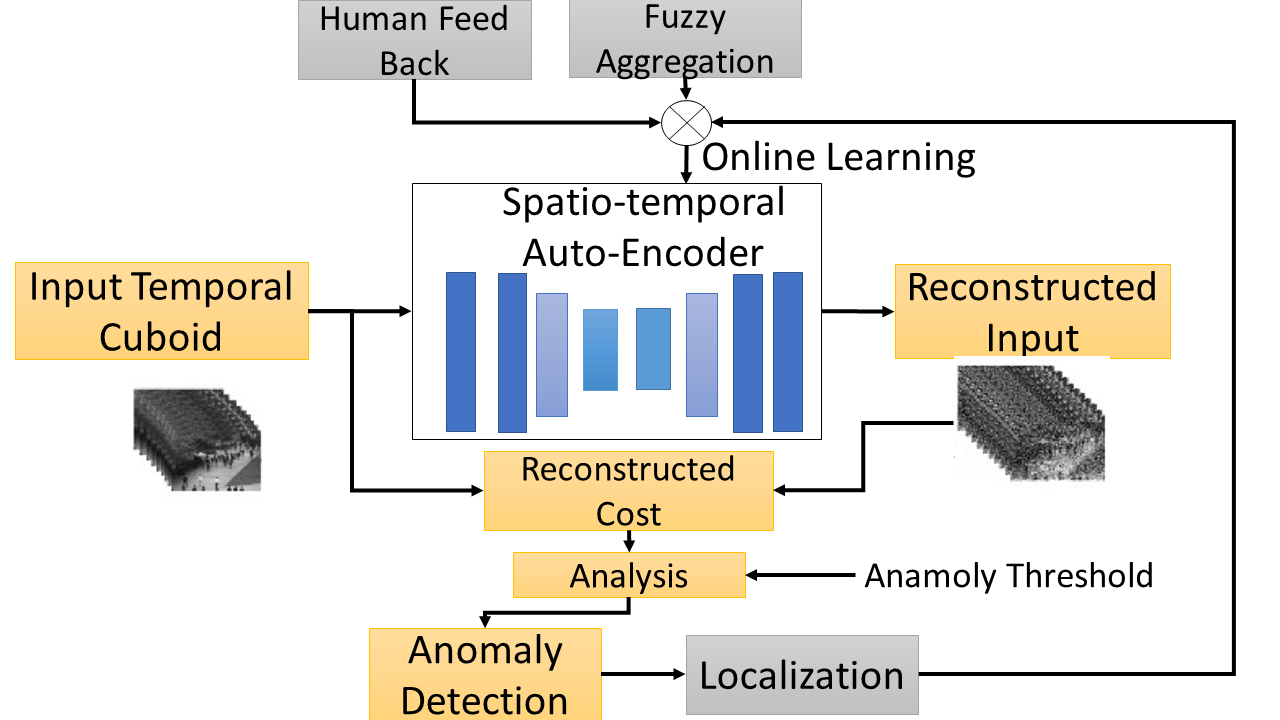}
    \caption{The anamoly detection performed by spatio-temporal auto-encoder. The online learning is performed through the fuzzy aggregation and human feedback~\cite{b71}.}
    \label{fig45}
\end{figure}
\subsection{Online deep learning-based energy forecasting}
\begin{figure}[ht!]
    \centering
    \includegraphics[width=0.75\linewidth]{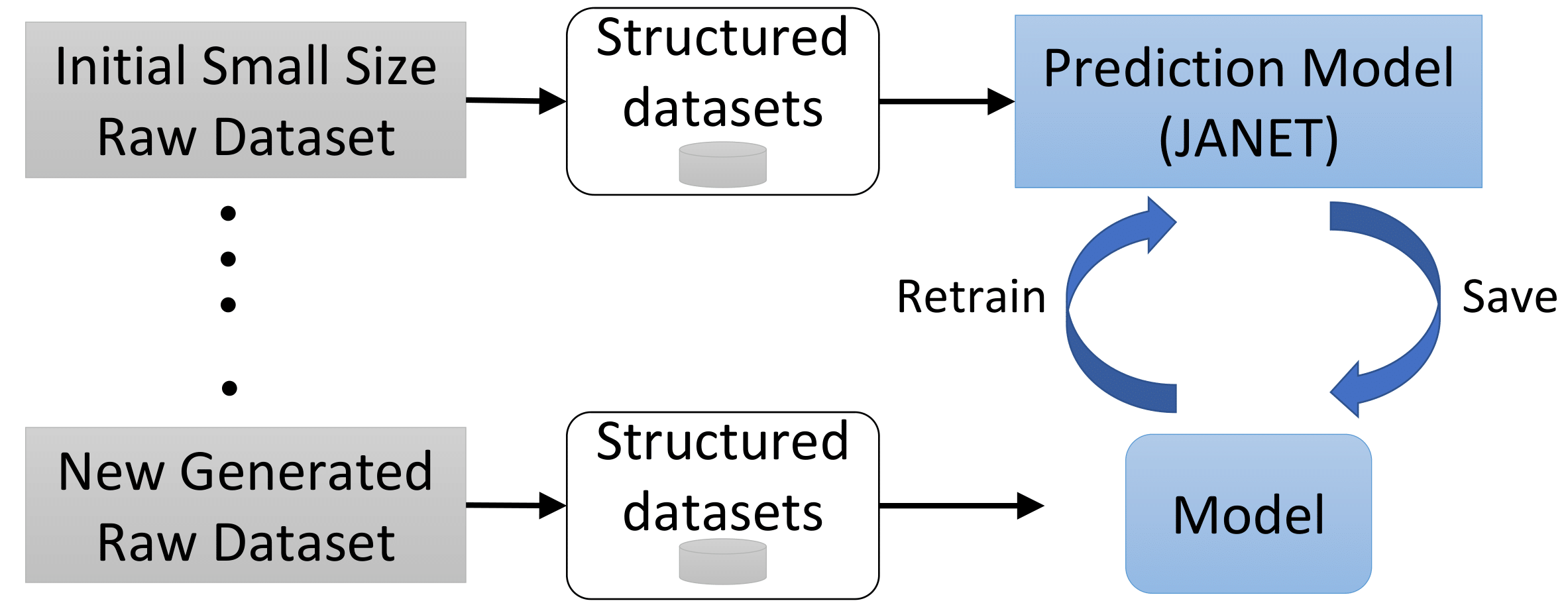}
    \caption{Online learning process of JANET on new incoming data~\cite{b72}.}
    \label{fig46}
\end{figure}

The energy prediction models work in a time-sentive and varying environment. The traditional centralized data analytics structure necessitates uploading the entire dataset to the cloud datacenter for analysis, which consumes a lot of network resources and causes congestion. To address these issues, a novel method~\cite{b72} is proposed that allows deep learning to be allocated at the network edge and developed an online learning approach to enable small data subset training and continuous model updating in time-sensitive environments. In this approach, the pre-trained DNN is learned incrementally on the small set of data acquired at a fixed time interval, as shown in Fig.~\ref{fig46}. This strategy employs the JANET (Just Another Network), an optimized variant of LSTM that avoids the high computation during online learning.    
\subsection{Incremental Continuous Load Prediction in Energy Management Systems}
Continuous energy load prediction is highly demanding as it requires continual learning of the novel scenarios. For the energy prediction, a novel method based on the long short-term memory (LSTM) network~\cite{b73} is developed that predicts and incrementally learns the new forecasting situations. The LSTM helps to retain the different temporal correlations of forecasting data.

\begin{figure}[ht!]
    \centering
    \includegraphics[width=0.7\linewidth]{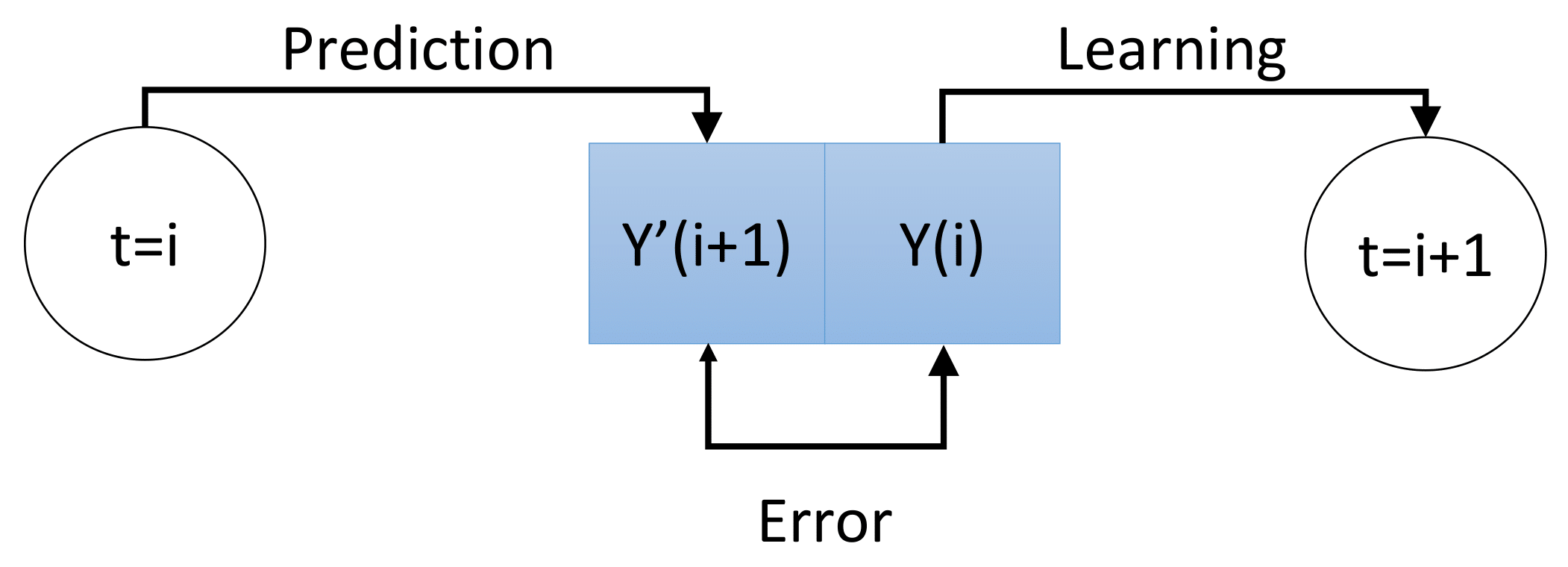}
    \caption{Incremental prediction and learning process in LSTM for energy forecasting~\cite{b73}}
    \label{fig47}
\end{figure}

The incremental learning is done by forecasting the load value and then computing the prediction error when the exact value arrives. It learns the novel incoming data based on the prediction error, as illustrated in Fig.~\ref{fig47}.

\subsection{Real-time adaption for deep stereo}
The online adaptation of a deep stereo network allows maintaining its performance in an unconstraint environment.
In ~\cite{b74}, a new lightweight, Modularly ADaptive Network (MADNet) along with a Modular ADaptation (MAD) algorithm is developed, which independently trains sub-portions of the network and keeps a high enough frame rate. 

\begin{figure}[ht!]
    \centering
    \includegraphics[width=3.5in]{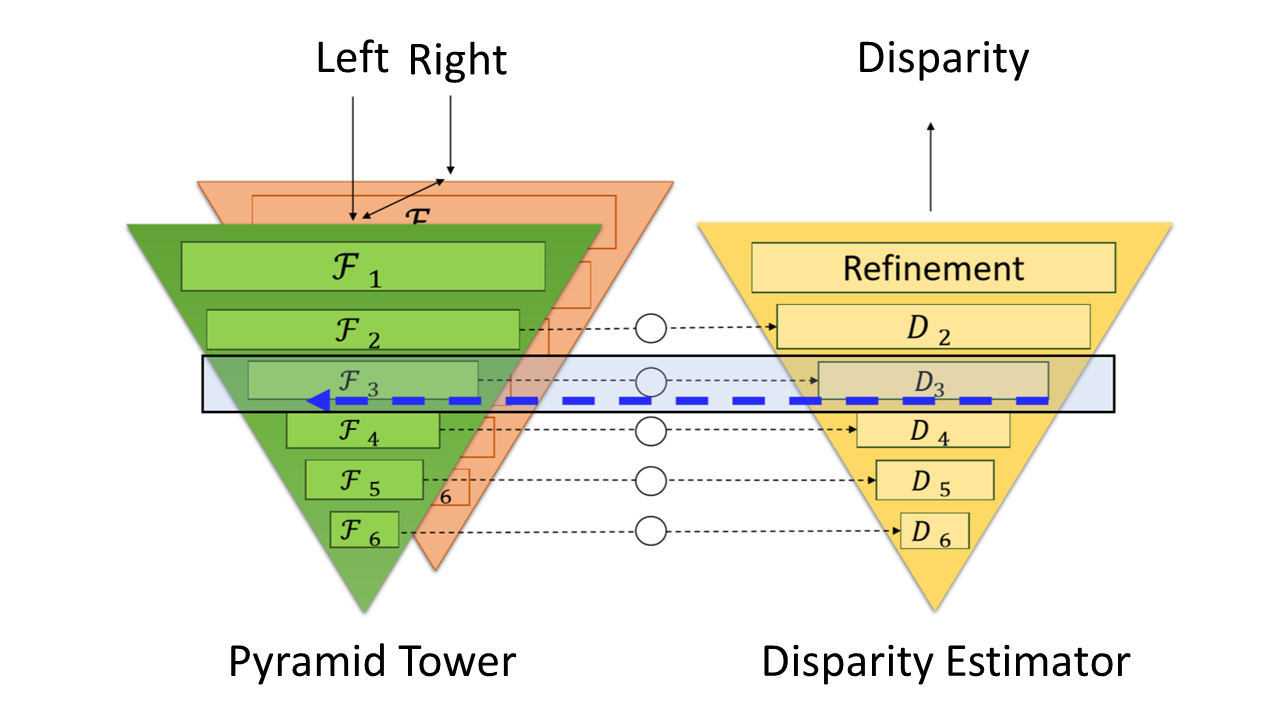}
    \caption{MAD architecture: it is divided in horizontal portions, the online learning is performed across the layers with high prediction error~\cite{b74}.}
    \label{fig48}
\end{figure}

MADNet divides the network into non-overlapping horizontal segments, each of which is referred to as a separate module, as shown in Fig.~\ref{fig48}. At each training iteration, one of the modules is optimized independently from the others by using their output prediction loss and then executing the shorter back-propagation only across the layers with high output loss. As a result, the learning time and computation complexity are significantly reduced. 

\subsection{Incremental Learning for Chemical Fault Diagnosis }
Incremental imbalance modified deep neural network (Incremental-IMDNN)~\cite{b75} is developed to promote the fault diagnosis for an imbalanced data stream. 

During incremental learning, the data is feed to the fault diagnosis module, which generates the balance input fault data. The active learning module is used in the feedback loop that will extract the most informative information from the diagnosis data. Then, this data is feed to a hierarchical neural network that performs incremental learning through network expansion. 
 \begin{figure}[ht!]
    \centering
    \includegraphics[width=3.5in]{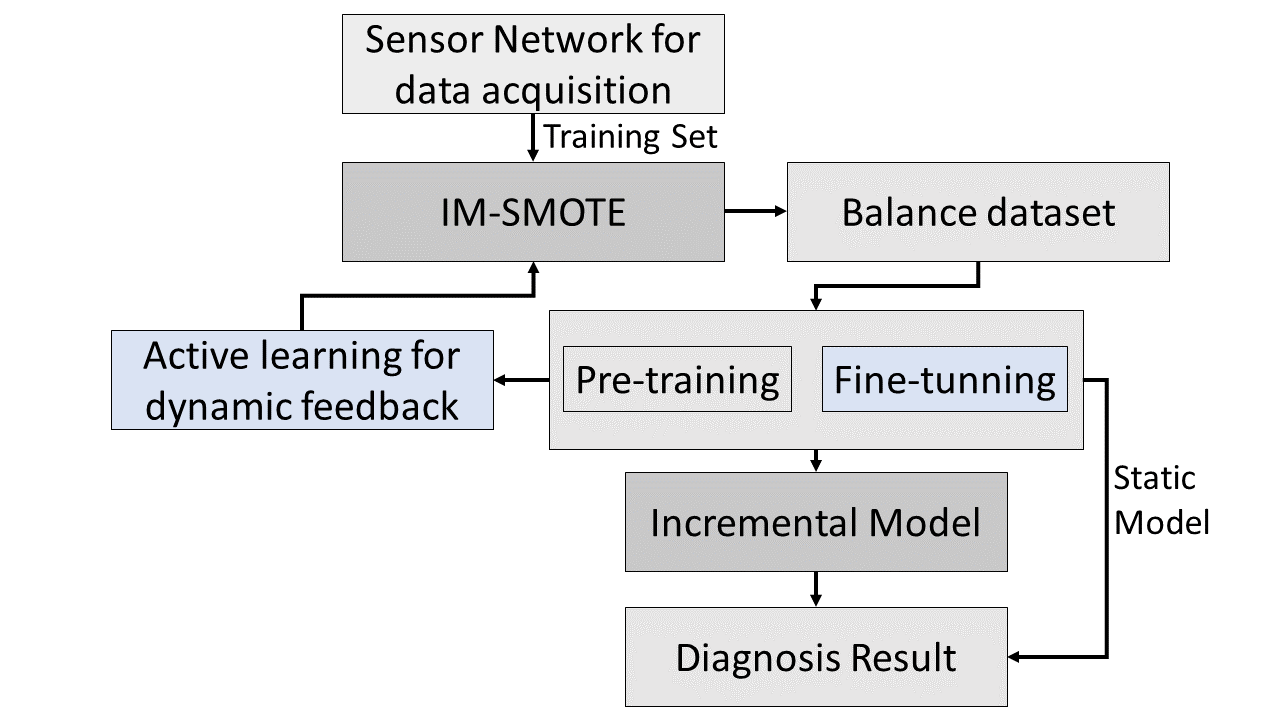}
    \caption{The block diagram of incremental learning during imbalanced chemical fault diagnosis~\cite{b75}}
    \label{fig49}
\end{figure}
\subsection{Event-Based Incremental Broad Learning System for Object Classification}
\begin{figure}[ht!]u
    \centering
    \includegraphics[width=3.5in]{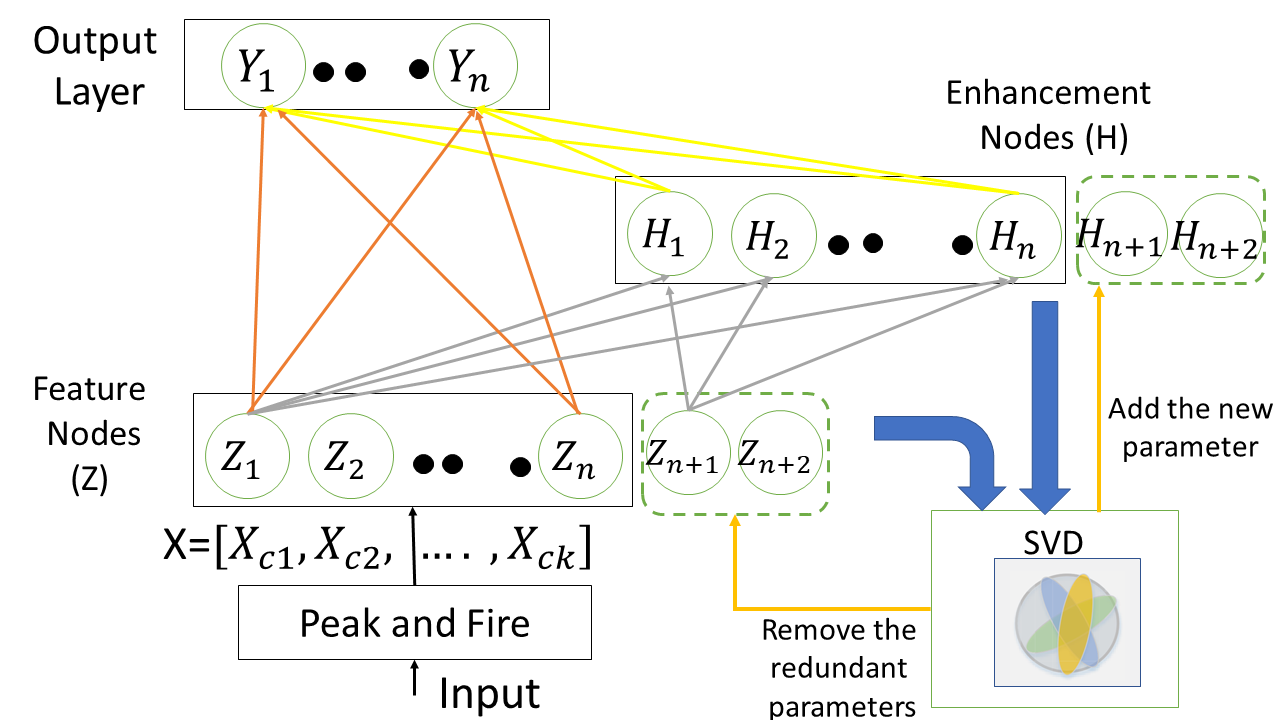}
    \caption{ Event based incremental broad learning: the network expands the feature nodes and enhancement node when the new data arrives~\cite{b76}.}
    \label{fig50}
\end{figure}

The event-based object recognition requires a high-speed learning platform. For this purpose, the incremental board learning system~\cite{b76} is developed that employs a flat neural network structure. It is composed of enhancement and feature nodes, as shown in Fig.~\ref{fig50}. These nodes are expanded dynamically to accommodate new knowledge. Moreover, the SVD is incorporated to avoid the addition of redundant nodes during dynamic expansion. 
\subsection{Incremental Learning for Wireless Big Data Feature Learning}
Big data feature learning is a critical problem for the Internet of Things service management. To learn their features in real-time, a tensor-based deep computation auto-encoder model~\cite{b77} is developed. The incremental learning of this model is performed by incorporating regularization and dynamic expansion-based methods. It also employs the higher-order back-propagation to achieve high learning speed.



\section{Looking Ahead}
\label{section7}
In this survey, we discuss few CL algorithms that addresses the real-world challenges of ML-based autonomous systems. First, we define CL in the context of task-based learning (Section \ref{section2}), in which tasks are assigned one at a time and offline training is performed on the corresponding training data until convergence. Multiple passes over large batches of training data are possible with knowledge of task boundaries. These assumptions aren't directly applicable to real-world scenarios in which an infinite stream of data is received in task-agnostic manner. To addresses actual real-world situations, we focus on discussing online continual learning (OCL) algorithms (Section \ref{section3}) that allow continuous learning from infinite data streams in an online fashion without the notion of task identity. 

We also discuss several practical use cases of CL in autonomous systems (Section \ref{section5}) that help us to find the proper approach to use and identifying similarities between different real-world settings. Since, many use cases are evaluated using image processing experiments or virtual environments, it is important to exactly clarify what would be their expected performance under realistic scenarios. 

We offer a set of recommendations for more concrete indications of what we consider to be important while defining a CL framework in a real-world setting.
\begin{itemize}

\item \textbf{Problem Agnostic}:
 
 Continuous learning  must be problem agnostic that is not limited to a specific setting (e.g., only classification).
 
\item \textbf{Adaptivity}:

CL should exhibit adaptive property that allow to learn from unlabeled data and adapt to specific user data.

\item \textbf{Graceful Forgetting}:
    
In an unbounded learning system with an infinite stream of data, selective forgetting of trivial information is an important mechanism for achieving a balance between stability and plasticity.

\item \textbf{Online Learning}:

To introduce fast acquisition of new information, no offline training of large batches or separate tasks is required.

\item \textbf{Network Complexity}:

When performing architectural dynamic changes, the learning model complexity must be kept below an upper bound of a linear growth in terms of the number of parameters.

\item \textbf{Evaluation of Scalability}:

We recommend evaluating algorithms on as many tasks as possible in order to provide a proper evaluation of scalability and continuous learning performance.

\item \textbf{Memory constraint}:

Realistic CL systems should not assume unlimited memory resources.

\item \textbf{Ablation studies}: 

Ablation studies are recommended on different components of the CL algorithm such as initialization (using pre-trained network or not), optimization methods and hyper-parameters, etc.

\item \textbf{Distributional Shift}:

The mechanism for dealing with distributional shifts must be formally described, not only when tasks change, but also between batches where data points conform to different distributions.

\item \textbf{Benchmarks}:

Complex datasets with realistic and higher resolution are recommended for robust evaluation of the CL algorithms.
\end{itemize}

\section{Conclusion}
\label{section8}
In this article, we presented a comprehensive overview of current CL techniques for real-world autonomous systems. We discussed methods for alleviating catastrophic forgetting with limited supervision and computing resources. Moreover, we summarized the limitations and challenges of existing CL techniques for real-world systems. We also explored the practical use cases of CL in self-driving cars, unmanned aerial vehicles, and urban robots. We hope that this survey will help to better categorize and compare the emerging CL approaches for real-world autonomous systems.

Future research will focus on developing fully autonomous, secure, and reliable learning frameworks which provide superior performance with reduced supervision, training time, and resources. The applicability of many proposed algorithms in a real world setting is rarely addressed. More robust evaluation benchmarks that represent actual real-world scenarios must be investigated.

\section*{Declarations}
\subsection*{Funding}
No funding was received to assist with the preparation of this manuscript.
\subsection*{Conflicts of interest}
The authors declare that they have no conflict of interest.
\subsection*{Code or data availability}
Not Applicable
\subsection*{Authors' Contributions}
This paper has four authors and their names and contributions to the paper are: Khadija Shaheen (conceptualization, writing, and preparation of original draft), Muhammad Abdullah Hanif (writing, reviewing, and editing the whole paper), Dr. Osman Hasan (editing and helping with the revisions) and Dr. Muhammad Shafique (editing and helping with the revisions)
\subsection*{Ethics approval}
This paper does not report research that requires ethical approval.
\subsection*{Consent to participate}
Not Applicable.
\subsection*{Consent for publication}
Not Applicable.


 
\bibliography{sn-bibliography}


\begin{thebibliography}{79}
\ifx \bisbn   \undefined \def \bisbn  #1{ISBN #1}\fi
\ifx \binits  \undefined \def \binits#1{#1}\fi
\ifx \bauthor  \undefined \def \bauthor#1{#1}\fi
\ifx \batitle  \undefined \def \batitle#1{#1}\fi
\ifx \bjtitle  \undefined \def \bjtitle#1{#1}\fi
\ifx \bvolume  \undefined \def \bvolume#1{\textbf{#1}}\fi
\ifx \byear  \undefined \def \byear#1{#1}\fi
\ifx \bissue  \undefined \def \bissue#1{#1}\fi
\ifx \bfpage  \undefined \def \bfpage#1{#1}\fi
\ifx \blpage  \undefined \def \blpage #1{#1}\fi
\ifx \burl  \undefined \def \burl#1{\textsf{#1}}\fi
\ifx \doiurl  \undefined \def \doiurl#1{\url{https://doi.org/#1}}\fi
\ifx \betal  \undefined \def \betal{\textit{et al.}}\fi
\ifx \binstitute  \undefined \def \binstitute#1{#1}\fi
\ifx \binstitutionaled  \undefined \def \binstitutionaled#1{#1}\fi
\ifx \bctitle  \undefined \def \bctitle#1{#1}\fi
\ifx \beditor  \undefined \def \beditor#1{#1}\fi
\ifx \bpublisher  \undefined \def \bpublisher#1{#1}\fi
\ifx \bbtitle  \undefined \def \bbtitle#1{#1}\fi
\ifx \bedition  \undefined \def \bedition#1{#1}\fi
\ifx \bseriesno  \undefined \def \bseriesno#1{#1}\fi
\ifx \blocation  \undefined \def \blocation#1{#1}\fi
\ifx \bsertitle  \undefined \def \bsertitle#1{#1}\fi
\ifx \bsnm \undefined \def \bsnm#1{#1}\fi
\ifx \bsuffix \undefined \def \bsuffix#1{#1}\fi
\ifx \bparticle \undefined \def \bparticle#1{#1}\fi
\ifx \barticle \undefined \def \barticle#1{#1}\fi
\bibcommenthead
\ifx \bconfdate \undefined \def \bconfdate #1{#1}\fi
\ifx \botherref \undefined \def \botherref #1{#1}\fi
\ifx \url \undefined \def \url#1{\textsf{#1}}\fi
\ifx \bchapter \undefined \def \bchapter#1{#1}\fi
\ifx \bbook \undefined \def \bbook#1{#1}\fi
\ifx \bcomment \undefined \def \bcomment#1{#1}\fi
\ifx \oauthor \undefined \def \oauthor#1{#1}\fi
\ifx \citeauthoryear \undefined \def \citeauthoryear#1{#1}\fi
\ifx \endbibitem  \undefined \def \endbibitem {}\fi
\ifx \bconflocation  \undefined \def \bconflocation#1{#1}\fi
\ifx \arxivurl  \undefined \def \arxivurl#1{\textsf{#1}}\fi
\csname PreBibitemsHook\endcsname

\bibitem{b1}
\begin{barticle}
\bauthor{\bsnm{Russakovsky}, \binits{O.}},
\bauthor{\bsnm{Deng}, \binits{J.}},
\bauthor{\bsnm{Su}, \binits{H.}},
\bauthor{\bsnm{Krause}, \binits{J.}},
\bauthor{\bsnm{Satheesh}, \binits{S.}},
\bauthor{\bsnm{Ma}, \binits{S.}},
\bauthor{\bsnm{Huang}, \binits{Z.}},
\bauthor{\bsnm{Karpathy}, \binits{A.}},
\bauthor{\bsnm{Khosla}, \binits{A.}},
\bauthor{\bsnm{Bernstein}, \binits{M.}}, \betal:
\batitle{Imagenet large scale visual recognition challenge}.
\bjtitle{International journal of computer vision}
\bvolume{115}(\bissue{3}),
\bfpage{211}--\blpage{252}
(\byear{2015})
\end{barticle}
\endbibitem

\bibitem{lecun2015deep}
\begin{barticle}
\bauthor{\bsnm{LeCun}, \binits{Y.}},
\bauthor{\bsnm{Bengio}, \binits{Y.}},
\bauthor{\bsnm{Hinton}, \binits{G.}}:
\batitle{Deep learning}.
\bjtitle{nature}
\bvolume{521}(\bissue{7553}),
\bfpage{436}--\blpage{444}
(\byear{2015})
\end{barticle}
\endbibitem

\bibitem{dehghani2019universal}
\begin{botherref}
\oauthor{\bsnm{Dehghani}, \binits{M.}},
\oauthor{\bsnm{Gouws}, \binits{S.}},
\oauthor{\bsnm{Vinyals}, \binits{O.}},
\oauthor{\bsnm{Uszkoreit}, \binits{J.}},
\oauthor{\bparticle{Łukasz} \bsnm{Kaiser}}:
Universal transformers
(2019)
{\href{https://arxiv.org/abs/1807.03819}{{arXiv:1807.03819}}}
\end{botherref}
\endbibitem

\bibitem{b2}
\begin{barticle}
\bauthor{\bsnm{Silver}, \binits{D.}},
\bauthor{\bsnm{Hubert}, \binits{T.}},
\bauthor{\bsnm{Schrittwieser}, \binits{J.}},
\bauthor{\bsnm{Antonoglou}, \binits{I.}},
\bauthor{\bsnm{Lai}, \binits{M.}},
\bauthor{\bsnm{Guez}, \binits{A.}},
\bauthor{\bsnm{Lanctot}, \binits{M.}},
\bauthor{\bsnm{Sifre}, \binits{L.}},
\bauthor{\bsnm{Kumaran}, \binits{D.}},
\bauthor{\bsnm{Graepel}, \binits{T.}}, \betal:
\batitle{A general reinforcement learning algorithm that masters chess, shogi,
  and go through self-play}.
\bjtitle{Science}
\bvolume{362}(\bissue{6419}),
\bfpage{1140}--\blpage{1144}
(\byear{2018})
\end{barticle}
\endbibitem

\bibitem{b3}
\begin{barticle}
\bauthor{\bsnm{French}, \binits{R.M.}}:
\batitle{Catastrophic forgetting in connectionist networks}.
\bjtitle{Trends in cognitive sciences}
\bvolume{3}(\bissue{4}),
\bfpage{128}--\blpage{135}
(\byear{1999})
\end{barticle}
\endbibitem

\bibitem{b4}
\begin{bchapter}
\bauthor{\bsnm{McCloskey}, \binits{M.}},
\bauthor{\bsnm{Cohen}, \binits{N.J.}}:
\bctitle{Catastrophic interference in connectionist networks: The sequential
  learning problem}.
In: \bbtitle{Psychology of Learning and Motivation}
vol. \bseriesno{24},
pp. \bfpage{109}--\blpage{165}
(\byear{1989})
\end{bchapter}
\endbibitem

\bibitem{b5}
\begin{bchapter}
\bauthor{\bsnm{Biesialska}, \binits{M.}},
\bauthor{\bsnm{Biesialska}, \binits{K.}},
\bauthor{\bsnm{Costa-juss{\`a}}, \binits{M.R.}}:
\bctitle{Continual lifelong learning in natural language processing: A survey}.
In: \bbtitle{Proceedings of the 28th International Conference on Computational
  Linguistics},
pp. \bfpage{6523}--\blpage{6541}
(\byear{2020})
\end{bchapter}
\endbibitem

\bibitem{b6}
\begin{botherref}
\oauthor{\bsnm{Parisi}, \binits{G.I.}},
\oauthor{\bsnm{Lomonaco}, \binits{V.}}:
Online continual learning on sequences.
Recent Trends in Learning From Data,
197--221
(2020)
\end{botherref}
\endbibitem

\bibitem{b7}
\begin{botherref}
\oauthor{\bsnm{Grossberg}, \binits{S.T.}}:
Studies of mind and brain: Neural principles of learning, perception,
  development, cognition, and motor control
\textbf{70}
(2012)
\end{botherref}
\endbibitem

\bibitem{b8}
\begin{barticle}
\bauthor{\bsnm{Grossberg}, \binits{S.}}:
\batitle{How does a brain build a cognitive code?}
\bjtitle{Studies of mind and brain}
\bvolume{87},
\bfpage{1}--\blpage{52}
(\byear{1982})
\end{barticle}
\endbibitem

\bibitem{b9}
\begin{barticle}
\bauthor{\bsnm{Parisi}, \binits{G.I.}},
\bauthor{\bsnm{Kemker}, \binits{R.}},
\bauthor{\bsnm{Part}, \binits{J.L.}},
\bauthor{\bsnm{Kanan}, \binits{C.}},
\bauthor{\bsnm{Wermter}, \binits{S.}}:
\batitle{Continual lifelong learning with neural networks: A review}.
\bjtitle{Neural Networks}
\bvolume{113},
\bfpage{54}--\blpage{71}
(\byear{2019})
\end{barticle}
\endbibitem

\bibitem{b10}
\begin{barticle}
\bauthor{\bsnm{Kirkpatrick}, \binits{J.}},
\bauthor{\bsnm{Pascanu}, \binits{R.}},
\bauthor{\bsnm{Rabinowitz}, \binits{N.}},
\bauthor{\bsnm{Veness}, \binits{J.}},
\bauthor{\bsnm{Desjardins}, \binits{G.}},
\bauthor{\bsnm{Rusu}, \binits{A.A.}},
\bauthor{\bsnm{Milan}, \binits{K.}},
\bauthor{\bsnm{Quan}, \binits{J.}},
\bauthor{\bsnm{Ramalho}, \binits{T.}},
\bauthor{\bsnm{Grabska-Barwinska}, \binits{A.}}, \betal:
\batitle{Overcoming catastrophic forgetting in neural networks}.
\bjtitle{Proceedings of the national academy of sciences}
\bvolume{114}(\bissue{13}),
\bfpage{3521}--\blpage{3526}
(\byear{2017})
\end{barticle}
\endbibitem

\bibitem{b11}
\begin{bchapter}
\bauthor{\bsnm{Zenke}, \binits{F.}},
\bauthor{\bsnm{Poole}, \binits{B.}},
\bauthor{\bsnm{Ganguli}, \binits{S.}}:
\bctitle{Continual learning through synaptic intelligence}.
In: \bbtitle{International Conference on Machine Learning},
vol. \bseriesno{70},
pp. \bfpage{3987}--\blpage{3995}
(\byear{2017}).
\bcomment{PMLR}
\end{bchapter}
\endbibitem

\bibitem{b12}
\begin{barticle}
\bauthor{\bsnm{Maltoni}, \binits{D.}},
\bauthor{\bsnm{Lomonaco}, \binits{V.}}:
\batitle{Continuous learning in single-incremental-task scenarios}.
\bjtitle{Neural Networks}
\bvolume{116},
\bfpage{56}--\blpage{73}
(\byear{2019})
\end{barticle}
\endbibitem

\bibitem{b23}
\begin{bchapter}
\bauthor{\bsnm{Aljundi}, \binits{R.}},
\bauthor{\bsnm{Kelchtermans}, \binits{K.}},
\bauthor{\bsnm{Tuytelaars}, \binits{T.}}:
\bctitle{Task-free continual learning}.
In: \bbtitle{Proceedings of the IEEE/CVF Conference on Computer Vision and
  Pattern Recognition},
pp. \bfpage{11254}--\blpage{11263}
(\byear{2019})
\end{bchapter}
\endbibitem

\bibitem{b24}
\begin{botherref}
\oauthor{\bsnm{Pellegrini}, \binits{L.}},
\oauthor{\bsnm{Graffieti}, \binits{G.}},
\oauthor{\bsnm{Lomonaco}, \binits{V.}},
\oauthor{\bsnm{Maltoni}, \binits{D.}}:
Latent replay for real-time continual learning.
In: 2020 IEEE/RSJ International Conference on Intelligent Robots and Systems
  (IROS),
pp. 10203--10209.
IEEE
\end{botherref}
\endbibitem

\bibitem{b25}
\begin{barticle}
\bauthor{\bsnm{Rao}, \binits{D.}},
\bauthor{\bsnm{Visin}, \binits{F.}},
\bauthor{\bsnm{Rusu}, \binits{A.}},
\bauthor{\bsnm{Pascanu}, \binits{R.}},
\bauthor{\bsnm{Teh}, \binits{Y.W.}},
\bauthor{\bsnm{Hadsell}, \binits{R.}}:
\batitle{Continual unsupervised representation learning}.
\bjtitle{Advances in Neural Information Processing Systems}
\bvolume{32},
\bfpage{7647}--\blpage{7657}
(\byear{2019})
\end{barticle}
\endbibitem

\bibitem{b26}
\begin{barticle}
\bauthor{\bsnm{Aljundi}, \binits{R.}},
\bauthor{\bsnm{Belilovsky}, \binits{E.}},
\bauthor{\bsnm{Tuytelaars}, \binits{T.}},
\bauthor{\bsnm{Charlin}, \binits{L.}},
\bauthor{\bsnm{Caccia}, \binits{M.}},
\bauthor{\bsnm{Lin}, \binits{M.}},
\bauthor{\bsnm{Page-Caccia}, \binits{L.}}:
\batitle{Online continual learning with maximal interfered retrieval}.
\bjtitle{Advances in Neural Information Processing Systems}
\bvolume{32},
\bfpage{11849}--\blpage{11860}
(\byear{2019})
\end{barticle}
\endbibitem

\bibitem{b27}
\begin{barticle}
\bauthor{\bsnm{Aljundi}, \binits{R.}},
\bauthor{\bsnm{Lin}, \binits{M.}},
\bauthor{\bsnm{Goujaud}, \binits{B.}},
\bauthor{\bsnm{Bengio}, \binits{Y.}}:
\batitle{Gradient based sample selection for online continual learning}.
\bjtitle{Advances in Neural Information Processing Systems}
\bvolume{32},
\bfpage{11816}--\blpage{11825}
(\byear{2019})
\end{barticle}
\endbibitem

\bibitem{b28}
\begin{bchapter}
\bauthor{\bsnm{Lee}, \binits{S.}},
\bauthor{\bsnm{Ha}, \binits{J.}},
\bauthor{\bsnm{Zhang}, \binits{D.}},
\bauthor{\bsnm{Kim}, \binits{G.}}:
\bctitle{A neural dirichlet process mixture model for task-free continual
  learning}.
In: \bbtitle{International Conference on Learning Representations}
(\byear{2019})
\end{bchapter}
\endbibitem

\bibitem{b29}
\begin{bchapter}
\bauthor{\bsnm{Ebrahimi}, \binits{S.}},
\bauthor{\bsnm{Elhoseiny}, \binits{M.}},
\bauthor{\bsnm{Darrell}, \binits{T.}},
\bauthor{\bsnm{Rohrbach}, \binits{M.}}:
\bctitle{Uncertainty-guided continual learning in bayesian neural networks}.
In: \bbtitle{Proceedings of the IEEE/CVF Conference on Computer Vision and
  Pattern Recognition Workshops},
pp. \bfpage{75}--\blpage{78}
(\byear{2019})
\end{bchapter}
\endbibitem

\bibitem{b30}
\begin{bchapter}
\bauthor{\bsnm{Rajasegaran}, \binits{J.}},
\bauthor{\bsnm{Khan}, \binits{S.}},
\bauthor{\bsnm{Hayat}, \binits{M.}},
\bauthor{\bsnm{Khan}, \binits{F.S.}},
\bauthor{\bsnm{Shah}, \binits{M.}}:
\bctitle{itaml: An incremental task-agnostic meta-learning approach}.
In: \bbtitle{Proceedings of the IEEE/CVF Conference on Computer Vision and
  Pattern Recognition},
pp. \bfpage{13588}--\blpage{13597}
(\byear{2020})
\end{bchapter}
\endbibitem

\bibitem{b31}
\begin{bchapter}
\bauthor{\bsnm{Javed}, \binits{K.}},
\bauthor{\bsnm{White}, \binits{M.}}:
\bctitle{Meta-learning representations for continual learning}.
In: \bbtitle{Proceedings of the 33rd International Conference on Neural
  Information Processing Systems},
pp. \bfpage{1820}--\blpage{1830}
(\byear{2019})
\end{bchapter}
\endbibitem

\bibitem{b39}
\begin{bchapter}
\bauthor{\bsnm{Nose}, \binits{Y.}},
\bauthor{\bsnm{Kojima}, \binits{A.}},
\bauthor{\bsnm{Kawabata}, \binits{H.}},
\bauthor{\bsnm{Hironaka}, \binits{T.}}:
\bctitle{A study on a lane keeping system using cnn for online learning of
  steering control from real time images}.
In: \bbtitle{2019 34th International Technical Conference on Circuits/Systems,
  Computers and Communications (ITC-CSCC)},
pp. \bfpage{1}--\blpage{4}
(\byear{2019}).
\bcomment{IEEE}
\end{bchapter}
\endbibitem

\bibitem{b40}
\begin{bchapter}
\bauthor{\bsnm{{\"O}fj{\"a}ll}, \binits{K.}},
\bauthor{\bsnm{Felsberg}, \binits{M.}},
\bauthor{\bsnm{Robinson}, \binits{A.}}:
\bctitle{Visual autonomous road following by symbiotic online learning}.
In: \bbtitle{2016 IEEE Intelligent Vehicles Symposium (IV)},
pp. \bfpage{136}--\blpage{143}
(\byear{2016}).
\bcomment{IEEE}
\end{bchapter}
\endbibitem

\bibitem{b41}
\begin{bchapter}
\bauthor{\bsnm{Kendall}, \binits{A.}},
\bauthor{\bsnm{Hawke}, \binits{J.}},
\bauthor{\bsnm{Janz}, \binits{D.}},
\bauthor{\bsnm{Mazur}, \binits{P.}},
\bauthor{\bsnm{Reda}, \binits{D.}},
\bauthor{\bsnm{Allen}, \binits{J.-M.}},
\bauthor{\bsnm{Lam}, \binits{V.-D.}},
\bauthor{\bsnm{Bewley}, \binits{A.}},
\bauthor{\bsnm{Shah}, \binits{A.}}:
\bctitle{Learning to drive in a day}.
In: \bbtitle{2019 International Conference on Robotics and Automation (ICRA)},
pp. \bfpage{8248}--\blpage{8254}
(\byear{2019}).
\bcomment{IEEE}
\end{bchapter}
\endbibitem

\bibitem{b42}
\begin{botherref}
\oauthor{\bsnm{Liaw}, \binits{R.}},
\oauthor{\bsnm{Krishnan}, \binits{S.}},
\oauthor{\bsnm{Garg}, \binits{A.}},
\oauthor{\bsnm{Crankshaw}, \binits{D.}},
\oauthor{\bsnm{Gonzalez}, \binits{J.E.}},
\oauthor{\bsnm{Goldberg}, \binits{K.}}:
Composing meta-policies for autonomous driving using hierarchical deep
  reinforcement learning
(2017)
{\href{https://arxiv.org/abs/1711.01503}{{arXiv:1711.01503}}}
\end{botherref}
\endbibitem

\bibitem{b43}
\begin{bchapter}
\bauthor{\bsnm{Klose}, \binits{P.}},
\bauthor{\bsnm{Mester}, \binits{R.}}:
\bctitle{Simulated autonomous driving in a realistic driving environment using
  deep reinforcement learning and a deterministic finite state machine}.
In: \bbtitle{Proceedings of the 2nd International Conference on Applications of
  Intelligent Systems},
pp. \bfpage{1}--\blpage{6}
(\byear{2019})
\end{bchapter}
\endbibitem

\bibitem{b44}
\begin{bchapter}
\bauthor{\bsnm{Soares}, \binits{E.}},
\bauthor{\bsnm{Angelov}, \binits{P.}},
\bauthor{\bsnm{Costa}, \binits{B.}},
\bauthor{\bsnm{Castro}, \binits{M.}}:
\bctitle{Actively semi-supervised deep rule-based classifier applied to adverse
  driving scenarios}.
In: \bbtitle{2019 International Joint Conference on Neural Networks (IJCNN)},
pp. \bfpage{1}--\blpage{8}
(\byear{2019}).
\bcomment{IEEE}
\end{bchapter}
\endbibitem

\bibitem{b45}
\begin{bchapter}
\bauthor{\bsnm{Zaal}, \binits{H.}},
\bauthor{\bsnm{Iqbal}, \binits{H.}},
\bauthor{\bsnm{Campo}, \binits{D.}},
\bauthor{\bsnm{Marcenaro}, \binits{L.}},
\bauthor{\bsnm{Regazzoni}, \binits{C.S.}}:
\bctitle{Incremental learning of abnormalities in autonomous systems}.
In: \bbtitle{2019 16th IEEE International Conference on Advanced Video and
  Signal Based Surveillance (AVSS)},
pp. \bfpage{1}--\blpage{8}
(\byear{2019}).
\bcomment{IEEE}
\end{bchapter}
\endbibitem

\bibitem{b46}
\begin{barticle}
\bauthor{\bsnm{Campo}, \binits{D.}},
\bauthor{\bsnm{Baydoun}, \binits{M.}},
\bauthor{\bsnm{Marin}, \binits{P.}},
\bauthor{\bsnm{Martin}, \binits{D.}},
\bauthor{\bsnm{Marcenaro}, \binits{L.}},
\bauthor{\bparticle{de~la} \bsnm{Escalera}, \binits{A.}},
\bauthor{\bsnm{Regazzoni}, \binits{C.}}:
\batitle{Learning probabilistic awareness models for detecting abnormalities in
  vehicle motions}.
\bjtitle{IEEE Transactions on Intelligent Transportation Systems}
\bvolume{21}(\bissue{3}),
\bfpage{1308}--\blpage{1320}
(\byear{2019})
\end{barticle}
\endbibitem

\bibitem{b47}
\begin{bchapter}
\bauthor{\bsnm{Williams}, \binits{G.R.}},
\bauthor{\bsnm{Goldfain}, \binits{B.}},
\bauthor{\bsnm{Lee}, \binits{K.}},
\bauthor{\bsnm{Gibson}, \binits{J.}},
\bauthor{\bsnm{Rehg}, \binits{J.M.}},
\bauthor{\bsnm{Theodorou}, \binits{E.A.}}:
\bctitle{Locally weighted regression pseudo-rehearsal for adaptive model
  predictive control}.
In: \bbtitle{Conference on Robot Learning},
pp. \bfpage{969}--\blpage{978}
(\byear{2020}).
\bcomment{PMLR}
\end{bchapter}
\endbibitem

\bibitem{b48}
\begin{bchapter}
\bauthor{\bsnm{Chen}, \binits{L.}},
\bauthor{\bsnm{Chen}, \binits{Y.}},
\bauthor{\bsnm{Yao}, \binits{X.}},
\bauthor{\bsnm{Shan}, \binits{Y.}},
\bauthor{\bsnm{Chen}, \binits{L.}}:
\bctitle{An adaptive path tracking controller based on reinforcement learning
  with urban driving application}.
In: \bbtitle{2019 IEEE Intelligent Vehicles Symposium (IV)},
pp. \bfpage{2411}--\blpage{2416}
(\byear{2019}).
\bcomment{IEEE}
\end{bchapter}
\endbibitem

\bibitem{b49}
\begin{bchapter}
\bauthor{\bsnm{Tang}, \binits{C.}},
\bauthor{\bsnm{Chen}, \binits{J.}},
\bauthor{\bsnm{Tomizuka}, \binits{M.}}:
\bctitle{Adaptive probabilistic vehicle trajectory prediction through
  physically feasible bayesian recurrent neural network}.
In: \bbtitle{2019 International Conference on Robotics and Automation (ICRA)},
pp. \bfpage{3846}--\blpage{3852}
(\byear{2019}).
\bcomment{IEEE}
\end{bchapter}
\endbibitem

\bibitem{b50}
\begin{bchapter}
\bauthor{\bsnm{Si}, \binits{W.}},
\bauthor{\bsnm{Wei}, \binits{T.}},
\bauthor{\bsnm{Liu}, \binits{C.}}:
\bctitle{Agen: Adaptable generative prediction networks for autonomous
  driving}.
In: \bbtitle{2019 IEEE Intelligent Vehicles Symposium (IV)},
pp. \bfpage{281}--\blpage{286}
(\byear{2019}).
\bcomment{IEEE}
\end{bchapter}
\endbibitem

\bibitem{b51}
\begin{botherref}
\oauthor{\bsnm{Habibi}, \binits{G.}},
\oauthor{\bsnm{Japuria}, \binits{N.}},
\oauthor{\bsnm{How}, \binits{J.P.}}:
Incremental learning of motion primitives for pedestrian trajectory prediction
  at intersections
(2019)
{\href{https://arxiv.org/abs/1911.09476}{{arXiv:1911.09476}}}
\end{botherref}
\endbibitem

\bibitem{b52}
\begin{bchapter}
\bauthor{\bsnm{Abdellatif}, \binits{A.A.}},
\bauthor{\bsnm{Chiasserini}, \binits{C.F.}},
\bauthor{\bsnm{Malandrino}, \binits{F.}}:
\bctitle{Active learning-based classification in automated connected vehicles}.
In: \bbtitle{IEEE INFOCOM 2020-IEEE Conference on Computer Communications
  Workshops (INFOCOM WKSHPS)},
pp. \bfpage{598}--\blpage{603}
(\byear{2020}).
\bcomment{IEEE}
\end{bchapter}
\endbibitem

\bibitem{b53}
\begin{bchapter}
\bauthor{\bsnm{Olariu}, \binits{C.}},
\bauthor{\bsnm{Assem}, \binits{H.}},
\bauthor{\bsnm{Ortega}, \binits{J.D.}},
\bauthor{\bsnm{Nieto}, \binits{M.}}:
\bctitle{A cloud-based ai framework for machine learning orchestration: A
  “driving or not-driving” case-study for self-driving cars}.
In: \bbtitle{2019 IEEE Intelligent Vehicles Symposium (IV)},
pp. \bfpage{1715}--\blpage{1722}
(\byear{2019}).
\bcomment{IEEE}
\end{bchapter}
\endbibitem

\bibitem{b54}
\begin{bchapter}
\bauthor{\bsnm{Cui}, \binits{Y.}},
\bauthor{\bsnm{Isele}, \binits{D.}},
\bauthor{\bsnm{Niekum}, \binits{S.}},
\bauthor{\bsnm{Fujimura}, \binits{K.}}:
\bctitle{Uncertainty-aware data aggregation for deep imitation learning}.
In: \bbtitle{2019 International Conference on Robotics and Automation (ICRA)},
pp. \bfpage{761}--\blpage{767}
(\byear{2019}).
\bcomment{IEEE}
\end{bchapter}
\endbibitem

\bibitem{b55}
\begin{bchapter}
\bauthor{\bsnm{Pierre}, \binits{J.M.}}:
\bctitle{Incremental lifelong deep learning for autonomous vehicles}.
In: \bbtitle{2018 21st International Conference on Intelligent Transportation
  Systems (ITSC)},
pp. \bfpage{3949}--\blpage{3954}
(\byear{2018}).
\bcomment{IEEE}
\end{bchapter}
\endbibitem

\bibitem{b57}
\begin{bchapter}
\bauthor{\bsnm{Sarabakha}, \binits{A.}},
\bauthor{\bsnm{Kayacan}, \binits{E.}}:
\bctitle{Online deep learning for improved trajectory tracking of unmanned
  aerial vehicles using expert knowledge}.
In: \bbtitle{2019 International Conference on Robotics and Automation (ICRA)},
pp. \bfpage{7727}--\blpage{7733}
(\byear{2019}).
\bcomment{IEEE}
\end{bchapter}
\endbibitem

\bibitem{b58}
\begin{barticle}
\bauthor{\bsnm{Sarabakha}, \binits{A.}},
\bauthor{\bsnm{Kayacan}, \binits{E.}}:
\batitle{Online deep fuzzy learning for control of nonlinear systems using
  expert knowledge}.
\bjtitle{IEEE Transactions on Fuzzy Systems}
\bvolume{28}(\bissue{7}),
\bfpage{1492}--\blpage{1503}
(\byear{2019})
\end{barticle}
\endbibitem

\bibitem{b59}
\begin{barticle}
\bauthor{\bsnm{Ferdaus}, \binits{M.M.}},
\bauthor{\bsnm{Pratama}, \binits{M.}},
\bauthor{\bsnm{Anavatti}, \binits{S.G.}},
\bauthor{\bsnm{Garratt}, \binits{M.A.}}:
\batitle{Online identification of a rotary wing unmanned aerial vehicle from
  data streams}.
\bjtitle{Applied Soft Computing}
\bvolume{76},
\bfpage{313}--\blpage{325}
(\byear{2019})
\end{barticle}
\endbibitem

\bibitem{b60}
\begin{botherref}
\oauthor{\bsnm{Maciel-Pearson}, \binits{B.G.}},
\oauthor{\bsnm{Marchegiani}, \binits{L.}},
\oauthor{\bsnm{Akcay}, \binits{S.}},
\oauthor{\bsnm{Atapour-Abarghouei}, \binits{A.}},
\oauthor{\bsnm{Garforth}, \binits{J.}},
\oauthor{\bsnm{Breckon}, \binits{T.P.}}:
Online deep reinforcement learning for autonomous uav navigation and
  exploration of outdoor environments
(2019)
{\href{https://arxiv.org/abs/1912.05684}{{arXiv:1912.05684}}}
\end{botherref}
\endbibitem

\bibitem{b61}
\begin{bchapter}
\bauthor{\bsnm{Sun}, \binits{Z.}},
\bauthor{\bsnm{Wang}, \binits{Y.}},
\bauthor{\bsnm{Lagani{\`e}re}, \binits{R.}}:
\bctitle{Online model adaptation for uav tracking with convolutional neural
  network}.
In: \bbtitle{2018 15th Conference on Computer and Robot Vision (CRV)},
pp. \bfpage{329}--\blpage{336}
(\byear{2018}).
\bcomment{IEEE}
\end{bchapter}
\endbibitem

\bibitem{b62}
\begin{bchapter}
\bauthor{\bsnm{Wehbe}, \binits{B.}},
\bauthor{\bsnm{Hildebrandt}, \binits{M.}},
\bauthor{\bsnm{Kirchner}, \binits{F.}}:
\bctitle{A framework for on-line learning of underwater vehicles dynamic
  models}.
In: \bbtitle{2019 International Conference on Robotics and Automation (ICRA)},
pp. \bfpage{7969}--\blpage{7975}
(\byear{2019}).
\bcomment{IEEE}
\end{bchapter}
\endbibitem

\bibitem{b63}
\begin{bchapter}
\bauthor{\bsnm{Chen}, \binits{S.}},
\bauthor{\bsnm{Wen}, \binits{J.T.}}:
\bctitle{Adaptive neural trajectory tracking control for flexible-joint robots
  with online learning}.
In: \bbtitle{2020 IEEE International Conference on Robotics and Automation
  (ICRA)},
pp. \bfpage{2358}--\blpage{2364}
(\byear{2020}).
\bcomment{IEEE}
\end{bchapter}
\endbibitem

\bibitem{b65}
\begin{botherref}
\oauthor{\bsnm{Julian}, \binits{R.}},
\oauthor{\bsnm{Swanson}, \binits{B.}},
\oauthor{\bsnm{Sukhatme}, \binits{G.S.}},
\oauthor{\bsnm{Levine}, \binits{S.}},
\oauthor{\bsnm{Finn}, \binits{C.}},
\oauthor{\bsnm{Hausman}, \binits{K.}}:
Never stop learning: The effectiveness of fine-tuning in robotic reinforcement
  learning
(2020)
{\href{https://arxiv.org/abs/2004.10190}{{arXiv:2004.10190}}}
\end{botherref}
\endbibitem

\bibitem{b66}
\begin{bchapter}
\bauthor{\bsnm{Mancini}, \binits{M.}},
\bauthor{\bsnm{Karaoguz}, \binits{H.}},
\bauthor{\bsnm{Ricci}, \binits{E.}},
\bauthor{\bsnm{Jensfelt}, \binits{P.}},
\bauthor{\bsnm{Caputo}, \binits{B.}}:
\bctitle{Knowledge is never enough: Towards web aided deep open world
  recognition}.
In: \bbtitle{2019 International Conference on Robotics and Automation (ICRA)},
pp. \bfpage{9537}--\blpage{9543}
(\byear{2019}).
\bcomment{IEEE}
\end{bchapter}
\endbibitem

\bibitem{b67}
\begin{barticle}
\bauthor{\bsnm{Zheng}, \binits{W.}},
\bauthor{\bsnm{Liu}, \binits{H.}},
\bauthor{\bsnm{Sun}, \binits{F.}}:
\batitle{Lifelong visual-tactile cross-modal learning for robotic material
  perception}.
\bjtitle{IEEE transactions on neural networks and learning systems}
\bvolume{32}(\bissue{3}),
\bfpage{1192}--\blpage{1203}
(\byear{2020})
\end{barticle}
\endbibitem

\bibitem{b68}
\begin{bchapter}
\bauthor{\bsnm{Liu}, \binits{H.}},
\bauthor{\bsnm{Zhang}, \binits{Z.}},
\bauthor{\bsnm{Zhu}, \binits{Y.}},
\bauthor{\bsnm{Zhu}, \binits{S.-C.}}:
\bctitle{Self-supervised incremental learning for sound source localization in
  complex indoor environment}.
In: \bbtitle{2019 International Conference on Robotics and Automation (ICRA)},
pp. \bfpage{2599}--\blpage{2605}
(\byear{2019}).
\bcomment{IEEE}
\end{bchapter}
\endbibitem

\bibitem{b69}
\begin{bchapter}
\bauthor{\bsnm{Dehghan}, \binits{M.}},
\bauthor{\bsnm{Zhang}, \binits{Z.}},
\bauthor{\bsnm{Siam}, \binits{M.}},
\bauthor{\bsnm{Jin}, \binits{J.}},
\bauthor{\bsnm{Petrich}, \binits{L.}},
\bauthor{\bsnm{Jagersand}, \binits{M.}}:
\bctitle{Online object and task learning via human robot interaction}.
In: \bbtitle{2019 International Conference on Robotics and Automation (ICRA)},
pp. \bfpage{2132}--\blpage{2138}
(\byear{2019}).
\bcomment{IEEE}
\end{bchapter}
\endbibitem

\bibitem{b70}
\begin{barticle}
\bauthor{\bsnm{Kahn}, \binits{G.}},
\bauthor{\bsnm{Abbeel}, \binits{P.}},
\bauthor{\bsnm{Levine}, \binits{S.}}:
\batitle{Badgr: An autonomous self-supervised learning-based navigation
  system}.
\bjtitle{IEEE Robotics and Automation Letters}
\bvolume{6}(\bissue{2}),
\bfpage{1312}--\blpage{1319}
(\byear{2021})
\end{barticle}
\endbibitem

\bibitem{b64}
\begin{bchapter}
\bauthor{\bsnm{Losing}, \binits{V.}},
\bauthor{\bsnm{Yoshikawa}, \binits{T.}},
\bauthor{\bsnm{Hasenjaeger}, \binits{M.}},
\bauthor{\bsnm{Hammer}, \binits{B.}},
\bauthor{\bsnm{Wersing}, \binits{H.}}:
\bctitle{Personalized online learning of whole-body motion classes using
  multiple inertial measurement units}.
In: \bbtitle{2019 International Conference on Robotics and Automation (ICRA)},
pp. \bfpage{9530}--\blpage{9536}
(\byear{2019}).
\bcomment{IEEE}
\end{bchapter}
\endbibitem

\bibitem{b13}
\begin{barticle}
\bauthor{\bsnm{Lesort}, \binits{T.}},
\bauthor{\bsnm{Lomonaco}, \binits{V.}},
\bauthor{\bsnm{Stoian}, \binits{A.}},
\bauthor{\bsnm{Maltoni}, \binits{D.}},
\bauthor{\bsnm{Filliat}, \binits{D.}},
\bauthor{\bsnm{D{\'\i}az-Rodr{\'\i}guez}, \binits{N.}}:
\batitle{Continual learning for robotics: Definition, framework, learning
  strategies, opportunities and challenges}.
\bjtitle{Information fusion}
\bvolume{58},
\bfpage{52}--\blpage{68}
(\byear{2020})
\end{barticle}
\endbibitem

\bibitem{b38}
\begin{barticle}
\bauthor{\bsnm{Chen}, \binits{Z.}},
\bauthor{\bsnm{Liu}, \binits{B.}}:
\batitle{Lifelong machine learning}.
\bjtitle{Synthesis Lectures on Artificial Intelligence and Machine Learning}
\bvolume{12}(\bissue{3}),
\bfpage{1}--\blpage{207}
(\byear{2018})
\end{barticle}
\endbibitem

\bibitem{b78}
\begin{botherref}
\oauthor{\bparticle{van~de} \bsnm{Ven}, \binits{G.M.}},
\oauthor{\bsnm{Tolias}, \binits{A.S.}}:
Three scenarios for continual learning
(2019)
{\href{https://arxiv.org/abs/1904.07734}{{arXiv:1904.07734}}}
\end{botherref}
\endbibitem

\bibitem{b14}
\begin{bchapter}
\bauthor{\bsnm{Rebuffi}, \binits{S.-A.}},
\bauthor{\bsnm{Kolesnikov}, \binits{A.}},
\bauthor{\bsnm{Sperl}, \binits{G.}},
\bauthor{\bsnm{Lampert}, \binits{C.H.}}:
\bctitle{icarl: Incremental classifier and representation learning}.
In: \bbtitle{Proceedings of the IEEE Conference on Computer Vision and Pattern
  Recognition},
pp. \bfpage{2001}--\blpage{2010}
(\byear{2017})
\end{bchapter}
\endbibitem

\bibitem{b15}
\begin{bchapter}
\bauthor{\bsnm{Shin}, \binits{H.}},
\bauthor{\bsnm{Lee}, \binits{J.K.}},
\bauthor{\bsnm{Kim}, \binits{J.}},
\bauthor{\bsnm{Kim}, \binits{J.}}:
\bctitle{Continual learning with deep generative replay}.
In: \bbtitle{Proceedings of the 31st International Conference on Neural
  Information Processing Systems},
pp. \bfpage{2994}--\blpage{3003}
(\byear{2017})
\end{bchapter}
\endbibitem

\bibitem{b16}
\begin{botherref}
\oauthor{\bsnm{Kemker}, \binits{R.}},
\oauthor{\bsnm{Kanan}, \binits{C.}}:
Fearnet: Brain-inspired model for incremental learning
(2018)
{\href{https://arxiv.org/abs/1711.10563}{{arXiv:1711.10563}}}
\end{botherref}
\endbibitem

\bibitem{b17}
\begin{barticle}
\bauthor{\bsnm{Kirkpatrick}, \binits{J.}},
\bauthor{\bsnm{Pascanu}, \binits{R.}},
\bauthor{\bsnm{Rabinowitz}, \binits{N.}},
\bauthor{\bsnm{Veness}, \binits{J.}},
\bauthor{\bsnm{Desjardins}, \binits{G.}},
\bauthor{\bsnm{Rusu}, \binits{A.A.}},
\bauthor{\bsnm{Milan}, \binits{K.}},
\bauthor{\bsnm{Quan}, \binits{J.}},
\bauthor{\bsnm{Ramalho}, \binits{T.}},
\bauthor{\bsnm{Grabska-Barwinska}, \binits{A.}}, \betal:
\batitle{Overcoming catastrophic forgetting in neural networks}.
\bjtitle{Proceedings of the national academy of sciences}
\bvolume{114}(\bissue{13}),
\bfpage{3521}--\blpage{3526}
(\byear{2017})
\end{barticle}
\endbibitem

\bibitem{b19}
\begin{bchapter}
\bauthor{\bsnm{Wei}, \binits{H.-R.}},
\bauthor{\bsnm{Huang}, \binits{S.}},
\bauthor{\bsnm{Wang}, \binits{R.}},
\bauthor{\bsnm{Dai}, \binits{X.}},
\bauthor{\bsnm{Chen}, \binits{J.}}:
\bctitle{Online distilling from checkpoints for neural machine translation}.
In: \bbtitle{Proceedings of the 2019 Conference of the North American Chapter
  of the Association for Computational Linguistics: Human Language
  Technologies, Volume 1 (Long and Short Papers)},
pp. \bfpage{1932}--\blpage{1941}
(\byear{2019})
\end{bchapter}
\endbibitem

\bibitem{b18}
\begin{barticle}
\bauthor{\bsnm{Li}, \binits{Z.}},
\bauthor{\bsnm{Hoiem}, \binits{D.}}:
\batitle{Learning without forgetting}.
\bjtitle{IEEE transactions on pattern analysis and machine intelligence}
\bvolume{40}(\bissue{12}),
\bfpage{2935}--\blpage{2947}
(\byear{2017})
\end{barticle}
\endbibitem

\bibitem{b21}
\begin{botherref}
\oauthor{\bsnm{Rusu}, \binits{A.A.}},
\oauthor{\bsnm{Rabinowitz}, \binits{N.C.}},
\oauthor{\bsnm{Desjardins}, \binits{G.}},
\oauthor{\bsnm{Soyer}, \binits{H.}},
\oauthor{\bsnm{Kirkpatrick}, \binits{J.}},
\oauthor{\bsnm{Kavukcuoglu}, \binits{K.}},
\oauthor{\bsnm{Pascanu}, \binits{R.}},
\oauthor{\bsnm{Hadsell}, \binits{R.}}:
Progressive neural networks
(2016)
{\href{https://arxiv.org/abs/1606.04671}{{arXiv:1606.04671}}}
\end{botherref}
\endbibitem

\bibitem{b22}
\begin{botherref}
\oauthor{\bsnm{Yoon}, \binits{J.}},
\oauthor{\bsnm{Yang}, \binits{E.}},
\oauthor{\bsnm{Lee}, \binits{J.}},
\oauthor{\bsnm{Hwang}, \binits{S.J.}}:
Lifelong learning with dynamically expandable networks
(2018)
{\href{https://arxiv.org/abs/1708.01547}{{arXiv:1708.01547}}}
\end{botherref}
\endbibitem

\bibitem{b32}
\begin{botherref}
\oauthor{\bsnm{Wong}, \binits{J.M.}}:
Towards lifelong self-supervision: A deep learning direction for robotics
(2016)
{\href{https://arxiv.org/abs/1611.00201}{{arXiv:1611.00201}}}
\end{botherref}
\endbibitem

\bibitem{b35}
\begin{barticle}
\bauthor{\bsnm{Cangelosi}, \binits{A.}},
\bauthor{\bsnm{Schlesinger}, \binits{M.}}:
\batitle{From babies to robots: the contribution of developmental robotics to
  developmental psychology}.
\bjtitle{Child Development Perspectives}
\bvolume{12}(\bissue{3}),
\bfpage{183}--\blpage{188}
(\byear{2018})
\end{barticle}
\endbibitem

\bibitem{b33}
\begin{botherref}
\oauthor{\bsnm{Burda}, \binits{Y.}},
\oauthor{\bsnm{Edwards}, \binits{H.}},
\oauthor{\bsnm{Pathak}, \binits{D.}},
\oauthor{\bsnm{Storkey}, \binits{A.}},
\oauthor{\bsnm{Darrell}, \binits{T.}},
\oauthor{\bsnm{Efros}, \binits{A.A.}}:
Large-scale study of curiosity-driven learning
(2018)
{\href{https://arxiv.org/abs/1808.04355}{{arXiv:1808.04355}}}
\end{botherref}
\endbibitem

\bibitem{b34}
\begin{bchapter}
\bauthor{\bsnm{Doersch}, \binits{C.}},
\bauthor{\bsnm{Zisserman}, \binits{A.}}:
\bctitle{Multi-task self-supervised visual learning}.
In: \bbtitle{Proceedings of the IEEE International Conference on Computer
  Vision},
pp. \bfpage{2051}--\blpage{2060}
(\byear{2017})
\end{bchapter}
\endbibitem

\bibitem{b37}
\begin{barticle}
\bauthor{\bsnm{Minsky}, \binits{M.}}:
\batitle{Steps toward artificial intelligence}.
\bjtitle{Proceedings of the IRE}
\bvolume{49}(\bissue{1}),
\bfpage{8}--\blpage{30}
(\byear{1961})
\end{barticle}
\endbibitem

\bibitem{b36}
\begin{barticle}
\bauthor{\bsnm{Gopnik}, \binits{A.}}:
\batitle{How babies think}.
\bjtitle{Scientific American}
\bvolume{303}(\bissue{1}),
\bfpage{76}--\blpage{81}
(\byear{2010})
\end{barticle}
\endbibitem

\bibitem{russell2009artificial}
\begin{barticle}
\bauthor{\bsnm{Russell}, \binits{S.}},
\bauthor{\bsnm{Norvig}, \binits{P.}}:
\batitle{Artificial intelligence: A modern approach. third edit}.
\bjtitle{Prentice Hall. doi}
\bvolume{10},
\bfpage{978}--\blpage{012161964}
(\byear{2010})
\end{barticle}
\endbibitem

\bibitem{b71}
\begin{barticle}
\bauthor{\bsnm{Nawaratne}, \binits{R.}},
\bauthor{\bsnm{Alahakoon}, \binits{D.}},
\bauthor{\bsnm{De~Silva}, \binits{D.}},
\bauthor{\bsnm{Yu}, \binits{X.}}:
\batitle{Spatiotemporal anomaly detection using deep learning for real-time
  video surveillance}.
\bjtitle{IEEE Transactions on Industrial Informatics}
\bvolume{16}(\bissue{1}),
\bfpage{393}--\blpage{402}
(\byear{2019})
\end{barticle}
\endbibitem

\bibitem{b72}
\begin{bchapter}
\bauthor{\bsnm{Liang}, \binits{F.}},
\bauthor{\bsnm{Hatcher}, \binits{W.G.}},
\bauthor{\bsnm{Xu}, \binits{G.}},
\bauthor{\bsnm{Nguyen}, \binits{J.}},
\bauthor{\bsnm{Liao}, \binits{W.}},
\bauthor{\bsnm{Yu}, \binits{W.}}:
\bctitle{Towards online deep learning-based energy forecasting}.
In: \bbtitle{2019 28th International Conference on Computer Communication and
  Networks (ICCCN)},
pp. \bfpage{1}--\blpage{9}
(\byear{2019}).
\bcomment{IEEE}
\end{bchapter}
\endbibitem

\bibitem{b73}
\begin{bchapter}
\bauthor{\bsnm{Arag{\'o}n}, \binits{G.}},
\bauthor{\bsnm{Puri}, \binits{H.}},
\bauthor{\bsnm{Grass}, \binits{A.}},
\bauthor{\bsnm{Chala}, \binits{S.}},
\bauthor{\bsnm{Beecks}, \binits{C.}}:
\bctitle{Incremental deep-learning for continuous load prediction in energy
  management systems}.
In: \bbtitle{2019 IEEE Milan PowerTech},
pp. \bfpage{1}--\blpage{6}
(\byear{2019}).
\bcomment{IEEE}
\end{bchapter}
\endbibitem

\bibitem{b74}
\begin{bchapter}
\bauthor{\bsnm{Tonioni}, \binits{A.}},
\bauthor{\bsnm{Tosi}, \binits{F.}},
\bauthor{\bsnm{Poggi}, \binits{M.}},
\bauthor{\bsnm{Mattoccia}, \binits{S.}},
\bauthor{\bsnm{Stefano}, \binits{L.D.}}:
\bctitle{Real-time self-adaptive deep stereo}.
In: \bbtitle{Proceedings of the IEEE/CVF Conference on Computer Vision and
  Pattern Recognition},
pp. \bfpage{195}--\blpage{204}
(\byear{2019})
\end{bchapter}
\endbibitem

\bibitem{b75}
\begin{barticle}
\bauthor{\bsnm{Hu}, \binits{Z.}},
\bauthor{\bsnm{Jiang}, \binits{P.}}:
\batitle{An imbalance modified deep neural network with dynamical incremental
  learning for chemical fault diagnosis}.
\bjtitle{IEEE Transactions on Industrial Electronics}
\bvolume{66}(\bissue{1}),
\bfpage{540}--\blpage{550}
(\byear{2018})
\end{barticle}
\endbibitem

\bibitem{b76}
\begin{bchapter}
\bauthor{\bsnm{Gao}, \binits{S.}},
\bauthor{\bsnm{Guo}, \binits{G.}},
\bauthor{\bsnm{Philip~Chen}, \binits{C.}}:
\bctitle{Event-based incremental broad learning system for object
  classification}.
In: \bbtitle{Proceedings of the IEEE/CVF International Conference on Computer
  Vision Workshops},
pp. \bfpage{0}--\blpage{0}
(\byear{2019})
\end{bchapter}
\endbibitem

\bibitem{b77}
\begin{barticle}
\bauthor{\bsnm{Zhang}, \binits{Q.}},
\bauthor{\bsnm{Yang}, \binits{L.T.}},
\bauthor{\bsnm{Chen}, \binits{Z.}},
\bauthor{\bsnm{Li}, \binits{P.}}:
\batitle{Incremental deep computation model for wireless big data feature
  learning}.
\bjtitle{IEEE Transactions on Big Data}
\bvolume{6}(\bissue{2}),
\bfpage{248}--\blpage{257}
(\byear{2019})
\end{barticle}
\endbibitem

\end{thebibliography}


\end{document}